\documentclass[10pt, a4paper]{dukenlp}

\usepackage[utf8]{inputenc}
\usepackage[T1]{fontenc}
\usepackage{hyperref}
\usepackage{url}

\usepackage{natbib}
\usepackage{subcaption}
\usepackage{latexsym}
\usepackage{graphicx}
\usepackage{amsmath}
\usepackage{amsfonts}
\usepackage{nicefrac}
\usepackage{bbm}
\usepackage{listings}
\usepackage{booktabs}
\usepackage{multirow}
\usepackage{comment}
\usepackage{xspace}
\usepackage{dialogue}
\usepackage{algorithm}
\usepackage{algorithmic}
\usepackage{array}
\usepackage{tabularx}

\usepackage{wrapfig}      
\PassOptionsToPackage{table}{xcolor}
\usepackage{xcolor}
\usepackage{float}
\usepackage{microtype}
\usepackage{pifont}
\usepackage{makecell}
\usepackage{tikz}
\usepackage{longtable}
\usepackage{enumitem}
\usepackage{pdflscape}
\usepackage{needspace}
\usepackage[most]{tcolorbox}
\usepackage{caption}      
\definecolor{darkblue}{RGB}{0,51,102}

\newtcolorbox{promptbox}[1]{
  enhanced,
  breakable,
  colback=gray!3,
  colframe=darkblue,
  boxrule=0.5pt,
  left=6pt,
  right=6pt,
  top=6pt,
  bottom=6pt,
  sharp corners,
  title={#1},
  fonttitle=\bfseries\small\sffamily,
  coltitle=black,
  colbacktitle=gray!10,
  before upper=\small\sffamily\setlength{\parskip}{3pt}\setlength{\parindent}{0pt}
}

\newcommand{\cmark}{\textcolor{green!60!black}{\ding{51}}}
\newcommand{\xmark}{\textcolor{red!70!black}{\ding{55}}}

\newcommand{\bandcolor}[1]{%
  \ifdim#1pt<1pt red!12%
  \else\ifdim#1pt<2pt orange!13%
  \else\ifdim#1pt<3pt yellow!13%
  \else\ifdim#1pt<4pt yellow!20%
  \else green!17%
  \fi\fi\fi\fi}

\newcommand{\score}[1]{\cellcolor{\bandcolor{#1}}#1}

\newcolumntype{C}[1]{>{\centering\arraybackslash}m{#1}}

\title{Vision2Code: A Multi-Domain Benchmark for Evaluating Image-to-Code Generation}

\author{
Ajay Vikram Periasami$^{1}$, Junlin Wang$^{1}$, Bhuwan Dhingra$^{1}$ \\
$^{1}$Duke University\\
\texttt{ajay.vikram@duke.edu, junlin.wang2@duke.edu, bdhingra@cs.duke.edu}
}

\begin{document}

\maketitle

\begin{abstract}
We study whether vision-language models (VLMs) can reconstruct structured, graphics-based images by writing executable code to generate them.
Existing benchmarks on image-to-code generation are generally limited in the domains they cover,
rely on reference source code for evaluation,
and do not provide metrics to quantify domain-specific issues.
To address this, we introduce \textbf{Vision2Code}, a multi-domain, reference-code-free benchmark containing 539 test examples from 15 different datasets,
spanning
charts and plots, geometry, graphs, scientific diagrams, documents and 3D spatial scenes.
Evaluation is done by executing the generated programs and comparing their rendered outputs to
the source images using
dataset- and domain-specific rubrics for a VLM judge.
We find that the final scores obtained in this manner have substantially higher 
correlation with human ratings than
those obtained via generic VLM rubrics or standard pixel-based, embedding-based or perceptual similarity metrics. 
We further demonstrate the reliability of our evaluation under
repeated sampling from the VLM judges,
varying rubric weights,
and in the presence of controlled errors.
Our results highlight the variability across domains of existing image-to-code generation models,
with accuracy on charts and graphs being significantly higher than other domains.
Lastly, we demonstrate the utility of our scoring method for rejection sampling of generated programs for fine-tuning, improving Qwen3.5-9B by $16\%$ and Phi-4-MM by $28\%$ compared to their base checkpoints.
\end{abstract}

\section{Introduction}
\label{sec:introduction}

Image-to-code generation evaluates whether a vision-language model (VLM) can write executable code to reproduce an image. Unlike raw pixel generation, code provides an editable, interpretable, and verifiable representation, which is also what makes tools like LaTeX, Excalidraw, and Figma useful. It is also a demanding test of visual understanding. A model might correctly summarize a chart but still get the axis scale wrong, or identify the nodes in a graph while reversing the edge directions. Reconstructing the image through code requires the model to preserve these details, rather than just understanding the overall visual content. While prior benchmarks have made progress in studying image-to-code generation, there are still several limitations. Most of them focus on a particular visual domain. ChartMimic~\citep{yang2025chartmimic}, Plot2Code~\citep{wu2025plot2code}, RealChart2Code~\citep{zhang2026realchart2code}, and Chart2Code~\citep{tang2025chart2code} evaluate charts and plots, while Design2Code~\citep{si-etal-2025-design2code}, Web2Code~\citep{yun2024webcode}, and Widget2Code~\citep{zhang2026widget2code} focus on webpages and UI elements. Another limitation is the dependence on reference source code. Broader benchmarks like Omni-I2C~\citep{zhou2026omni} require the original rendering code, which makes code-level evaluation straightforward but also restricts which images can be included. For most real-world visual data, this code is simply not available. Another issue is the evaluation of model-generated images. Current methods often use the same evaluation criteria across very different domains~\citep{si-etal-2025-design2code,roberts2024image2struct,zhou2026omni} which makes it hard to detect fine-grained domain-specific errors. If the evaluator cannot detect such errors, it becomes difficult to reliably measure reconstruction quality across domains without reference code.

To bridge this gap, we introduce \textbf{Vision2Code}, a multi-domain, reference-code-free benchmark containing 539 test examples from 15 different datasets, spanning charts and plots, geometry, graphs, scientific diagrams, documents and 3D spatial scenes. Given the heterogeneous nature of the dataset with lack of source code, we perform evaluation on the rendered images and not their intermediate code. The images are compared against the source using dataset-specific rubrics and guardrails, where the different categories were semantically motivated based on visual inspection of different failure modes. We compared these scores against ratings from four human annotators and found stronger correlations than generic VLM rubrics and traditional metrics, including SSIM, PSNR, CLIP ViT-B/32, DreamSim, and Qwen3-VL-Embedding-8B cosine similarity. Another aspect we studied was VLM evaluator robustness, by modifying generated code to see if it catches the fine-grained details for respective domains. We also compared two independent VLM judges, repeated the evaluations and studied the sensitivity to rubric weights. Both judges gave the exact same leaderboard ranking across all nine models, with very little variance between runs, and were able to catch most of the manually introduced errors. We also found that perturbing the weights still preserves the leaderboard ranking, suggesting that the results are not overly dependent on the particular choice of weights. From the evaluation of nine open-weight and proprietary models, we observe that image-to-code performance varies considerably across domains. While models recreate charts and graphs relatively well, they still struggle with chemistry, scientific diagrams, document layouts, and 3D spatial scenes. Since ground-truth source code is missing from most visual datasets, we also investigated whether the evaluator could be used to select training data from a model's own generations. Our evaluator-guided self-training approach filters image-code pairs by identifying cases where the model generates a good reconstruction initially but struggles to reproduce the same output when it is given back as input. Fine-tuning with these selected examples improves Qwen3.5-9B from 1.60 to 1.86 and Phi-4-MM from 0.58 to 0.74, with the gains holding under independent evaluators as well.

In summary, our contributions include: (1) \textbf{Vision2Code}, a multi-domain reference-code-free benchmark for image-to-code generation across 15 heterogeneous datasets and six domains: charts and plots, geometry, graphs, scientific imagery, documents, and 3D spatial scenes, (2) a domain-aware evaluation framework that scores rendered outputs using dataset-specific rubrics and guardrails, showing strong agreement with human annotators and across VLM judges and (3) an evaluator-guided self-training pipeline that uses rendered-output scores to filter model-generated image and code pairs into useful supervision, improving VLM image-to-code performance in the absence of ground truth code.

\section{Related Work}

\paragraph{Code as a Medium for Visual Reasoning.}
There are several works in this domain using code for visual reasoning, although the way it is used varies across different systems. VisProg~\citep{gupta2023visual}, ViperGPT~\citep{suris2023vipergpt}, and CodeVQA~\citep{subramanian2023modular} generate programs that combine perception modules to answer visual questions. Chameleon~\citep{NEURIPS2023_871ed095} and CodeAct~\citep{10.5555/3692070.3694124} use code for planning and tool execution. Another line of work is using code as an intermediate representation of visual information. Here, RECODE~\citep{shen2025recode} derenders images into executable code, CodePlot-CoT~\citep{duan2025codeplot} generates plots as part of mathematical reasoning, while Visual Sketchpad~\citep{hu2024visual} allows VLMs to draw through code. Visual Program Distillation~\citep{Hu_2024_CVPR} studies how much of this programmatic reasoning can be distilled into a standalone VLM. While these approaches differ in how they use code, in most cases it serves as an intermediate step for solving another task. What we study in Vision2Code is the generated code itself, specifically how well the program can reconstruct the source image when executed.

\begin{figure}[!ht]
  \centering
  \definecolor{benchborder}{HTML}{1F3A68}
  \definecolor{benchbg}{HTML}{F4F6FA}
  \begin{tikzpicture}
    \node[inner sep=4pt, outer sep=0pt,
          draw=benchborder, line width=1.2pt,
          rounded corners=4pt,
          fill=benchbg] (benchmark)
      {\includegraphics[width=\dimexpr\linewidth-2pt-8pt\relax]{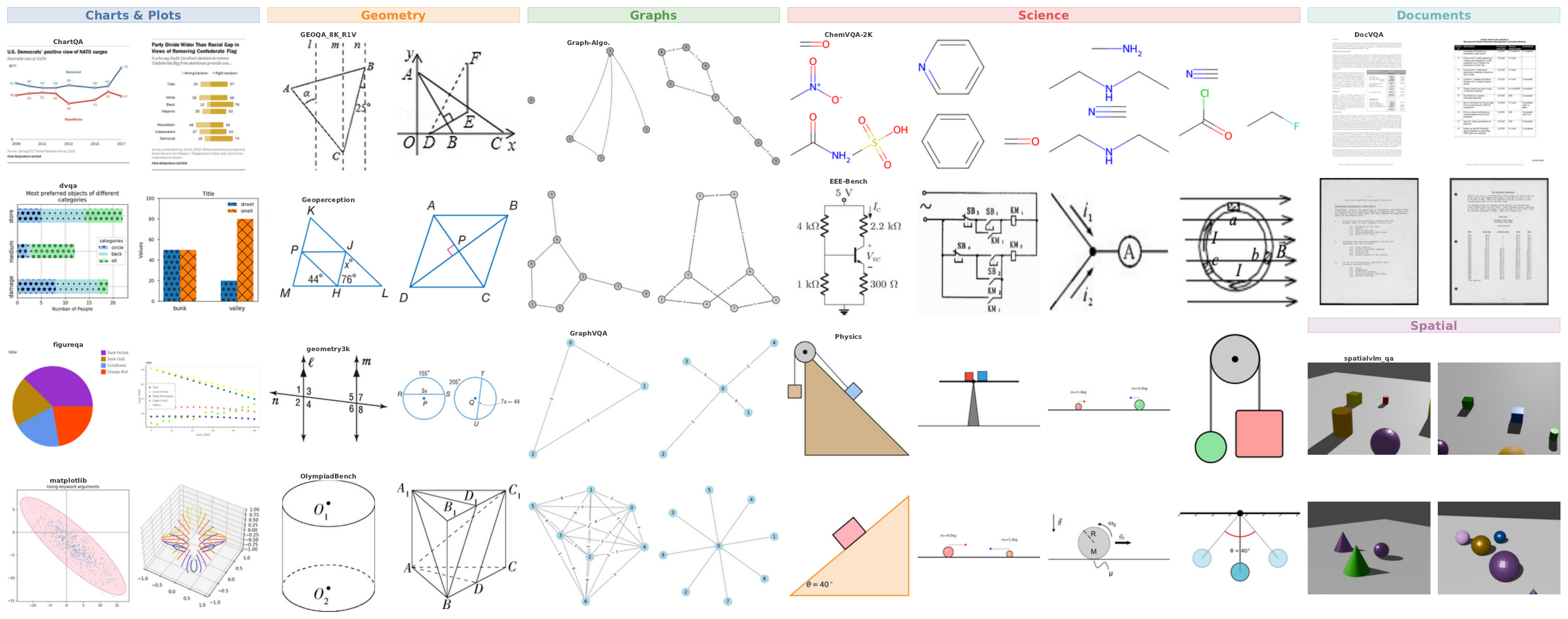}};
  \end{tikzpicture}
  \caption{Vision2Code spans 15 heterogeneous datasets across six domains: charts and plots, geometry, graphs, science, documents, and spatial scenes. The benchmark is reference-code-free and multi-domain.}
  \label{fig:main_overview}
\end{figure}

\paragraph{Image-to-Code Benchmarks.}
Prior image-to-code benchmarks mainly focus on a particular type of output, restricting them to narrow scope. ChartMimic~\citep{yang2025chartmimic}, Plot2Code~\citep{wu2025plot2code}, RealChart2Code~\citep{zhang2026realchart2code}, and Chart2Code~\citep{tang2025chart2code} evaluate charts and scientific plots. Design2Code~\citep{si-etal-2025-design2code}, Web2Code~\citep{yun2024webcode}, WebSight~\citep{laurenccon2024unlocking}, Widget2Code~\citep{zhang2026widget2code}, and Figma2Code~\citep{gui2026figma2code} focus on webpages and UI generation. Image2Struct~\citep{roberts2024image2struct} looks at webpages, LaTeX, and musical scores, while UniSVG~\citep{li2025unisvg}, VCode~\citep{lin2025vcode}, SVGenius~\citep{chen2025svgenius}, VectorGym~\citep{rodriguez2026vectorgym}, and OmniSVG~\citep{yang2025omnisvg} study vector graphics. Concurrent with our work, Omni-I2C~\citep{zhou2026omni} evaluates several types of technical graphics and multiple programming languages, but relies on reference code. Vision2Code does not require such reference programs, which allows us to include datasets where only the image is available. HumanEval-V~\citep{zhang2024humanevalv} and MMCode~\citep{li2024mmcode} also take images as inputs, but the goal there is to solve a programming problem shown in the image rather than reconstruct the image itself.

\paragraph{Evaluation of Image-to-Code Generation.}
A generated program may run successfully and still reconstruct the source poorly, therefore motivating image-level evaluation. Standard image metrics are useful baselines, but the information they measure is different: SSIM, PSNR and MSE are sensitive to pixel-level differences, while CLIP~\citep{radford2021learning} and DreamSim~\citep{fu2023dreamsim} measure broader visual similarity. Neither directly checks the structural details important to every dataset. There are several works in this domain using different combinations of evaluation measures. Plot2Code~\citep{wu2025plot2code} reports code pass rate, text-match ratio and GPT-4V scores. For ChartMimic~\citep{yang2025chartmimic}, both the generated code and rendered chart are evaluated, whereas Design2Code~\citep{si-etal-2025-design2code} uses visual metrics together with human evaluation. Omni-I2C~\citep{zhou2026omni} combines perceptual and symbolic measures. Another line of work is using MLLMs directly as judges~\citep{chen2024mllmjudge}. For Vision2Code, the different categories were semantically motivated based on visual inspection of different failure modes. We use these manually designed, dataset-specific criteria to evaluate reconstruction quality, comparing the scores against human ratings, a second VLM judge and standard image similarity metrics. The VLM evaluator is also tested by modifying generated code to see if it catches the fine-grained details for respective domains.

\begin{table*}[t]
\caption{Comparison with recent image-to-code benchmarks. N is the number of evaluation samples; Multi indicates multiple domains; Ref.-free indicates evaluation without paired source code; Fine eval. indicates category-level or multi-criteria reconstruction evaluation;
Domain eval. indicates evaluation criteria tailored to a visual domain or dataset. Render diag. indicates separate render-failure reporting; Human val. indicates validation against humans. \cmark\ = completely or partially supported, \xmark\ = not supported.
}
\label{tab:benchmark-comparison}
\centering
\scriptsize
\setlength{\tabcolsep}{4pt}
\renewcommand{\arraystretch}{1.08}

\resizebox{\textwidth}{!}{%
\begin{tabular}{llrcccccc}
\toprule
\textbf{Benchmark} &
\textbf{Scope} &
\textbf{Samples} &
\textbf{Multi} &
\textbf{Ref.-free} &
\textbf{Fine eval.} &
\textbf{Domain eval.} &
\textbf{Render diag.} &
\textbf{Human val.} \\
\midrule

ChartMimic
& Charts
& 4{,}800
& \xmark & \cmark & \cmark & \cmark & \cmark & \cmark \\

Plot2Code
& Scientific plots
& 368
& \xmark & \cmark & \xmark & \xmark & \cmark & \cmark \\

RealChart2Code
& Real-world charts
& 2{,}896
& \xmark & \cmark & \cmark & \cmark & \cmark & \cmark \\

Chart2Code
& Charts
& 2{,}186
& \xmark & \cmark & \cmark & \cmark & \cmark & \cmark \\

Design2Code
& Web pages
& 484
& \xmark & \cmark & \cmark & \cmark & \xmark & \cmark \\

Web2Code
& Web pages
& 1{,}198
& \xmark & \cmark & \cmark & \cmark & \xmark & \xmark \\

Widget2Code
& UI widgets
& 1{,}000
& \xmark & \cmark & \cmark & \cmark & \xmark & \cmark \\

Image2Struct
& Web, LaTeX, music
& 2{,}400
& \cmark & \cmark & \xmark & \xmark & \cmark & \xmark \\

Omni-I2C
& Technical graphics
& 1{,}080
& \cmark & \xmark & \cmark & \xmark & \cmark & \cmark \\

\midrule
\textbf{Vision2Code}
& 6 visual domains
& 539
& \cmark & \cmark & \cmark & \cmark & \cmark & \cmark \\
\bottomrule
\end{tabular}%
}
\end{table*}

\section{Benchmark}
\label{sec:benchmark}

\paragraph{Task.} Vision2Code studies image reconstruction by asking a VLM to generate executable code from a given image. Given a source image, a model must generate code whose execution recreates the image as faithfully as possible, but the goal is not to have an exact pixel-level match. We are interested in whether the model is able to preserve the actual information represented in the image, including plotted values, labels, topology, notation, document layout and spatial relations. Models are prompted to generate Python code using Matplotlib and NumPy, and are evaluated based on the rendered output rather than the specific code that produced it.

\paragraph{Domain coverage.}
We include six different domains in Vision2Code, each testing different capabilities. In charts and plots, models need to understand and recover numerical values, axes, scales, labels and the mapping between legends and plotted values. Geometry and graph images depend more on shape, proportions, connectivity and direction. Scientific figures in physics, chemistry and electrical circuits are particularly sensitive to small details where getting these wrong could change the meaning of the image. Documents test text and layout recovering capability, while spatial scenes require the model to capture object identity, depth and relative 3D placement. This gives us a well balanced mix of quantitative, symbolic, text-heavy and spatial images.

\begin{figure}[H]
  \centering
  \includegraphics[width=\textwidth]{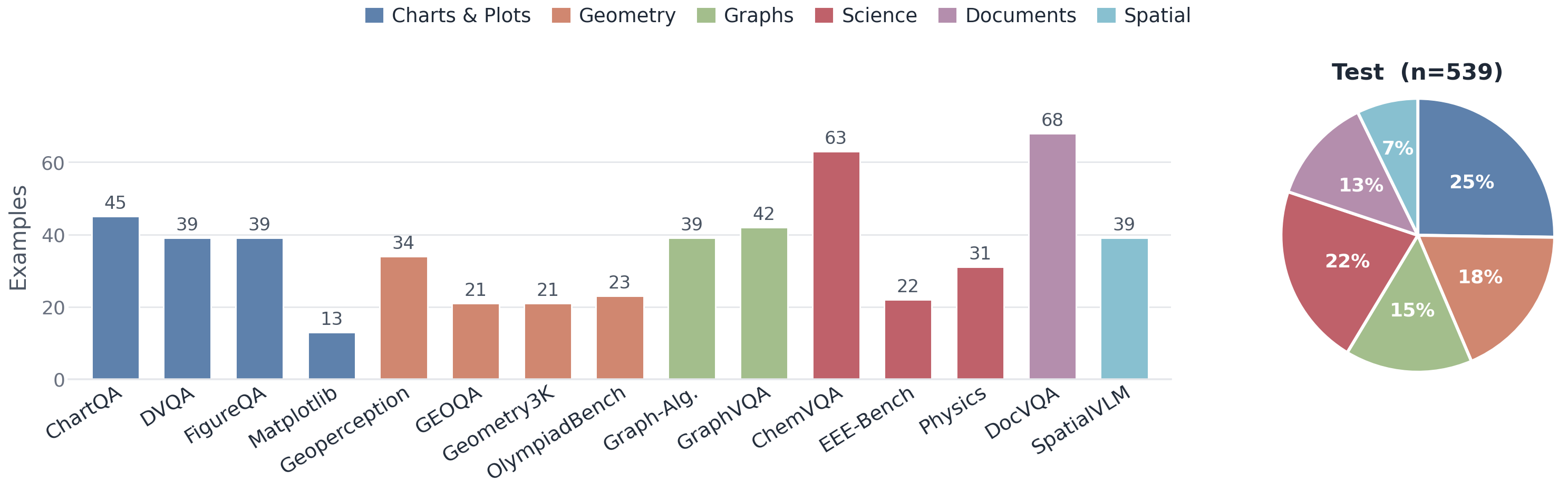}
  \caption{Vision2Code composition. Left: examples per dataset. Right: distribution across six  visual domains.}
  \label{fig:benchmark-statistics}
\end{figure}

\paragraph{Source data and sampling.}
Vision2Code uses 15 public visual datasets covering the six domains mentioned above. We use the official test split of these datasets if available and manually create a held-out set for the remaining. After manual filtering, we end up with 539 examples in the final benchmark. Figure~\ref{fig:benchmark-statistics} shows the distribution of these samples across the different datasets and domains. The source datasets, split construction, sampling procedure and filtering details are provided in Appendix~\ref{app:dataset-datasheet}.

\section{Evaluation Framework}
\label{sec:evaluation}

\begin{figure}[t]
  \centering
  \includegraphics[width=\textwidth]{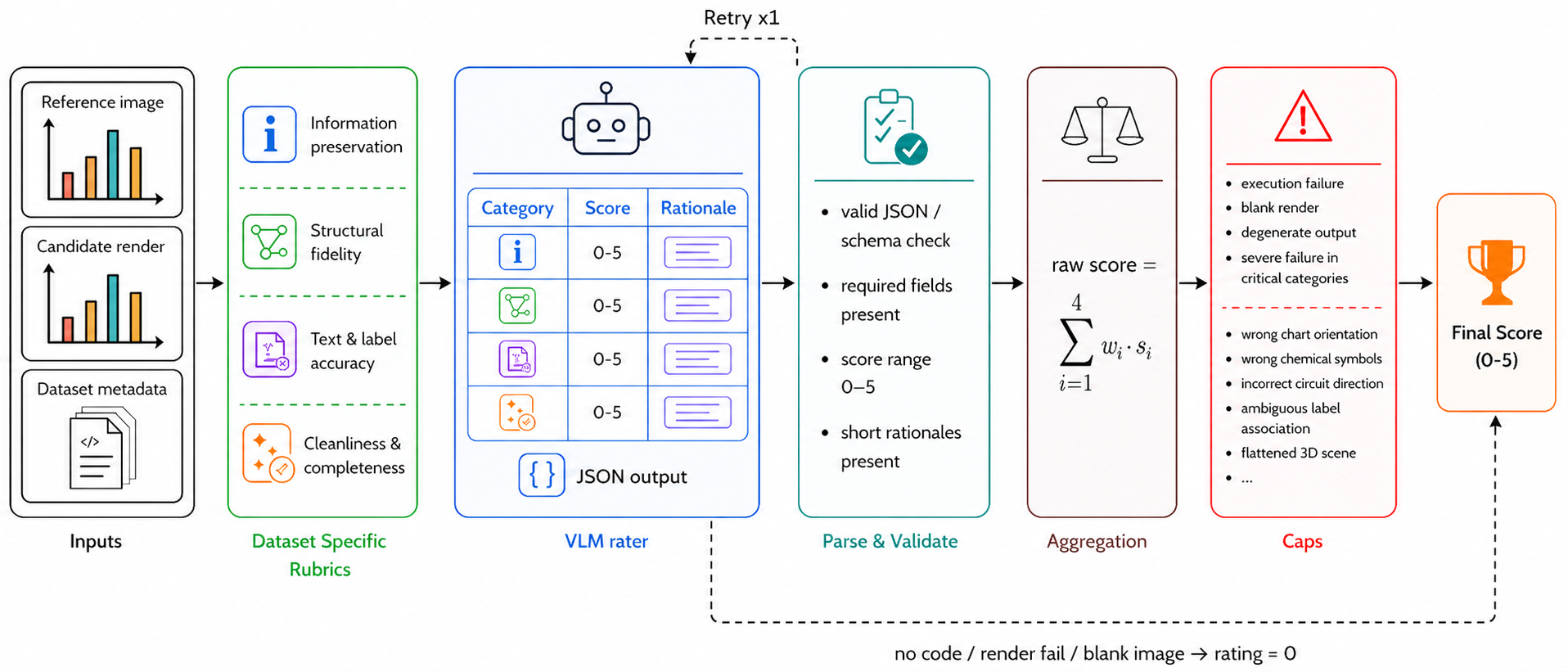}
  \caption{Vision2Code evaluation pipeline. Generated code is rendered into an image, evaluated by comparing against the source image with a dataset-specific rubric consisting of four categories, and aggregated with deterministic caps.}
  \label{fig:evaluation_pipeline}
\end{figure}

\paragraph{Code execution.}
We first execute the code generated by the model. For this, we use Python with Matplotlib and NumPy, since the same runtime can support all six domains. The evaluation itself is not restricted to Python. As long as the generated artifact produces an image, we can evaluate it using the same approach. We also consider other output formats like Excalidraw scene JSON and standalone LaTeX, with examples given in Appendix~\ref{app:tool-use-ablations}. The generated code is run in a fixed environment and we save the rendered image at the same resolution as the source. If the code fails, times out or does not produce a valid image, we directly assign a score of 0. Otherwise, the rendered image is passed to the VLM rater. Figure~\ref{fig:evaluation_pipeline} shows the complete pipeline.

\paragraph{Dataset-specific rubric.}
For every sample, the rater is given the source image, the rendered reconstruction, reference metadata and the rubric corresponding to that dataset. Each rubric consists of four categories, with weights that sum to one. The different categories are scored from 0 to 5 in increments of 0.1, and we calculate the final score as their weighted sum. The first two categories are meant to capture the most important aspects of the image. If either receives a score below a specified threshold, the overall score is capped. We also have additional dataset-specific caps to account for severe errors, since getting certain details wrong could change the actual meaning of the image even if the other parts are reconstructed well. In cases where the rater returns an invalid response or misses a required category, we ask it to evaluate the same pair again. The complete rules and prompts are provided in Appendix~\ref{app:vlm-rubrics}.

\paragraph{Validation protocol.}
We validate our evaluation framework by correlating our scores with human scores and compare these correlations with several other baselines. To test the effect of our dataset-specific rubric, we compare against a dataset invariant rubric with four equally weighted criteria for all samples in the dataset. These include \emph{core information fidelity}, \emph{structure and layout fidelity}, \emph{text and annotation accuracy}, and \emph{visual completeness}. We also include standard similarity methods like SSIM, PSNR, CLIP ViT-B/32, DreamSim, and Qwen3-VL-Embedding-8B. For Qwen3-VL, we test similarity on both the image embedding alone and a version that includes short dataset-specific text describing what should be preserved in the image. We use Pearson and Spearman correlation against the mean score of the four human annotators and also report Inter-Annotator Agreement (IAA) using the Intraclass Correlation Coefficient (ICC). Appendix~\ref{app:human-validation} provides the rating instructions, interface and annotator details.

\section{Experiments}
\label{sec:experiments}

We evaluate nine VLMs on Vision2Code, covering both open-weight (Qwen3.5-9B, Qwen3.5-397B-A17B and Kimi-K2.5, with reasoning variants of the latter two) and proprietary models (GPT-5.4, GPT-5.4 Mini with and without reasoning, and Gemini-3.1-Flash-Lite-Preview). The process is the same for all models: given a source image, we generate code and execute it to obtain the reconstruction, which is then evaluated using the procedure in Section~\ref{sec:evaluation}. For this, we use two independent VLM judges, Qwen3.5-122B-A10B-GPTQ-Int4 and Kimi-K2.6, so every reconstruction is scored separately by both. If the generated code fails to produce a valid image, we assign a score of 0. The overall benchmark score is the macro-average across the 15 datasets.

\subsection{Benchmark Results}

Table~\ref{tab:main-results} reports the per-dataset scores and overall macro-average for all nine models on test. Results from both Qwen3.5-122B and Kimi-K2.6 are included to check whether the ranking changes with the choice of evaluator. Appendix~\ref{app:recreation-comparisons} shows example reconstructions from each model.

\begin{table*}[t]
\caption{Vision2Code leaderboard using two VLM judges: Qwen3.5-122B-A10B-GPTQ-Int4 and Kimi-K2.6. Avg.\ is the macro-average across datasets. Abbreviations: +R = reasoning enabled; MPL = Matplotlib; GeoP = Geoperception; GEOQA = GEOQA\_8K\_R1V; Geom3K = Geometry3K; GAlg = Graph-Algorithms; GVQA = GraphVQA-Swift; Chem = ChemVQA-2K; EEE = EEE-Bench; Olymp = OlympiadBench; Doc = DocVQA; Spat = SpatialVLM-QA.}
\label{tab:main-results}
\centering
\scriptsize
\setlength{\tabcolsep}{1.2pt}
\renewcommand{\arraystretch}{1.12}
VLM Judge: Qwen3.5-122B-A10B-GPTQ-Int4
\vspace{2pt}

\begin{tabular*}{\textwidth}{@{\extracolsep{\fill}}lcccccccccccccccc@{}}
\toprule
Model & \makecell{Chart} & DVQA & FigQA & MPL & GeoP & GEOQA & \makecell{Geom3K} & \makecell{GAlg} & \makecell{GVQA} & \makecell{Chem} & EEE & Phys & Olymp & \makecell{Doc} & Spat & Avg. \\
\midrule
\multicolumn{17}{l}{\textit{Open-weight models}} \\
Qwen3.5-9B & \score{3.10} & \score{3.39} & \score{2.83} & \score{1.91} & \score{2.00} & \score{2.00} & \score{0.98} & \score{0.69} & \score{1.79} & \score{0.07} & \score{0.50} & \score{2.08} & \score{1.28} & \score{1.23} & \score{0.10} & \score{1.60} \\
Kimi-K2.5 & \score{3.07} & \score{4.42} & \score{3.77} & \score{3.63} & \score{3.32} & \score{3.18} & \score{3.48} & \score{2.86} & \score{3.73} & \score{1.73} & \score{2.72} & \score{4.18} & \score{2.34} & \score{2.44} & \score{1.11} & \score{3.06} \\
Kimi-K2.5+R & \score{3.39} & \score{4.57} & \score{4.15} & \score{3.55} & \score{3.61} & \score{3.55} & \score{3.31} & \score{2.59} & \score{4.08} & \score{3.02} & \score{2.87} & \score{4.38} & \score{2.71} & \score{2.92} & \score{0.70} & \score{3.29} \\
Qwen3.5-397B-A17B & \score{3.22} & \score{3.51} & \score{2.93} & \score{2.40} & \score{2.81} & \score{2.81} & \score{2.44} & \score{2.93} & \score{3.61} & \score{0.92} & \score{1.41} & \score{2.63} & \score{2.42} & \score{1.54} & \score{0.54} & \score{2.41} \\
Qwen3.5-397B-A17B+R & \score{3.91} & \score{4.49} & \score{3.84} & \score{2.89} & \score{3.51} & \score{3.52} & \score{2.48} & \score{3.01} & \score{4.03} & \score{1.59} & \score{2.72} & \score{3.42} & \score{3.27} & \score{2.07} & \score{0.27} & \score{3.00} \\
\midrule
\multicolumn{17}{l}{\textit{Proprietary models}} \\
GPT-5.4 & \score{4.60} & \score{4.84} & \score{4.28} & \score{3.71} & \score{4.64} & \score{4.07} & \score{4.43} & \score{3.81} & \score{4.69} & \score{3.83} & \score{4.20} & \score{4.61} & \score{4.64} & \score{4.08} & \score{1.30} & \score{4.12} \\
GPT-5.4 Mini & \score{3.62} & \score{4.70} & \score{4.45} & \score{3.39} & \score{4.50} & \score{3.56} & \score{4.43} & \score{3.89} & \score{4.04} & \score{3.48} & \score{3.56} & \score{4.51} & \score{4.23} & \score{3.36} & \score{1.98} & \score{3.85} \\
GPT-5.4 Mini+R & \score{4.45} & \score{4.70} & \score{4.78} & \score{4.35} & \score{4.56} & \score{3.47} & \score{4.10} & \score{4.12} & \score{4.26} & \score{3.80} & \score{4.00} & \score{4.46} & \score{4.14} & \score{3.90} & \score{2.43} & \score{4.10} \\
Gemini-3.1-Flash-LitePr & \score{4.17} & \score{4.71} & \score{4.21} & \score{3.48} & \score{3.66} & \score{3.41} & \score{3.64} & \score{3.22} & \score{4.29} & \score{0.73} & \score{2.92} & \score{3.81} & \score{3.10} & \score{2.14} & \score{0.05} & \score{3.17} \\
\bottomrule
\end{tabular*}

\vspace{0.7em}
VLM Judge: Kimi-K2.6
\vspace{2pt}

\begin{tabular*}{\textwidth}{@{\extracolsep{\fill}}lcccccccccccccccc@{}}
\toprule
Model & \makecell{Chart} & DVQA & FigQA & MPL & GeoP & GEOQA & \makecell{Geom3K} & \makecell{GAlg} & \makecell{GVQA} & \makecell{Chem} & EEE & Phys & Olymp & \makecell{Doc} & Spat & Avg. \\
\midrule
\multicolumn{17}{l}{\textit{Open-weight models}} \\
Qwen3.5-9B & \score{2.92} & \score{3.14} & \score{2.63} & \score{2.11} & \score{1.68} & \score{2.02} & \score{1.15} & \score{0.72} & \score{1.86} & \score{0.11} & \score{0.82} & \score{2.27} & \score{1.50} & \score{1.22} & \score{0.12} & \score{1.62} \\
Kimi-K2.5 & \score{2.83} & \score{4.23} & \score{3.39} & \score{3.78} & \score{2.88} & \score{2.97} & \score{2.91} & \score{2.77} & \score{3.67} & \score{1.62} & \score{2.86} & \score{3.95} & \score{2.45} & \score{2.21} & \score{1.18} & \score{2.91} \\
Kimi-K2.5+R & \score{3.05} & \score{4.31} & \score{3.58} & \score{3.67} & \score{2.90} & \score{2.89} & \score{2.92} & \score{2.72} & \score{4.06} & \score{2.62} & \score{2.98} & \score{4.02} & \score{2.52} & \score{2.68} & \score{0.92} & \score{3.06} \\
Qwen3.5-397B-A17B & \score{2.62} & \score{3.02} & \score{2.67} & \score{2.59} & \score{2.35} & \score{2.77} & \score{1.92} & \score{2.98} & \score{3.51} & \score{0.76} & \score{1.51} & \score{2.68} & \score{2.52} & \score{1.43} & \score{0.78} & \score{2.27} \\
Qwen3.5-397B-A17B+R & \score{3.76} & \score{4.12} & \score{3.39} & \score{2.99} & \score{2.86} & \score{2.81} & \score{2.13} & \score{2.93} & \score{3.84} & \score{1.36} & \score{2.58} & \score{3.33} & \score{3.03} & \score{1.85} & \score{0.29} & \score{2.75} \\
\midrule
\multicolumn{17}{l}{\textit{Proprietary models}} \\
GPT-5.4 & \score{4.32} & \score{4.74} & \score{3.99} & \score{3.60} & \score{3.84} & \score{3.71} & \score{3.69} & \score{3.82} & \score{4.44} & \score{3.41} & \score{3.76} & \score{4.55} & \score{4.23} & \score{3.84} & \score{1.46} & \score{3.83} \\
GPT-5.4 Mini & \score{3.43} & \score{4.44} & \score{4.03} & \score{3.38} & \score{3.81} & \score{3.20} & \score{3.70} & \score{3.73} & \score{3.96} & \score{3.15} & \score{3.16} & \score{4.45} & \score{3.99} & \score{3.12} & \score{1.97} & \score{3.57} \\
GPT-5.4 Mini+R & \score{4.37} & \score{4.47} & \score{4.36} & \score{4.09} & \score{3.89} & \score{3.35} & \score{3.69} & \score{4.01} & \score{3.98} & \score{3.51} & \score{3.38} & \score{4.11} & \score{4.03} & \score{3.66} & \score{2.25} & \score{3.81} \\
Gemini-3.1-Flash-LitePr & \score{3.96} & \score{4.54} & \score{3.85} & \score{3.37} & \score{3.18} & \score{2.97} & \score{3.08} & \score{3.13} & \score{4.19} & \score{0.88} & \score{2.69} & \score{3.45} & \score{2.97} & \score{1.96} & \score{0.09} & \score{2.95} \\
\bottomrule
\end{tabular*}
\end{table*}

\paragraph{Overall performance.}
GPT-5.4 achieves the highest score on Vision2Code, followed closely by GPT-5.4 Mini with reasoning. Among the open-weight models, Kimi-K2.5 with reasoning performs the best, whereas Qwen3.5-9B is considerably weaker. This shows the performance gap between the frontier proprietary and open-weight models. At the same time, even the highest-scoring models are still far from achieving full reconstruction accuracy. Even with this progress, reference-free image-to-code generation remains far from solved.

\paragraph{Domain variation.}
Models can reconstruct standard charts, plots and graphs well but struggle on chemistry, circuits, document layouts and 3D scenes. One possible explanation for this is that code for different mathematical plots and visualizations frequently appear in pretraining data. The other domains require more precise symbolic, geometric or spatial reasoning. SpatialVLM-QA is the hardest dataset by a large margin, where even the best score is only 2.43. Performance patterns are observed across domains for both VLM judges. We discuss the different failure modes qualitatively in Section~\ref{subsec:reconstruction-failures}.

\paragraph{Reasoning effects.}
While test-time reasoning helps improve the score for all three models under both judges, the gains vary across domains. Some datasets have lower scores with reasoning, and several of these decreases appear under both judges. This is particularly clear on SpatialVLM-QA for Kimi-K2.5 and Qwen3.5-397B-A17B. We also observe decreases on Matplotlib and Graph-Algorithms for Kimi-K2.5, and on Geometry3K and Physics for GPT-5.4 Mini. This could happen because reasoning may help the model decompose the image and plan the overall code structure but not with the extraction of exact visual details.

\subsection{Evaluator Robustness}

\begin{table}[t]
\centering
\caption{Evaluator validation.
(a) Human alignment and (b) Rubric sensitivity, measured by
Pearson $r$ and Spearman $\rho$.
(c) Judge Repeatability, original score (Orig.), 3-run mean, and variance
(Var., in units of $10^{-3}$).}
\label{tab:evaluator-validation}

\scriptsize
\setlength{\tabcolsep}{4pt}
\renewcommand{\arraystretch}{1.08}

\begin{minipage}[t]{0.25\linewidth}
\vspace{0pt}
\begin{tabular*}{\linewidth}{@{\extracolsep{\fill}}lrr@{}}
\multicolumn{3}{@{}l}{\textbf{(a) Human alignment}} \\
\midrule
\textbf{Metric} & $r$ & $\rho$ \\
\midrule
SSIM                & -0.054 & -0.023 \\
PSNR                &  0.057 &  0.092 \\
CLIP                &  0.443 &  0.460 \\
Cosine (image)      &  0.627 &  0.716 \\
Cosine (+text)      &  0.576 &  0.663 \\
DreamSim            &  0.686 &  0.726 \\
Generic (Qwen)      &  0.785 &  0.817 \\
Generic (Kimi)      &  0.807 &  0.820 \\
\midrule
\textbf{Ours (Qwen)} & \textbf{0.836} & \textbf{0.826} \\
\textbf{Ours (Kimi)} & \textbf{0.849} & \textbf{0.861} \\
\bottomrule
\end{tabular*}
\end{minipage}\hfill
\begin{minipage}[t]{0.20\linewidth}
\vspace{0pt}
\begin{tabular*}{\linewidth}{@{\extracolsep{\fill}}lrr@{}}
\multicolumn{3}{@{}l}{\textbf{(b) Rubric sensitivity}} \\
\midrule
\textbf{Variant} & $r$ & $\rho$ \\
\midrule
\multicolumn{3}{@{}l}{\textit{VLM Judge: Qwen3.5-122B}} \\
Uniform          & 0.835 & 0.823 \\
Uncapped         & 0.817 & 0.794 \\
\textbf{Complete} & \textbf{0.836} & \textbf{0.826} \\
\midrule
\multicolumn{3}{@{}l}{\textit{VLM Judge: Kimi-K2.6}} \\
Uniform          & 0.844 & 0.860 \\
Uncapped         & 0.832 & 0.836 \\
\textbf{Complete} & \textbf{0.849} & \textbf{0.861} \\
\bottomrule
\end{tabular*}
\end{minipage}\hfill
\begin{minipage}[t]{0.52\linewidth}
\vspace{0pt}
\begin{tabular*}{\linewidth}{@{\extracolsep{\fill}}lrrrrrr@{}}
\multicolumn{7}{@{}l}{\textbf{(c) Judge repeatability}} \\
\midrule
\textbf{Model} &
\multicolumn{3}{c}{\textbf{Qwen judge}} &
\multicolumn{3}{c}{\textbf{Kimi judge}} \\
\cmidrule(lr){2-4}\cmidrule(lr){5-7}
& Orig. & Mean & Var. & Orig. & Mean & Var. \\
\midrule
Qwen3.5-9B
& 2.086 & 2.077 & 1.052 & 1.981 & 1.979 & 0.012 \\
Kimi-K2.5
& 3.413 & 3.421 & 0.290 & 3.110 & 3.097 & 0.144 \\
Kimi-K2.5+R
& 3.625 & 3.638 & 0.973 & 3.340 & 3.360 & 0.636 \\
Qwen3.5-397B
& 2.555 & 2.627 & 4.053 & 2.545 & 2.539 & 0.068 \\
Qwen3.5-397B+R
& 3.635 & 3.615 & 0.844 & 3.212 & 3.211 & 0.506 \\
GPT-5.4
& 4.133 & 4.124 & 0.223 & 3.750 & 3.726 & 0.795 \\
GPT-5.4 Mini
& 3.950 & 3.967 & 0.306 & 3.770 & 3.760 & 0.300 \\
GPT-5.4 Mini+R
& 4.310 & 4.276 & 1.812 & 3.823 & 3.787 & 1.012 \\
Gemini Flash-Lite
& 3.348 & 3.341 & 0.181 & 3.022 & 3.041 & 0.337 \\
\midrule
\textbf{All pairs}
& \textbf{3.453} & \textbf{3.456} & \textbf{0.016}
& \textbf{3.175} & \textbf{3.169} & \textbf{0.029} \\
\bottomrule
\end{tabular*}
\end{minipage}
\end{table}

\paragraph{Human alignment.}
We validate the evaluator by checking correlations with human ratings on a balanced subset of 180 source-output pairs from the test set, covering six models and all 15 datasets. Each pair was scored independently by four annotators on a 0--5 scale, giving 720 ratings in total. We use their mean as the human reference score. Agreement is high, with ICC(2,1) = 0.753 for a single annotator and ICC(2,4) = 0.924 for the mean of four. The comparison with perceptual, pixel-level, embedding and generic-rubric baselines is shown in Table~\ref{tab:evaluator-validation}(a). Under both VLM judges, the dataset-specific rubric has the highest correlation with human ratings: Pearson/Spearman correlations are 0.849/0.861 for Kimi-K2.6 and 0.836/0.826 for Qwen3.5-122B. Both judges therefore agree more closely with humans when using the dataset-specific rubric than their corresponding generic-rubric version. DreamSim performs best among the non-VLM methods, while CLIP, embedding similarity, SSIM and PSNR have lower agreement. Additional details are given in Appendix~\ref{app:human-validation}.

\paragraph{Judge agreement.}
Another aspect is whether the leaderboard depends on the VLM used for evaluation. For this, we rescore all the images using Kimi-K2.6 and compare the results with our primary judge, Qwen3.5-122B (Table~\ref{tab:main-results}). While Kimi-K2.6 generally assigns lower scores, the ranking of all nine models remains exactly the same. The model-level scores have a Pearson correlation of 0.999, and across the 4{,}845 outputs evaluated by both judges, the correlation is 0.936. We also wanted to check if there is any model-family preference, since Kimi-K2.6 is evaluating models from the same family as well. However, the Kimi-K2.5 models do not improve in rank or receive unusually favorable changes in their scores compared with the other models. Kimi-K2.6 also has slightly better agreement with human ratings than Qwen3.5-122B for both the generic and dataset-specific rubrics (Table~\ref{tab:evaluator-validation}(a)).

\paragraph{Rubric sensitivity.}
The different categories and their weights were chosen based on the information that needs to be preserved in each dataset. But the exact weights may not necessarily be optimal, so we wanted to understand how much the results depend on the chosen weights and the additional caps. We ran an ablation with two settings: a \textit{uniform} version where all four categories have equal weights but the original caps and guardrails are retained, and an \textit{uncapped} version where we keep the original weights but remove the caps. We then compare these settings by checking correlations with human ratings. As shown in Table~\ref{tab:evaluator-validation}(b), the complete rubric has the highest Pearson and Spearman correlations under both judges. Perturbing the weights through uniform weighting changes the correlations only slightly, whereas removing the caps causes a more noticeable drop. This suggests that there is some flexibility in choosing the weights, but the caps are important since a severe error in one category should not be compensated by high scores in the other categories.

\paragraph{Judge repeatability.}
We also study how much the evaluator scores change when the same images are rated multiple times. For each judge, we evaluate the same 539 source-output pairs three times, with the pairs distributed as evenly as possible across all nine models. This includes the original run and two reruns. Table~\ref{tab:evaluator-validation}(c) shows that the three-run means are close to the original scores, with very little variance between runs. Comparing the two reruns, the individual-output Pearson/Spearman correlations are 0.985/0.983 for Qwen3.5-122B and 0.983/0.974 for Kimi-K2.6. The mean absolute score differences are 0.089 and 0.157 respectively, and 95.0\% and 93.3\% of the scores differ by at most 0.5 points. Both judges achieve a single-run agreement ICC of 0.981 across the three runs. Kimi preserves the same ranking of all nine models in every run, while Qwen has one swap between the fourth and fifth ranked models.

\paragraph{Domain-specific error sensitivity.}
Another thing we studied was VLM evaluator robustness by modifying generated code to see if it catches the fine-grained details for respective domains. For this experiment, we selected 71 reconstructions from 14 datasets across all nine evaluated models, where every reconstruction had received a score of at least 4.5 from both judges. We then modified the corresponding programs to introduce one controlled error targeting a particular rubric category, while keeping the remaining content unchanged. The idea is to see if the score of that category decreases when an error is introduced. We find that the targeted score drops in 65/71 cases for Qwen3.5-122B and 68/71 for Kimi-K2.6, with mean decreases of 2.36 and 2.02 points on the 0--5 scale. The final score also decreases in 66/71 cases for Qwen and all 71 cases for Kimi, and neither judge assigns a higher score to any of the modified images. Both judges struggle the most with geometry, missing errors such as a shortened trapezoid segment, a broken parallelogram connection or a removed right-angle label. Qwen also misses an absent chart bar, a changed chemical bond and a circuit gap, while Kimi misses a modified graph edge weight. Qualitative examples are provided in Appendix~\ref{app:evaluator-robustness-examples}.

\subsection{Failure Analysis}

We separate failures into two types. Some generated programs fail to run, time out, or produce no valid image. Others execute successfully but still reconstruct the source poorly. We analyze these execution and reconstruction failures separately.

\subsubsection{Execution Failures}

The render success rates and counts of the different failure modes are reported in Table~\ref{tab:render-error-analysis}. Success ranges from 75.1\% for Qwen3.5-9B to 98.9\% for GPT-5.4 Mini with reasoning. A high success rate, however, does not necessarily mean that the reconstructions are good. For Qwen3.5-397B-A17B, enabling reasoning gives a higher benchmark score but fewer successful renders. GPT-5.4 and GPT-5.4 Mini with reasoning both render almost all examples, and at the same time there is still a difference in how well they reconstruct the source images. The failure types also differ across models. Syntax errors are the most frequent reason for failure in Qwen3.5-9B, whereas Gemini-Flash-Lite-Preview mostly fails through invalid or hallucinated API calls. For GPT-5.4 Mini without reasoning, the main failures are runtime errors or cases where no output is produced. Its reasoning variant has very few failures. Reasoning can also change the type of error rather than simply reduce the total number, as we see with Qwen3.5-397B-A17B: Shape/3D errors become less common, but API-related failures increase and the overall render success rate drops.

\begin{table*}[t]
\caption{Execution failure analysis on test. Render success is reported as a percentage, while failure columns report counts among failed renders. API = hallucinated APIs or invalid arguments; Shape/3D = shape mismatch or 3D geometry errors; Other = runtime errors, no-output exits, or timeouts; Syntax = syntax errors or truncated code; MissDep = missing dependencies or nonexistent files.}
\label{tab:render-error-analysis}
\centering
\scriptsize
\setlength{\tabcolsep}{4pt}
\renewcommand{\arraystretch}{1.08}

\begin{tabular*}{\textwidth}{@{\extracolsep{\fill}}lcrrrrrr@{}}
\toprule
\textbf{Model} &
\textbf{Render success (\%)} &
\multicolumn{6}{c}{\textbf{Counts by failure type}} \\
\cmidrule(lr){3-8}
& & Total & API & Shape/3D & Other & Syntax & MissDep \\
\midrule

Qwen3.5-9B
& 75.1 & 134 & 28 & 37 & 22 & 45 & 2 \\

Kimi-K2.5
& 88.5 & 62 & 21 & 8 & 3 & 13 & 17 \\

Kimi-K2.5+R
& 90.9 & 49 & 5 & 11 & 28 & 4 & 1 \\

Qwen3.5-397B-A17B
& 87.9 & 65 & 16 & 22 & 12 & 14 & 1 \\

Qwen3.5-397B-A17B+R
& 81.6 & 99 & 40 & 12 & 27 & 14 & 6 \\

GPT-5.4
& 97.6 & 13 & 4 & 5 & 3 & 1 & 0 \\

GPT-5.4 Mini
& 94.1 & 32 & 7 & 6 & 18 & 0 & 1 \\

GPT-5.4 Mini+R
& 98.9 & 6 & 1 & 5 & 0 & 0 & 0 \\

Gemini-3.1-Flash-LitePr
& 92.2 & 42 & 37 & 1 & 2 & 2 & 0 \\

\midrule
All models
& 89.7 & 502 & 159 & 107 & 115 & 93 & 28 \\
\bottomrule
\end{tabular*}
\end{table*}

\subsubsection{Reconstruction Failures}
\label{subsec:reconstruction-failures}

For charts and plots, we observe flipped axes, plot trends that are not preserved, and overlapping legends or labels. Geometry reconstructions often have incorrect proportions, missing components, wrong shapes and overlapping text. In graphs, the errors include reversed arrow directions, incorrect overall topology and wrong node or edge labels. Chemistry examples frequently have incorrect molecular shapes, missing double or triple bonds, wrong atom identities and heavy overlap. Object placement is a problem in physics examples, where objects may overlap slopes or appear to float. Circuit reconstructions often have missing components, malformed component shapes or disconnected paths. While document reconstructions may capture the rough content, they can still lose the formatting, relative proportions and layout. Spatial scenes are particularly difficult, with models often failing to preserve depth and relative object placement. Qualitative examples are provided in Appendix~\ref{app:reconstruction-failures}.

\subsection{Evaluator-Guided Self-Training}
\label{sec:self-training}

\begin{figure*}[t]
\centering
\includegraphics[width=\textwidth]{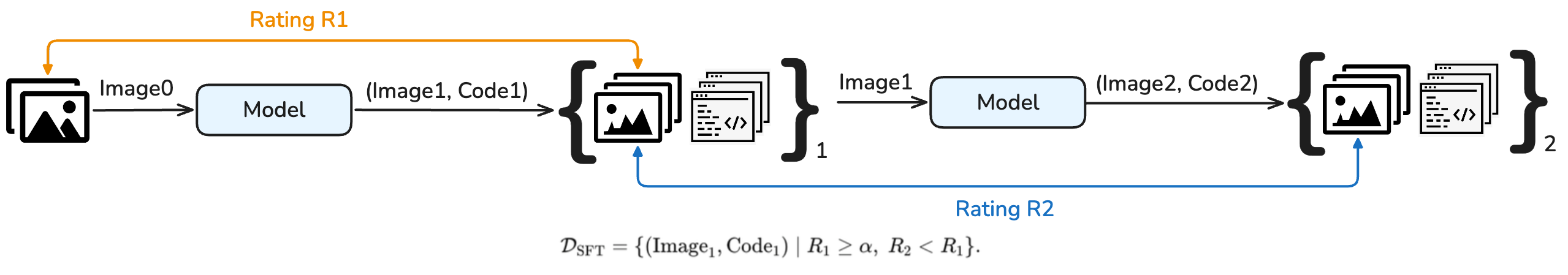}
\caption{Evaluator-guided self-training pipeline. First-stage image-code pairs are used for SFT when the render scores well against the source ($R_1 \geq \alpha$) but is not stably reconstructed in the second stage ($R_2 < R_1$).}
\label{fig:self-training-ablation}
\end{figure*}

Since most visual datasets do not have the source code to generate the images, direct image-to-code supervision is difficult. For training data, we instead investigate whether the model's own successful generations can be used for supervised fine-tuning. Figure~\ref{fig:self-training-ablation} shows the pipeline. Given a source image $I_0$, the model generates code $C_1$ which renders into $I_1$. We then give $I_1$ back to the same model to generate another program $C_2$, producing a second rendered image $I_2$. The evaluator assigns two scores, $R_1 = R(I_0,I_1)$ for the first reconstruction and $R_2 = R(I_1,I_2)$ for the reconstruction of the model's own output. Since we have the scores from both reconstructions, we can use them to decide which of the first-stage $(I_1,C_1)$ pairs should be used for training. We are particularly interested in cases where the model has successfully reconstructed an image once, but still struggles to reproduce the same image when given it again. Our primary filter, $R_1\ge\alpha,\; R_2<R_1$, identifies such examples by requiring the initial reconstruction to have a high score while the second reconstruction has a lower score. This allows us to focus on images that the model can generate but cannot yet reconstruct consistently. We compare this with four other filtering strategies in Table~\ref{tab:self-training-ablation}, and find that this particular filter gives the best results. Using these filtered examples for fine-tuning improves Qwen3.5-9B from 1.60 to 1.86 and Phi-4-MM from 0.58 to 0.74. We also evaluate the base and fine-tuned models using Kimi-K2.6 and Qwen3.6-Plus since Qwen3.5-122B is used for both selecting the training examples and evaluating the final models and may introduce bias. The improvements are significant and hold under both judges. The other filtering strategies are less effective due to the quality of the filtered examples. Full dataset counts and hyperparameter details are provided in Appendix~\ref{app:self-training-details}. 

\begin{table}[h]
\centering
\caption{Evaluator-guided self-training. The ablation reports
test scores under the primary Qwen3.5-122B evaluator. The final three columns
report the best-filter improvement over baseline under the primary judge
and two additional evaluators, Kimi-K2.6 and Qwen3.6-Plus. Here,
$\alpha=4$.}
\label{tab:self-training-ablation}
\scriptsize
\setlength{\tabcolsep}{2.5pt}
\renewcommand{\arraystretch}{1.12}

\begin{tabular*}{\linewidth}{
    @{\extracolsep{\fill}}lccccccccc@{}}
\toprule
&
\multicolumn{6}{c}{\textbf{Performance of different evaluator-guided filters on test}} &
\multicolumn{3}{c}{\textbf{Best-filter improvement over baseline}} \\
\cmidrule(lr){2-7}\cmidrule(l){8-10}

\textbf{Model} &
\textbf{Baseline} &
\textbf{All Valid} &
\makecell{$R_1\ge\alpha$\\$R_2\ge R_1$} &
$R_1<\alpha$ &
$R_1\ge\alpha$ &
\makecell{$R_1\ge\alpha$\\$R_2<R_1$} &
\makecell{\textbf{Qwen3.5-122B}\\$\boldsymbol{\Delta}$} &
\makecell{\textbf{Kimi-K2.6}\\$\boldsymbol{\Delta}$} &
\makecell{\textbf{Qwen3.6-Plus}\\$\boldsymbol{\Delta}$} \\
\midrule

Qwen3.5-9B
& 1.5970
& 1.7233
& 1.7722
& 1.6962
& 1.8567
& \textbf{1.8645}
& \textbf{+0.2675}
& \textbf{+0.2113}
& \textbf{+0.2118} \\

Phi-4-MM
& 0.5756
& 0.4097
& 0.5450
& 0.6349
& 0.6299
& \textbf{0.7414}
& \textbf{+0.1658}
& \textbf{+0.1655}
& \textbf{+0.1895} \\

\bottomrule
\end{tabular*}
\end{table}

\section{Conclusion}
\label{sec:conclusion}

\textbf{Vision2Code} is a comprehensive, multi-domain, reference-code-free benchmark for evaluating image-to-code generation capability in VLMs. Our experiments show that our dataset-specific evaluator agrees more closely with human ratings than several other metrics and is robust to various ablations. The benchmark reveals that current models are much stronger on charts and graphs than on spatial scenes, scientific diagrams, and documents, which are signals for directions of improvement for frontier models. We also demonstrated that using our evaluator to curate model-generated outputs is a viable strategy for self-training, yielding clear performance boosts for both Qwen3.5-9B and Phi-4-MM. That being said, several limitations remain. The benchmark's scope is restricted to its source datasets and can be extended to many more domains. The leaderboard scores are based on Python generated code (specifically Matplotlib and NumPy). Exploring the impact of the intermediate code representation could be an interesting future direction. Additionally, while robust, our VLM evaluator can still miss subtle errors that humans can catch. Future work can expand the benchmark to include more datasets, programming languages, and evaluation methods to further advance image-to-code generation capabilities.

\bibliography{references}
\bibliographystyle{unsrtnat}

\appendix

\section{VLM Evaluation Rubrics}
\label{app:vlm-rubrics}

We specify the rubric-based VLM evaluation protocol used for the reported benchmark scores. The same rubric prompt is used across raters and the default rater is Qwen3.5-122B-A10B-GPTQ-Int4. The evaluator receives the reference image, the rendered candidate image, dataset metadata, and a dataset-specific rubric. It returns only category-level scores and short rationales; the final score is computed deterministically outside the model using the rubric weights and additional guardrails.

\subsection{Rater Prompt}

\begin{promptbox}{System prompt}
You are a rigorous but calibrated evaluator for image recreation quality. Compare the reference image against the candidate rendered image. Judge only what is visibly present in the images.

Use the dataset-specific rubric and instructions provided by the user message. Be strict about semantically wrong recreations, but do not over-penalize differences that the rubric explicitly says to treat as secondary.

Scoring anchors on a 0.0 to 5.0 scale: 5.0 = near-exact or clearly excellent for that dataset; 4.0 = strong recreation with minor issues; 3.0 = good but with noticeable structural, text, or semantic problems; 2.0 = partial recreation with important errors; 1.0 = major mismatch; 0.0 = missing, unreadable, blank, or fundamentally broken.

Return JSON only. Do not include markdown fences.
\end{promptbox}

For each sample, the user message presents the reference image, the candidate rendered image, and the following instruction template:
\begin{promptbox}{User prompt template}
Evaluate the candidate recreation against the reference image using the dataset-specific rubric below. The prompt includes the rubric version, resolved dataset name, dataset-specific guidance, dataset-specific notes, category definitions with weights, and reference metadata.

Return JSON with exactly these top-level keys: \texttt{category\_scores}, \texttt{rationales}, \texttt{strengths}, \texttt{issues}, and \texttt{overall\_summary}. The \texttt{category\_scores} object must contain every required category id with a numeric score from 0.0 to 5.0, using 0.1 increments when needed. The \texttt{rationales} object must contain the same category ids, with one short rationale per category. \texttt{strengths} should contain 1--3 short bullets, \texttt{issues} should contain 1--5 short bullets, and \texttt{overall\_summary} should contain 1--3 concise sentences.

Judge what matters for the dataset-specific rubric, not superficial similarity alone. For easy datasets, preserve recreation structure, not just semantic content. For hard or noisy references, be strict but do not penalize differences that the source itself makes ambiguous. Use 0.0 or 1.0 for severe failures. Do not compute the final 0.0 to 5.0 rating yourself; provide category scores only.
\end{promptbox}

If the rater output is not valid JSON or is missing required category ids, we issue a repair prompt asking for corrected JSON only with the same top-level keys and the exact required category ids. The repair prompt does not change the images or rubric.

\subsection{Aggregation and Guardrails}

Each rubric contains four weighted categories whose weights sum to 1.0. The rater assigns a score $s_i\in[0,5]$ to each category, rounded to one decimal place. We compute the provisional visual rating as
\[
  r_{\mathrm{raw}} = \sum_i w_i s_i.
\]
The final score is $r_{\mathrm{raw}}$ after deterministic caps. If code execution fails or no valid render is produced, the final score is 0.0. If the candidate image is unreadable, missing, blank, near-monochrome, or otherwise degenerate, the score is capped at 0.5. The first two rubric categories are treated as critical semantic categories; low scores on these categories trigger the profile-dependent caps in Table~\ref{tab:cap-profiles}. Dataset-specific caps in Table~\ref{tab:dataset-rubrics} further prevent high final scores when a key semantic failure is present.

\begin{table}[H]
\caption{Cap profiles used by the rubric aggregator. A cap limits the final score to at most the listed value when its condition is met.}
\label{tab:cap-profiles}
\centering
\small
\setlength{\tabcolsep}{4pt}
\renewcommand{\arraystretch}{1.12}
\begin{tabularx}{\textwidth}{@{}>{\raggedright\arraybackslash}p{0.29\textwidth}XXX@{}}
\toprule
\textbf{Condition} & \textbf{Strict} & \textbf{Balanced} & \textbf{Hard} \\
\midrule
Any critical category $\leq$ threshold & $\leq 1.5 \Rightarrow 2.5$ & $\leq 1.5 \Rightarrow 2.8$ & $\leq 1.2 \Rightarrow 3.0$ \\
Both critical categories $\leq$ threshold & $\leq 2.5 \Rightarrow 3.5$ & $\leq 2.5 \Rightarrow 4.0$ & $\leq 2.2 \Rightarrow 4.0$ \\
Near-perfect guardrail & $r_{\mathrm{raw}}\geq4.8$ and critical $<4.6 \Rightarrow 4.5$ & $r_{\mathrm{raw}}\geq4.8$ and critical $<4.4 \Rightarrow 4.5$ & $r_{\mathrm{raw}}\geq4.8$ and critical $<4.2 \Rightarrow 4.5$ \\
\bottomrule
\end{tabularx}
\end{table}

\subsection{Dataset-Specific Rubrics}

Table~\ref{tab:dataset-rubrics} gives the complete dataset-specific rubrics. The table uses human-readable category names; the evaluator prompt uses corresponding machine-readable category ids with the same weights and order. The category order is meaningful: the first two categories are the critical categories used by the guardrails above.

\begingroup
\scriptsize
\setlength{\tabcolsep}{2.5pt}
\renewcommand{\arraystretch}{1.14}

\begin{longtable}{@{}>{\raggedright\arraybackslash}p{0.16\textwidth}>{\raggedright\arraybackslash}p{0.30\textwidth}>{\raggedright\arraybackslash}p{0.47\textwidth}@{}}
\caption{Dataset-specific rubrics. Category weights are shown in parentheses. Caps are applied after weighted aggregation.}
\label{tab:dataset-rubrics}\\
\toprule
\textbf{Dataset} & \textbf{Categories and weights} & \textbf{Guidance and dataset-specific caps} \\
\midrule
\endfirsthead

\caption[]{Dataset-specific VLM rating rubrics (continued).}\\
\toprule
\textbf{Dataset} & \textbf{Categories and weights} & \textbf{Guidance and dataset-specific caps} \\
\midrule
\endhead

\midrule
\multicolumn{3}{r}{\textit{Continued on next page}}\\
\endfoot

\bottomrule
\endlastfoot

ChartQA\newline\textit{Strict} &
Numeric information (.30)\newline
Structure and layout (.30)\newline
Text, labels, and legibility (.25)\newline
Cleanliness and completeness (.15) &
Easy chart recreation; preserve information and chart structure. Penalize wrong orientation/type, crowded labels, illegible text, bad title placement, or legend overlap. Cap at 3.5 for orientation/type mismatch when numeric information is mostly preserved but structure is weak; cap at 2.5 for material numeric or text errors. \\
\addlinespace

DVQA\newline\textit{Strict} &
Numeric information (.30)\newline
Structure and layout (.30)\newline
Text, labels, and legibility (.25)\newline
Cleanliness and completeness (.15) &
Easy chart recreation; preserve exact plotted information, faithful structure, and readable labels. Cap at 3.5 for orientation/type mismatch when numeric information is mostly preserved but structure is weak; cap at 2.5 for material numeric or text errors. \\
\addlinespace

FigureQA\newline\textit{Strict} &
Numeric information (.30)\newline
Structure and layout (.30)\newline
Text, labels, and legibility (.25)\newline
Cleanliness and completeness (.15) &
Easy chart recreation; semantic preservation alone is insufficient for top scores. Penalize meaningful plot-structure changes and crowded or illegible labels. Cap at 3.5 for orientation/type mismatch when numeric information is mostly preserved but structure is weak; cap at 2.5 for material numeric or text errors. \\
\addlinespace

Matplotlib\newline\textit{Strict} &
Layout and composition (.30)\newline
Plot elements (.25)\newline
Text, ticks, and annotations (.25)\newline
Style and rendering (.20) &
Coding-heavy recreation with ground-truth plotting code. High scores require genuinely close layout, plot elements, ticks, annotations, and rendering. Unknown library-specific style differences matter less than missing structure. Cap at 3.0 for wrong layout or missing major plot elements. \\
\addlinespace

Geoperception\newline\textit{Strict} &
Geometric structure (.35)\newline
Numbers, labels, and markers (.30)\newline
Association and placement (.20)\newline
Readability and completeness (.15) &
Geometry-heavy figures; structure, labels, markers, and clear annotation binding are central. Small label shifts are acceptable only when the target remains unambiguous. Cap at 2.5 for wrong geometric relations or ambiguous label association. \\
\addlinespace

GEOQA\_8K\newline R1V\newline\textit{Strict} &
Geometric structure (.35)\newline
Numbers, labels, and markers (.30)\newline
Association and placement (.20)\newline
Readability and completeness (.15) &
Geometry-heavy figures; precise structure, labels, markers, numbers, and annotation association matter. Slight label shifts are acceptable if association remains unambiguous. Cap at 2.5 for wrong geometric relations or ambiguous label association. \\
\addlinespace

Geometry3K\newline\textit{Strict} &
Geometric structure (.35)\newline
Numbers, labels, and markers (.30)\newline
Association and placement (.20)\newline
Readability and completeness (.15) &
Geometry problem diagrams; approximate correctness is not sufficient for high scores. Angle or length text may move slightly only if association remains obvious. Cap at 2.5 for wrong geometric relations or ambiguous label association. \\
\addlinespace

Graph-Algorithms\newline\textit{Strict} &
Graph structure (.40)\newline
Ordering and direction semantics (.25)\newline
Labels and numbers (.20)\newline
Readability and completeness (.15) &
Graph topology, directions, ordering, weights, and labels are central. Cosmetic layout differences matter less than exact structure. If the source is cluttered, avoid over-penalizing ambiguity already present. Cap at 2.5 for wrong topology, ordering, or direction semantics. \\
\addlinespace

GraphVQA-Swift\newline\textit{Strict} &
Graph structure (.35)\newline
Labels and numbers (.30)\newline
Attribute mapping (.20)\newline
Readability and completeness (.15) &
Graph structure, label precision, and attribute-to-node/edge binding are central. Wrong label ordering or wrong attribute attachment is important. Cap at 2.5 for wrong label ordering or attribute binding. \\
\addlinespace

ChemVQA-2K\newline\textit{Balanced} &
Molecular structure (.40)\newline
Chemical symbols (.30)\newline
Bond lines and markers (.20)\newline
Readability and completeness (.10) &
Focus on chemistry semantics: molecular structure, chemical symbols, bond lines, and semantically meaningful colored difference markers. Small orientation changes matter less than structure and symbols. Cap at 2.5 for wrong molecular structure or wrong chemical symbols. \\
\addlinespace

EEE-Bench\newline\textit{Balanced} &
Circuit topology and connectivity (.35)\newline
Component identity and symbols (.25)\newline
Labels, values, and directions (.25)\newline
Readability and completeness (.15) &
Circuit topology, component identity, labels, values, signs, and current/voltage direction matter most. Be lenient on cosmetic symbol style when identity is clear. Cap at 2.5 for wrong topology, labels/values, or current/voltage direction. \\
\addlinespace

Physics\newline\textit{Balanced} &
Physical setup (.35)\newline
Vectors, labels, and numbers (.30)\newline
Object geometry consistency (.20)\newline
Readability and completeness (.15) &
Focus on physical setup, object placement, vectors, labels, and numerical givens. Wrong force direction or physically inconsistent placement should be penalized strongly. Cap at 2.5 for wrong vector direction or physically inconsistent placement. \\
\addlinespace

OlympiadBench\newline\textit{Balanced} &
Problem structure and semantics (.35)\newline
Labels, numbers, and symbols (.25)\newline
Spatial precision (.20)\newline
Readability and completeness (.20) &
Harder and denser problem figures. Judge whether the recreated figure supports the same downstream reasoning problem. Do not over-penalize small cosmetic differences that do not change meaning. Cap at 2.5 for wrong core problem semantics. \\
\addlinespace

DocVQA\newline\textit{Balanced} &
Text information (.35)\newline
Document structure and layout (.30)\newline
Legibility and non-crowding (.20)\newline
Key visual elements (.15) &
Judge text and structure first. Be strict but fair for dense or difficult infographics; exact decorative styling matters less than information and structure. Cap at 2.5 for materially wrong key text or numbers. \\
\addlinespace

SpatialVLM-QA\newline\textit{Hard} &
Object shape and identity (.25)\newline
Spatial relations (.35)\newline
3D depth rendering (.25)\newline
Color and scene completeness (.15) &
Hard 3D spatial reconstruction. Judge object identity, spatial relations, depth, and scene completion carefully. Correct objects with a flattened 2D recreation should not score highly. Cap at 2.5 for flattened 2D scenes that preserve objects but lack depth; cap at 2.0 for wrong core spatial relations. \\
\end{longtable}
\endgroup

\section{Benchmark Construction}
\label{app:dataset-datasheet}

\paragraph{Data sources.}
Vision2Code is built from 15 visual source datasets. Each example is converted to a common schema containing a source image, source metadata, source provenance, dataset identity, and a stable sample identifier. The sources are ChartQA~\citep{masry2022chartqa}, DVQA~\citep{kafle2018dvqa}, FigureQA~\citep{kahou2018figureqa}, Matplotlib Gallery examples~\citep{matplotlib_gallery,hunter2007matplotlib}, Geoperception~\citep{zhang2024euclid}, GEOQA\_8K\_R1V from the GeoQA-8K direct-synthesizing release associated with GeoQA and MM-UPT~\citep{chen2021geoqa,wei2025mmupt}, Geometry3K from Inter-GPS~\citep{lu2021intergps}, Graph-Algorithms and Physics from Zebra-CoT~\citep{li2026zebracot}, GraphVQA-Swift~\citep{graphvqa_swift_hf}, ChemVQA-2K~\citep{bhuma2025chemvqa}, EEE-Bench~\citep{li2025eeebench}, OlympiadBench~\citep{he2024olympiadbench}, DocVQA~\citep{mathew2021docvqa}, and SpatialVLM-QA~\citep{litian2025spatialvlmqa,chen2024spatialvlm}. All sources except Matplotlib are obtained from public Hugging Face releases; the Matplotlib examples are scraped from the official gallery, retaining only successfully rendered figures. For release, we document the exact source URI, license or access status, split policy, and redistribution status for each dataset in the accompanying license and provenance table.

\paragraph{Train/test splits.} Native train/test splits are preserved when available (ChartQA, ChemVQA-2K, DocVQA, GEOQA\_8K\_R1V, GraphVQA-Swift, Geometry3K). The remaining sources are deterministically re-split with seed 42 to produce held-out pools. We sample
600 candidate test examples from these held-out pools. Each dataset receives a minimum quota to preserve small sources, with the remaining budget allocated proportionally to held-out pool size via the largest-remainder method. Training pools are not used for leaderboard evaluation and are reserved for self-training experiments.

\paragraph{Filtering.} All candidate test examples are manually reviewed for suitability as reconstruction targets. A reviewer sees only the source image and accepts or rejects the example; examples are rejected if they are blank, corrupted, dominated by unreadable fine print, dependent on hidden metadata, or otherwise infeasible to reconstruct from visible structure alone. Rejected examples are not replaced. Of the 600 candidate examples, 539 are retained as the final Vision2Code test set.

\begin{table*}[t]
\centering
\caption{Per-dataset benchmark counts and split policy after manual filtering.}
\label{tab:appendix-dataset-datasheet}
\scriptsize
\setlength{\tabcolsep}{4pt}
\renewcommand{\arraystretch}{1.10}
\begin{tabular}{@{}llrl@{}}
\toprule
\textbf{Dataset} & \textbf{Domain} & \textbf{Test} & \textbf{Split policy} \\
\midrule
ChartQA          & Charts \& Plots      & 45 & Native split \\
DVQA             & Charts \& Plots      & 39 & Deterministic, seed 42 \\
FigureQA         & Charts \& Plots      & 39 & Deterministic, seed 42 \\
Matplotlib       & Charts \& Plots      & 13 & Deterministic, seed 42 \\
Geoperception    & Geometry             & 34 & Deterministic, seed 42 \\
GEOQA\_8K\_R1V   & Geometry             & 21 & Native split \\
Geometry3K       & Geometry             & 21 & Native split \\
OlympiadBench    & Geometry             & 23 & Deterministic, seed 42 \\
Graph-Algorithms & Graphs               & 39 & Deterministic, seed 42 \\
GraphVQA-Swift   & Graphs               & 42 & Native split \\
ChemVQA-2K       & Scientific Imagery   & 63 & Native split \\
EEE-Bench        & Scientific Imagery   & 22 & Deterministic, seed 42 \\
Physics          & Scientific Imagery   & 31 & Deterministic, seed 42 \\
DocVQA           & Documents            & 68 & Native split \\
SpatialVLM-QA    & 3D Spatial Scenes    & 39 & Deterministic, seed 42 \\
\midrule
\textbf{Total}   & --                   & \textbf{539} & -- \\
\bottomrule
\end{tabular}
\end{table*}

\section{Generation and Rendering Protocol}
\label{app:generation-rendering}

\paragraph{Generation prompt.}
All benchmarked models receive the same image-to-code instruction. For local Qwen models the prompt is applied through the model chat template with \texttt{enable\_thinking=False}; local decoding is greedy (\texttt{do\_sample=False}). API-hosted models use the same system and user content, with provider-specific message formatting only.

\begin{promptbox}{Code-generation system prompt}
You are an expert Python plotting assistant. Given an input image, write Python code to recreate it using matplotlib/numpy as needed. Save the image only to the Python variable OUTPUT\_PATH. Do not quote OUTPUT\_PATH and do not append a file extension to it. Return code only.
\end{promptbox}

\begin{promptbox}{Code-generation user prompt}
Generate Python code to recreate this image. Return code only.
\end{promptbox}

\paragraph{Code normalization.}
Model outputs are normalized before execution. We strip leading and trailing whitespace, remove \texttt{<think>} blocks when present, and extract the final Python code block when the model returns markdown fences. The normalized string is written as \texttt{generated\_code.py}. No semantic repair is applied before the primary benchmark render.

\paragraph{Rendering environment.}
Generated code is executed in a temporary directory with \texttt{MPLBACKEND=Agg}. The evaluator defines \texttt{OUTPUT\_PATH} and monkey-patches Matplotlib \texttt{savefig} calls so that the output is saved to the expected path at the same pixel dimensions as the source image. The renderer sets the figure size from the source width and height with target DPI 100, enforces \texttt{bbox\_inches=None} and \texttt{pad\_inches=0}, and times out executions after 30 seconds unless overridden by a run configuration. If code execution fails, exits without producing an image, or produces an unreadable image, the sample is counted as a render failure and receives benchmark score 0.0.

\paragraph{Software stack.}
The main local rendering environment used Python 3.10.19, Matplotlib 3.10.6, NumPy 2.2.6, Pillow 12.1.0, and PyTorch 2.10.0+cu128. Cluster launchers set Hugging Face, Torch, Triton, Matplotlib, and package caches to writable scratch directories to avoid shared-cache lock contention.

\section{Generic Rubric and Embedding Baselines}
\label{app:generic-rubric-embedding}

\subsection{Generic Rubric Baseline}
The generic rubric baseline uses the same VLM rater, image inputs, output schema, and render-failure rule as the dataset-specific evaluator. The only difference is the rubric: instead of using a source-dataset-specific rubric, every example is scored with the same four equal-weight categories as shown in Table \ref{tab:appendix-generic-rubric}. This baseline tests whether the improvement in human alignment comes from dataset-specific evaluation criteria rather than from the rater model alone. The evaluation prompts are shown below.

\begin{promptbox}{Generic-rubric system prompt}
You are a rigorous but calibrated evaluator for image recreation quality. Compare the reference image against the candidate rendered image. Judge only what is visibly present in the images.

Use the rubric provided by the user message. It applies to every dataset. Be strict about semantically wrong recreations, but do not over-focus on small cosmetic differences.

Scoring anchors on a 0.0 to 5.0 scale: 5.0 = near-exact or clearly excellent; 4.0 = strong recreation with minor issues; 3.0 = good but with noticeable structural, text, or semantic problems; 2.0 = partial recreation with important errors; 1.0 = major mismatch; 0.0 = missing, unreadable, blank, or fundamentally broken.

Return JSON only. Do not include markdown fences.
\end{promptbox}

\begin{promptbox}{Generic-rubric user prompt template}
Evaluate the candidate recreation against the reference image using this rubric.

Rubric version: \texttt{generic\_recreation\_rubric\_v1\_equal\_weight}

Rubric dataset: \texttt{generic}

Return JSON with exactly these top-level keys:
\texttt{category\_scores}, \texttt{rationales}, \texttt{strengths}, \texttt{issues}, and \texttt{overall\_summary}.

The \texttt{category\_scores} object must contain every category id with a numeric score from 0.0 to 5.0. The \texttt{rationales} object must contain the same category ids, with one short rationale per category. Use the same four generic categories for every dataset. Do not compute the final 0.0 to 5.0 rating yourself; provide category scores only.
\end{promptbox}

\begin{table}[H]
\caption{Rubric categories used for the generic-rubric evaluator.}
\label{tab:appendix-generic-rubric}
\centering
\small
\setlength{\tabcolsep}{5pt}
\renewcommand{\arraystretch}{1.12}
\begin{tabularx}{\textwidth}{@{}l c X@{}}
\toprule
\textbf{Category} & \textbf{Weight} & \textbf{Meaning} \\
\midrule
Core information fidelity & 0.25 & Whether the rendered image preserves the main objects, values, symbols, and visual facts. \\
Structure and layout fidelity & 0.25 & Whether spatial arrangement, object relationships, axes, topology, or document organization are preserved. \\
Text and annotation accuracy & 0.25 & Whether labels, numbers, legends, symbols, and annotations are correct and legible. \\
Visual completeness and cleanliness & 0.25 & Whether the render is complete, non-degenerate, readable, and free from severe artifacts. \\
\bottomrule
\end{tabularx}
\end{table}

For the generic baseline, the final score is computed outside the rater as the weighted average of the four category scores. The same render-failure rule is used as in the main evaluator: if the generated code fails to execute or does not produce a valid image, the final score is set to 0.0. Unlike the dataset-specific evaluator, the generic baseline does not use source-dataset-specific critical-category caps or domain-specific guardrails.

\subsection{Embedding Similarity baseline}

We also compare against embedding-similarity baselines using Qwen3-VL-Embedding-8B. These baselines do not use a VLM rater or rubric. Instead, they embed the source image and rendered output and compute cosine similarity between the two embeddings. We evaluate two variants. The \emph{image-only} variant embeds each image directly. The \emph{image+text} variant embeds each image together with a short dataset-specific focus text that describes the visual information that should matter for that source dataset. The same focus text is used for the source image and rendered output of a given example. Appendix~\ref{app:embedding-focus-texts} lists the focus texts. For both embedding variants, the common embedding instruction is: ``Represent the input for comparing the visual fidelity of a recreated figure to its source.''

\subsubsection{Embedding Focus Texts}
\label{app:embedding-focus-texts}

For the image+text cosine-similarity validation metric, each source-render pair is embedded with the same dataset-specific focus text. These short prompts specify the visual attributes that should dominate the embedding comparison for each dataset.

\begin{table}[H]
\caption{Dataset-specific focus texts used for the image+text cosine-similarity validation metric.}
\label{tab:embedding-focus-texts}
\centering
\small
\setlength{\tabcolsep}{6pt}
\renewcommand{\arraystretch}{1.25}
\begin{tabularx}{\textwidth}{@{}>{\bfseries\raggedright\arraybackslash}p{0.22\textwidth} X@{}}
\toprule
\normalfont\textbf{Dataset} & \textbf{Focus text} \\
\midrule
ChartQA           & chart type, plotted values, axis labels, legend entries, overall layout. \\
DVQA              & chart structure, plotted values, readable labels, legend entries, information content. \\
FigureQA          & chart structure, plotted trends, labels, legend entries, visual relationships. \\
Matplotlib        & plot layout, data traces, axes, titles, legends, annotations, figure styling. \\
Geoperception     & geometry structure, labels, markers, relationships, annotation placement. \\
GEOQA\_8K\_R1V    & geometry structure, point labels, lengths, angles, markers, annotation placement. \\
Geometry3K        & geometry shapes, point labels, markers, numerical annotations, exact relations. \\
Graph-Algorithms  & graph topology, node labels, edge directions, edge weights, ordering. \\
GraphVQA-Swift    & graph structure, node labels, edge labels, attributes, label binding. \\
ChemVQA-2K        & molecular structure, chemical symbols, bond lines, annotations, color-coded semantics. \\
EEE-Bench         & circuit topology, component identity, labels, values, symbols, current or voltage direction. \\
Physics           & object placement, vectors, labels, numerical givens, physically correct structure. \\
OlympiadBench     & geometry structure, labels, givens, markers, precise relationships. \\
DocVQA            & text accuracy, document layout, table structure, labels, visual organization. \\
SpatialVLM-QA     & object identity, spatial relations, depth cues, layout, scene completeness. \\
\bottomrule
\end{tabularx}
\end{table}

\section{Human Validation}
\label{app:human-validation}

\begin{figure}[H]
\centering
\includegraphics[width=0.8\linewidth]{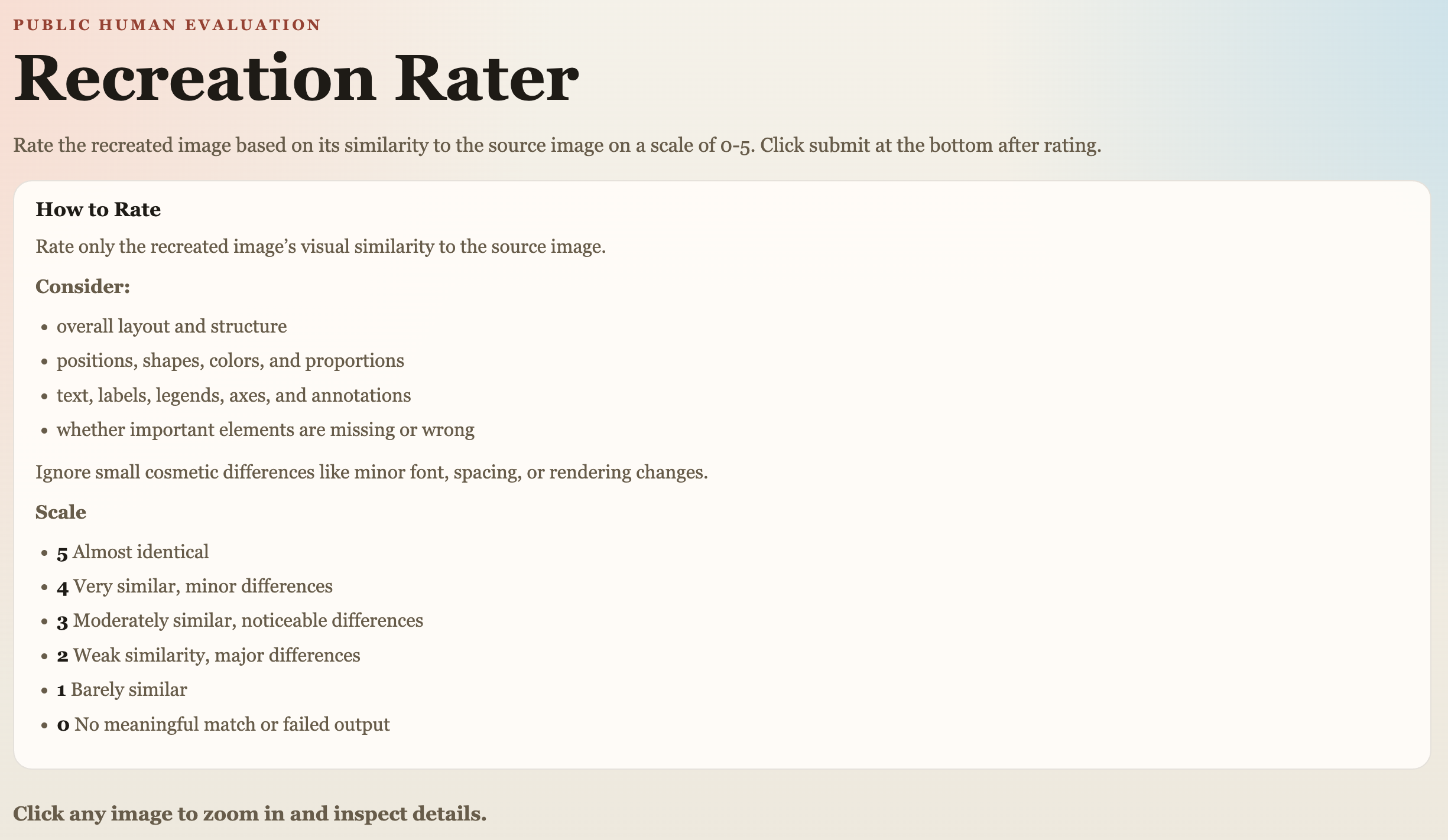}
\caption{Rater web interface: general rating instructions shown to human annotators.}
\label{fig:appendix-human-rater-1}
\end{figure}

\begin{figure}[H]
\centering
\includegraphics[width=0.8\linewidth]{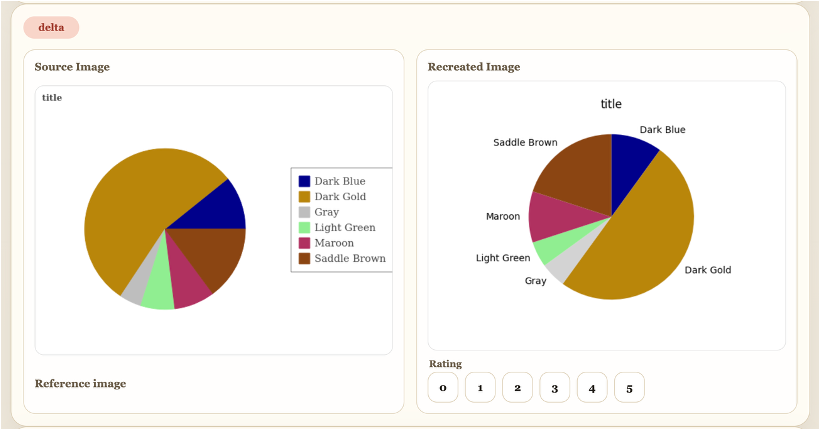}
\caption{Rater web interface: source image (left) and candidate render (right) with the 0--5 rating controls.}
\label{fig:appendix-human-rater-2}
\end{figure}

We evaluate human alignment on 180 source--output pairs from test. The set is balanced across six models and all 15 datasets, with 30 outputs per model, and 12 per dataset. The six models are Qwen3.5-9B, Kimi-K2.5, Qwen3.5-397B-A17B, GPT-5.4, GPT-5.4 Mini, and Gemini-3.1-Flash-LitePreview. Four annotators independently rate all 180 pairs, yielding 720 ratings. Model and dataset identities are hidden, and examples are shown in a different random order for each annotator. Annotators compare the source and rendered image and rate reconstruction quality from 0--5, considering structure, layout, shapes, positions, text, labels, and missing or incorrect elements while ignoring minor cosmetic differences. The scale ranges from 5 (almost identical) to 0 (no meaningful match). Figures
\ref{fig:appendix-human-rater-1} and \ref{fig:appendix-human-rater-2} show the
annotation interface.

\subsection{Inter-Annotator Agreement}
\label{app:human-iaa}

Because all four annotators rate the same examples, we use a two-way random-effects, absolute-agreement ICC. ICC(2,1) measures the reliability of one annotator, while ICC(2,4) measures the reliability of the four-rater mean used in our correlation analysis. Confidence intervals are computed by bootstrapping the 180 examples. Pairwise Pearson correlations range from 0.754 to 0.879, Spearman correlations from 0.760 to 0.869, and quadratic-weighted Cohen's $\kappa$ from 0.619 to 0.863. The high ICC(2,4) supports using the mean of the four ratings as the human target.

\begin{table}[H]
\centering
\scriptsize
\setlength{\tabcolsep}{6pt}
\renewcommand{\arraystretch}{1.08}
\caption{Inter-annotator agreement across four human annotators.}
\label{tab:appendix-human-iaa}
\begin{tabular}{@{}lcc@{}}
\toprule
\textbf{Measure} & \textbf{Result} & \textbf{95\% CI} \\
\midrule
ICC(2,1), single annotator & 0.753 & [0.701, 0.795] \\
ICC(2,4), four-rater mean & 0.924 & [0.904, 0.939] \\
\midrule
All scores within 1 point & 63.3\% & -- \\
All scores within 2 points & 91.7\% & -- \\
\bottomrule
\end{tabular}
\end{table}

\subsection{Correlation Analysis}

We compute Pearson and Spearman correlations between each automatic metric and the mean of the four human ratings over the same 180 examples. All metrics are available for all examples, so no values are removed or imputed. We compare SSIM, PSNR, CLIP, DreamSim, Qwen3-VL embedding similarity with and without dataset-specific focus text, and both generic and dataset-specific rubrics evaluated with Qwen3.5-122B and Kimi-K2.6. As reported in Table~\ref{tab:evaluator-validation}(a), the dataset-specific rubric achieves the highest correlation with human judgments under both VLM judges. It also consistently improves over the corresponding generic rubric, showing that the gain is not only due to using a VLM judge but also to the dataset-specific evaluation criteria.

\section{Evaluator-Guided Self-Training Details}
\label{app:self-training-details}

For each self-training source image $I_0$, the model first generates code $C_1$, whose render is $I_1$. The model is then given $I_1$ and generates code again, producing a second render $I_2$. The evaluator assigns $R_1$ to $(I_0,I_1)$ and $R_2$ to $(I_1,I_2)$. Thus, $R_1$ measures fidelity to the source, while $R_2$ measures whether the model can consistently re-encode its own rendered output. We fine-tune on $(I_1,C_1)$ pairs: high-$R_1$ samples provide code that reconstructs the source accurately, while $R_2<R_1$ identifies visual structures that the model produces once but cannot reconstruct consistently.

We generate separate sets of 20{,}000 trajectories for Qwen3.5-9B and Phi-4 Multimodal Multimodal using the same source-dataset training cohort, which is disjoint from the 539-example Vision2Code test set. For Qwen3.5-9B, 12{,}973 trajectories have valid paired ratings; 5{,}215 are missing a paired rating and 1{,}812 have a non-ok or missing rating status. For Phi-4 Multimodal, 9{,}199 trajectories have valid paired ratings; 6{,}761 are missing a paired rating and 4{,}040 have a non-ok or missing rating status.

We set $\alpha=4$ and construct five filtered datasets for each model. To compare filters using equal training-set sizes within each model, we match all variants to the size of the smallest filter. This gives 1{,}412 examples per Qwen variant, split into 1{,}271 training and 141 development examples, and 122 examples per Phi-4 Multimodal variant, split into 110 training and 12 development examples. Table~\ref{tab:appendix-self-training-filters} reports the available candidates and the resulting dataset sizes.

\begin{table}[H]
\caption{Self-training dataset filters. Variants are sampled with seed 42 from valid stage-paired trajectories and matched in size within each model.}
\label{tab:appendix-self-training-filters}
\centering
\small
\setlength{\tabcolsep}{6pt}
\renewcommand{\arraystretch}{1.12}
\begin{tabular}{@{}l l r r r@{}}
\toprule
\textbf{Variant} & \textbf{Filter} & \textbf{Candidates} & \textbf{Train} & \textbf{Dev} \\
\midrule
\multicolumn{5}{@{}l}{\textit{Qwen3.5-9B}} \\
All valid rated samples & Any valid paired $(R_1,R_2)$ & 12{,}973 & 1{,}271 & 141 \\
Self-consistent high & $R_1 \ge \alpha,\;R_2 \ge R_1$ & 1{,}808 & 1{,}271 & 141 \\
Self-drop high & $R_1 \ge \alpha,\;R_2 < R_1$ & 1{,}412 & 1{,}271 & 141 \\
Low first rating & $R_1 < \alpha$ & 9{,}753 & 1{,}271 & 141 \\
All high first rating & $R_1 \ge \alpha$ & 3{,}220 & 1{,}271 & 141 \\
\midrule
\multicolumn{5}{@{}l}{\textit{Phi-4 Multimodal Multimodal}} \\
All valid rated samples & Any valid paired $(R_1,R_2)$ & 9{,}199 & 110 & 12 \\
Self-consistent high & $R_1 \ge \alpha,\;R_2 \ge R_1$ & 541 & 110 & 12 \\
Self-drop high & $R_1 \ge \alpha,\;R_2 < R_1$ & 122 & 110 & 12 \\
Low first rating & $R_1 < \alpha$ & 8{,}536 & 110 & 12 \\
All high first rating & $R_1 \ge \alpha$ & 663 & 110 & 12 \\
\bottomrule
\end{tabular}
\end{table}

\paragraph{Fine-tuning configuration.}
Each variant fine-tunes its corresponding base checkpoint using full fine-tuning with DeepSpeed ZeRO-3 on four A5000 GPUs. Both Qwen3.5-9B and Phi-4 Multimodal use a maximum sequence length of 2{,}048, 6 epochs, a learning rate of $5\times10^{-6}$, per-device training and evaluation batch sizes of 1, evaluation and checkpointing every 50 steps, and seed 42. Qwen uses gradient accumulation 8 and logs every 10 steps. Because each Phi-4 Multimodal variant contains only 110 training examples, Phi-4 Multimodal uses gradient accumulation 1 and logs every 5 steps. For each variant, we select the checkpoint with the lowest development loss.

\begin{figure}[H]
\centering
\includegraphics[width=\linewidth]{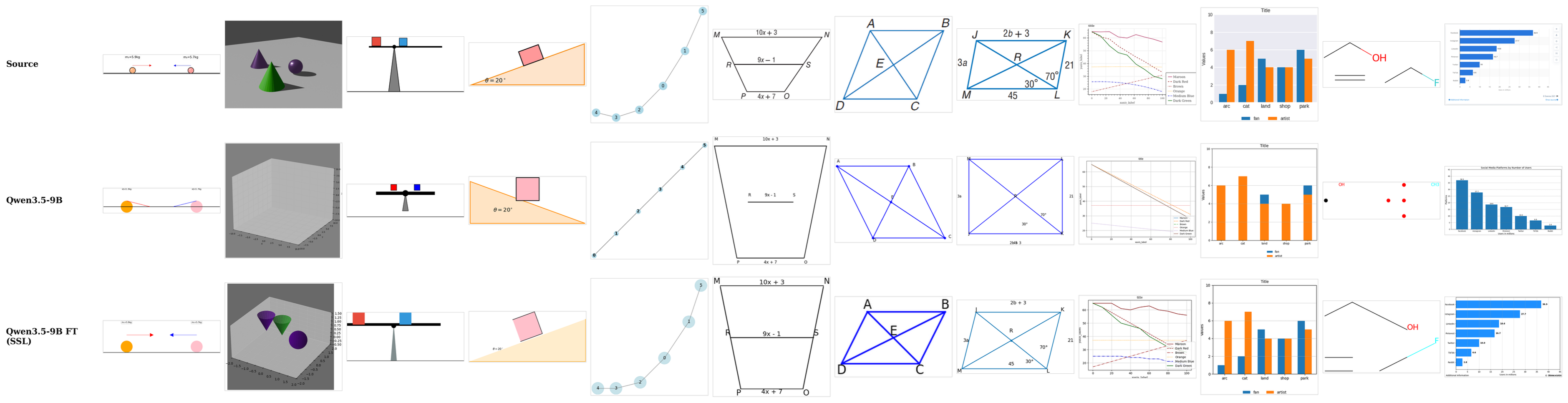}
\caption{Qualitative examples before and after rater-filtered self-training. Rows show source images, off-the-shelf Qwen3.5-9B renders, and renders after fine-tuning on the $(R_1 \geq \alpha,\ R_2 < R_1)$ subset.}
\label{fig:appendix-self-training-examples}
\end{figure}

Figure~\ref{fig:appendix-self-training-examples} provides a qualitative check on the self-training result. The fine-tuned model often improves visible structure rather than only changing style: examples show more complete layouts, better object placement, and fewer collapsed or missing elements compared with the base Qwen3.5-9B outputs.

\paragraph{Independent evaluation.}
Training-data selection uses Qwen3.5-122B as the evaluator. To check whether
the measured improvement is specific to this judge, we separately rescore the
baseline and best-filter models with Kimi-K2.6 and Qwen3.6-Plus. Neither
additional judge is used for selecting the self-training data.

\section{Failure Taxonomy}
\label{app:failure-taxonomy}

Render failures are separated from semantic reconstruction errors. Render failures receive score 0.0 because no valid candidate image is available for the rubric evaluator. We classify failures using the execution status and traceback in Table \ref{tab:appendix-failure-taxonomy}. 

\begin{table}[H]
\caption{Execution failure taxonomy used in render-success diagnostics.}
\label{tab:appendix-failure-taxonomy}
\centering
\small
\setlength{\tabcolsep}{5pt}
\renewcommand{\arraystretch}{1.12}
\begin{tabularx}{\textwidth}{@{}l X X@{}}
\toprule
\textbf{Failure type} & \textbf{Definition} & \textbf{Typical cause} \\
\midrule
Syntax/truncation & Code cannot be parsed or stops mid-expression. & Output length limit, unterminated strings, unmatched brackets. \\
Missing dependency or file & Code imports unavailable packages or reads absent local files. & Assumed external assets, unavailable plotting libraries. \\
No image saved & Code exits without producing the expected output file. & Missing save call, saving to another path, interactive-only display. \\
Hallucinated API & Code calls nonexistent methods or unsupported keyword arguments. & Invented Matplotlib helpers, invalid patch/axes methods. \\
Shape/3D error & Runtime failure from incompatible array shapes or invalid 3D geometry calls. & Incorrect coordinate dimensionality, wrong patch type for 3D axes. \\
Other runtime error & Any execution failure not covered above. & Type errors, value errors, or unhandled corner cases. \\
\bottomrule
\end{tabularx}
\end{table}

\section{Additional Analyses}
\label{app:additional-analyses}

\paragraph{Test-time scaling.}

\newsavebox{\testtimescalingtbl}
\begin{lrbox}{\testtimescalingtbl}
\footnotesize
\setlength{\tabcolsep}{4pt}
\renewcommand{\arraystretch}{1.10}
\begin{tabular}{@{}ccc@{}}
\toprule
\textbf{Stage} & \textbf{Qwen3.5-9B} & \textbf{GPT-5.4 Mini} \\
\midrule
1 & 1.60 & 3.85 \\
2 & 1.63 & 3.84 \\
3 & 1.61 & 3.83 \\
4 & 1.63 & 3.81 \\
\bottomrule
\end{tabular}
\end{lrbox}

\begin{wraptable}{r}{\dimexpr\wd\testtimescalingtbl+2\fboxsep\relax}
\vspace{-\baselineskip}
\centering
\captionsetup{
  width=\dimexpr\wd\testtimescalingtbl+2\fboxsep\relax,
  justification=raggedright,
  singlelinecheck=false
}
\caption{Test-time scaling on test across four inference stages.}
\label{tab:test-time-scaling}
\usebox{\testtimescalingtbl}
\vspace{-0.75em}
\end{wraptable}

We also test whether repeated refinement over four stages improves reconstruction at inference time. At stage 1, the model receives only the source image. At later stages, it receives the source image together with its previous code and rendered output, and is asked to improve the reconstruction. For Qwen3.5-9B, the first refinement gives a small gain, but later stages do not improve further. For GPT-5.4 Mini, repeated refinement slightly reduces performance. One possible reason is that after the first pass, the main challenge is no longer generating code but identifying exactly what is wrong with the rendered image. Without a separate verifier, the model has to judge its own mistakes and may keep modifying an already incorrect structure.

Stage 1 uses the standard image-to-code prompt. For stages 2--4, the model receives the original reference image, the previous rendered output, and the previous Python code, and is asked to rewrite the code so that the rendered output better matches the reference.

\begin{promptbox}{Refinement system prompt}
You are an expert Python plotting assistant. Write Python code to recreate the reference image using matplotlib/numpy as needed. Save the image only to the Python variable OUTPUT\_PATH. Do not quote OUTPUT\_PATH and do not append a file extension to it. Return code only.
\end{promptbox}

\begin{promptbox}{Refinement user prompt}
You are improving Python code that recreates a reference image.

The first image is the reference image to recreate. The second image is the previous rendered output.

Previous Python code:
\texttt{\{previous\_code\}}

Rewrite the Python code so the rendered output better matches the reference image. Save the image only to the Python variable OUTPUT\_PATH. Do not quote OUTPUT\_PATH and do not append a file extension to it. Return code only.
\end{promptbox}

\paragraph{Code format.}

Vision2Code does not require a specific programming language or rendering tool; it only requires an executable artifact that can be rendered into an image. To test this, we rerun GPT-5.4 Mini on a separate 292-example DocVQA held-out set with LaTeX instead of the default Python/Matplotlib target, then compile and rasterize the LaTeX output before applying the same DocVQA rater and rubric. LaTeX performs worse in both render success and reconstruction score, likely because many DocVQA pages contain scanned forms, dense tables, and irregular layouts that are difficult to reproduce with LaTeX. We also test Excalidraw on a small set of 48 web-scraped examples, where it reaches 100\% render success and a generic-rubric score of 3.90. Appendix~\ref{app:tool-use-ablations} shows qualitative reconstruction examples for LaTeX and Excalidraw.

\begin{table}[H]
\centering
\scriptsize
\setlength{\tabcolsep}{6pt}
\renewcommand{\arraystretch}{1.10}
\caption{Code format ablation on a separate 292-example DocVQA held-out set for GPT-5.4 Mini.}
\label{tab:tool_use_ablation_docvqa}
\begin{tabular}{@{}lcc@{}}
\toprule
\textbf{Target} & \textbf{Render success} & \textbf{Score} \\
\midrule
Python & 84.9\% & 3.35 \\
LaTeX  & 65.8\% & 1.52 \\
\bottomrule
\end{tabular}
\end{table}

\newpage

\section{Reconstruction Failures}
\label{app:reconstruction-failures}

\begin{figure}[H]
\centering
\includegraphics[width=0.9\textwidth]{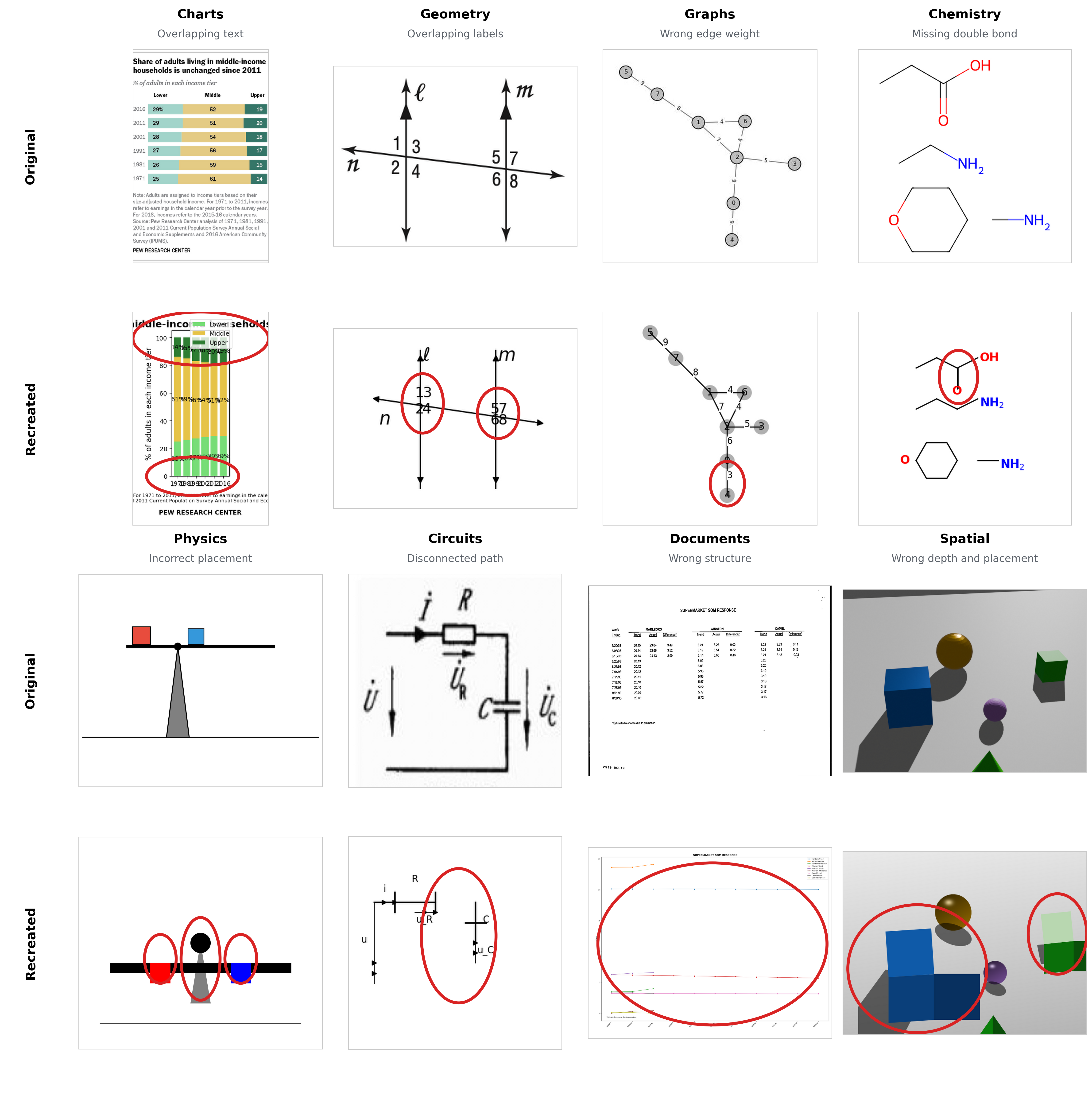}
\caption{\textbf{Representative reconstruction failures.}
Each original image is shown above its reconstruction, with the principal
error circled in red. The examples show overlapping chart text, overlapping
geometry labels, an incorrect graph edge weight, a missing chemical double
bond, incorrect object placement in a physics diagram, a disconnected
circuit, a document table rendered as a plot, and incorrect depth and
relative placement in a spatial scene.}
\label{fig:reconstruction-failures}
\end{figure}

\section{Recreation Comparisons}
\label{app:recreation-comparisons}

\begin{figure}[H]
\centering
\includegraphics[width=\linewidth,height=0.80\textheight,keepaspectratio]{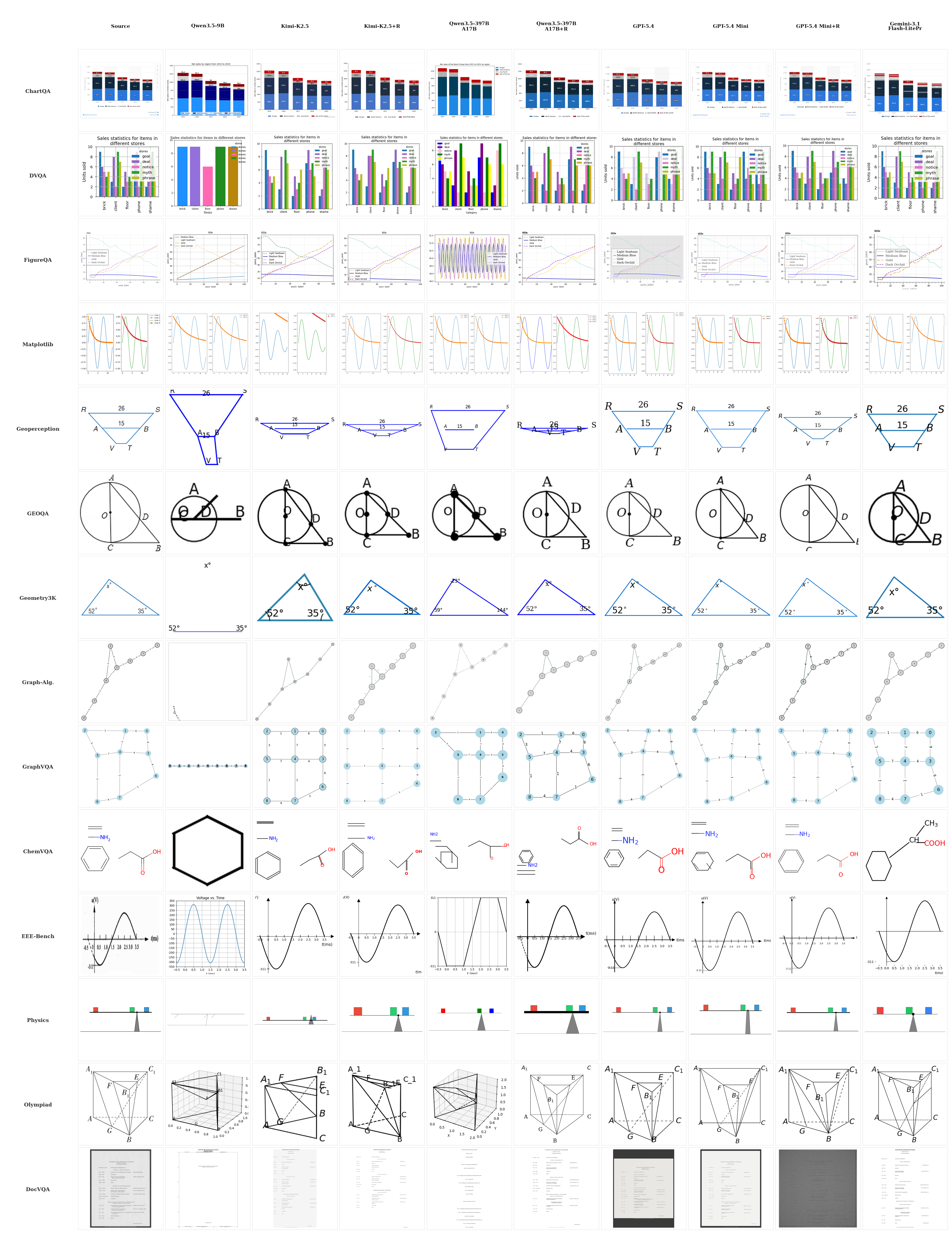}
\caption{Source and rendered-output comparisons for the nine benchmarked models.}
\label{fig:appendix-recreation-comparison-1}
\end{figure}

\clearpage
\begin{figure}[H]
\centering
\includegraphics[width=\linewidth,height=0.88\textheight,keepaspectratio]{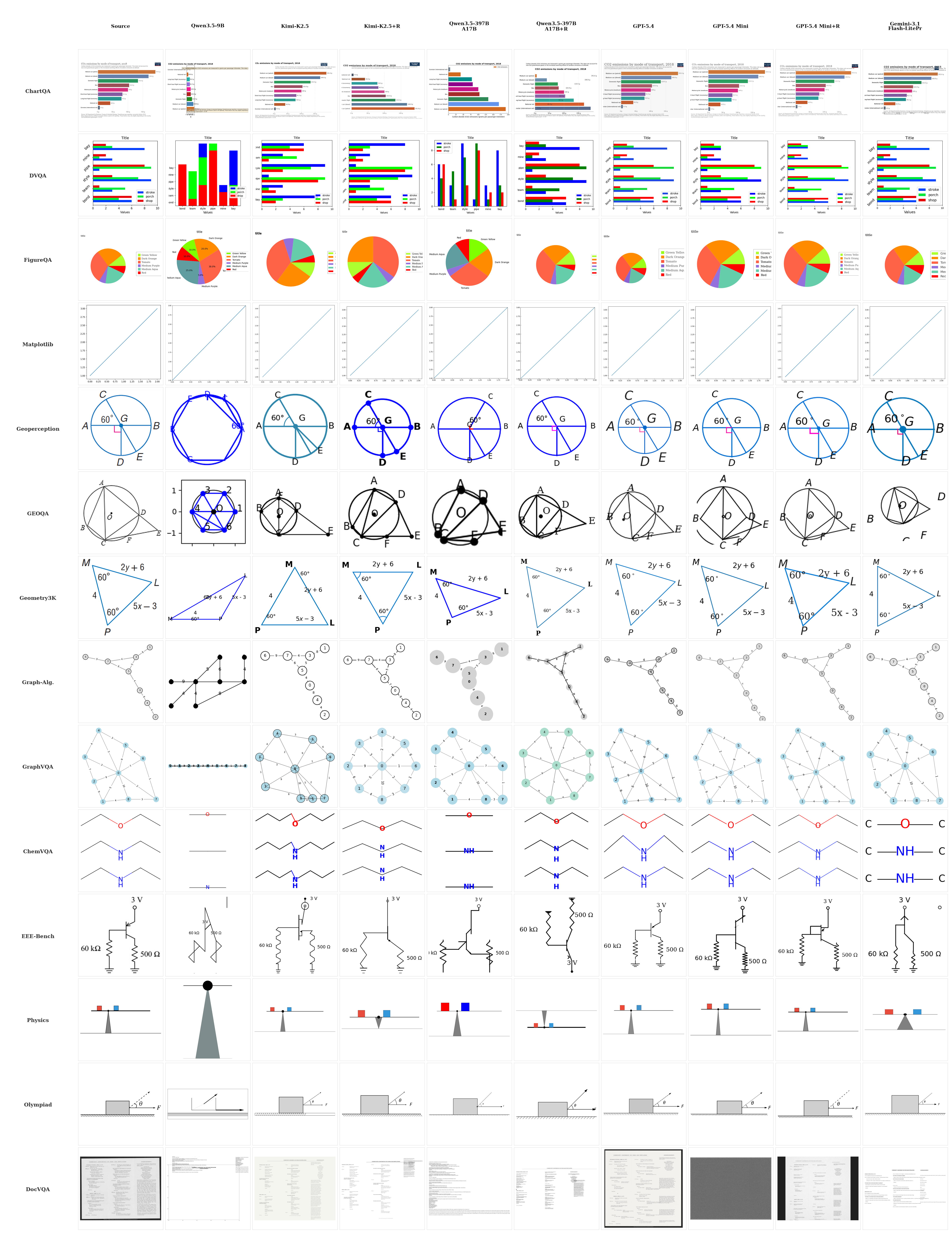}
\caption{Additional source and rendered-output comparisons for the nine benchmarked models.}
\label{fig:appendix-recreation-comparison-2}
\end{figure}

\clearpage
\begin{figure}[H]
\centering
\includegraphics[width=\linewidth,height=0.88\textheight,keepaspectratio]{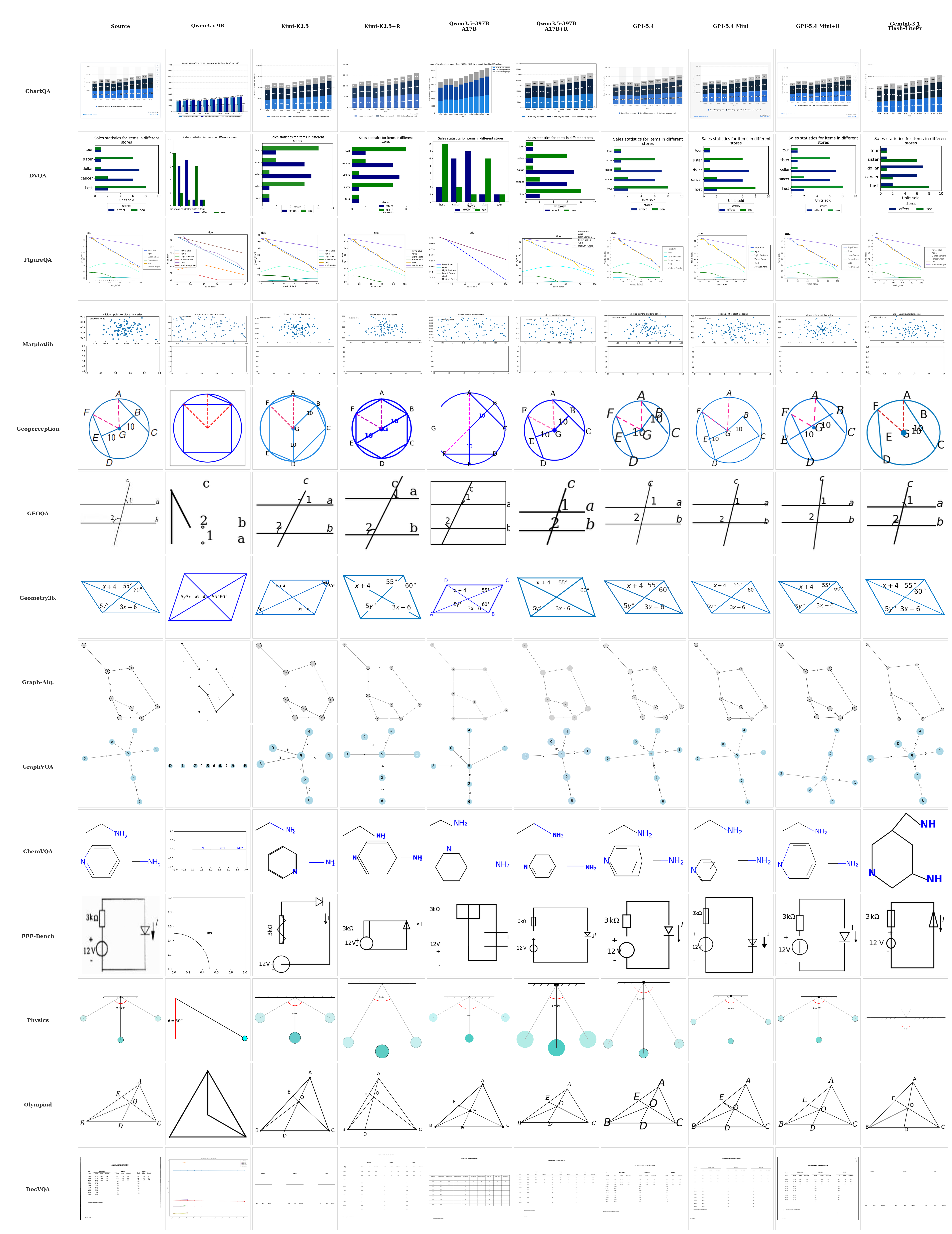}
\caption{Additional source and rendered-output comparisons for the nine benchmarked models.}
\label{fig:appendix-recreation-comparison-3}
\end{figure}

\clearpage
\begin{figure}[H]
\centering
\includegraphics[width=\linewidth,height=0.88\textheight,keepaspectratio]{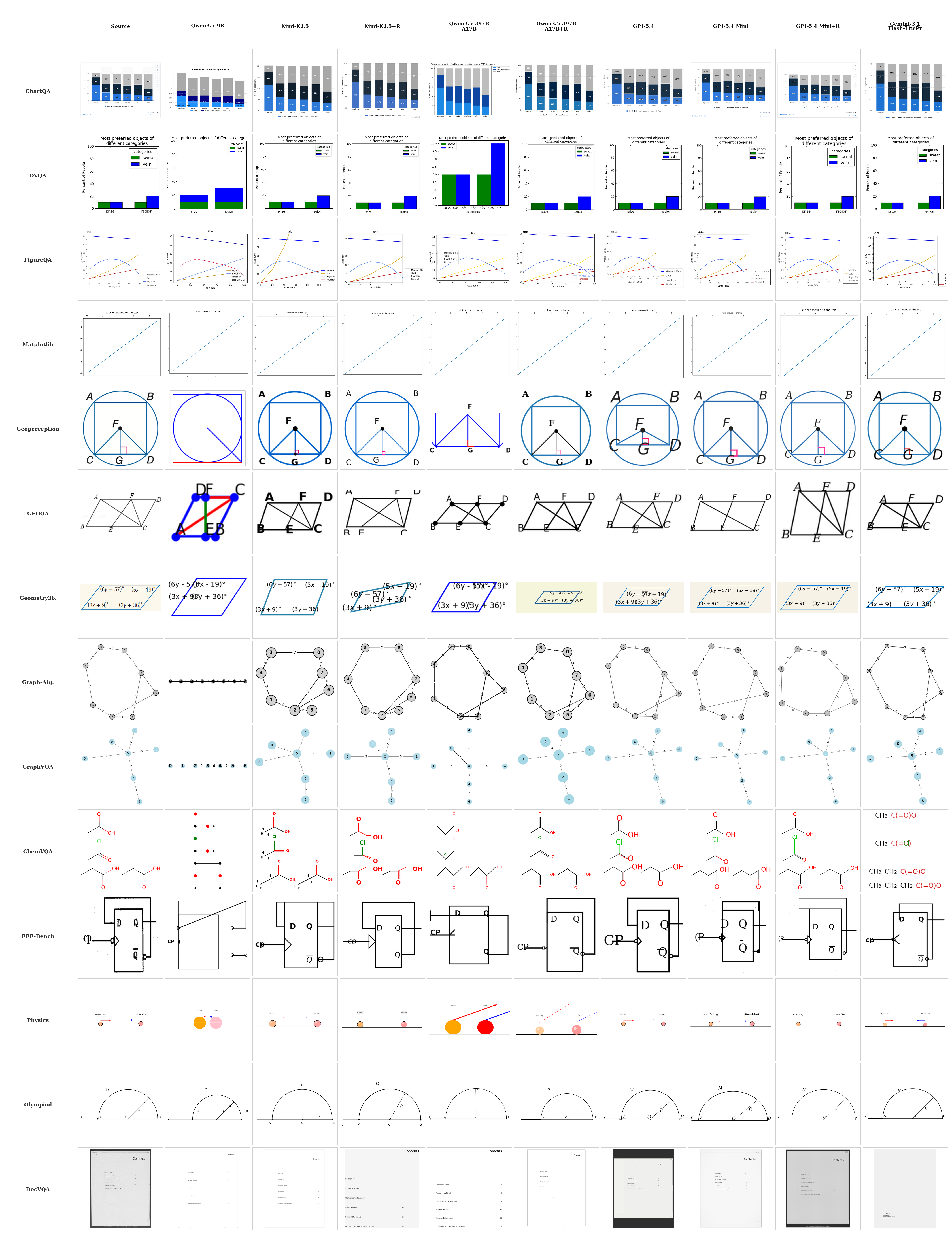}
\caption{Additional source and rendered-output comparisons for the nine benchmarked models.}
\label{fig:appendix-recreation-comparison-4}
\end{figure}

\clearpage
\section{Qualitative Rating Examples}
\label{app:qualitative-rating-examples}

We include representative test examples showing the source image, model-rendered outputs, and the score assigned by the Qwen3.5-122B-A10B-GPTQ-Int4 rater. These examples illustrate how the rater distinguishes high-fidelity recreations from outputs with missing structure, incorrect visual elements, or degraded layout quality.

\begin{figure}[H]
    \centering
    \includegraphics[width=\textwidth]{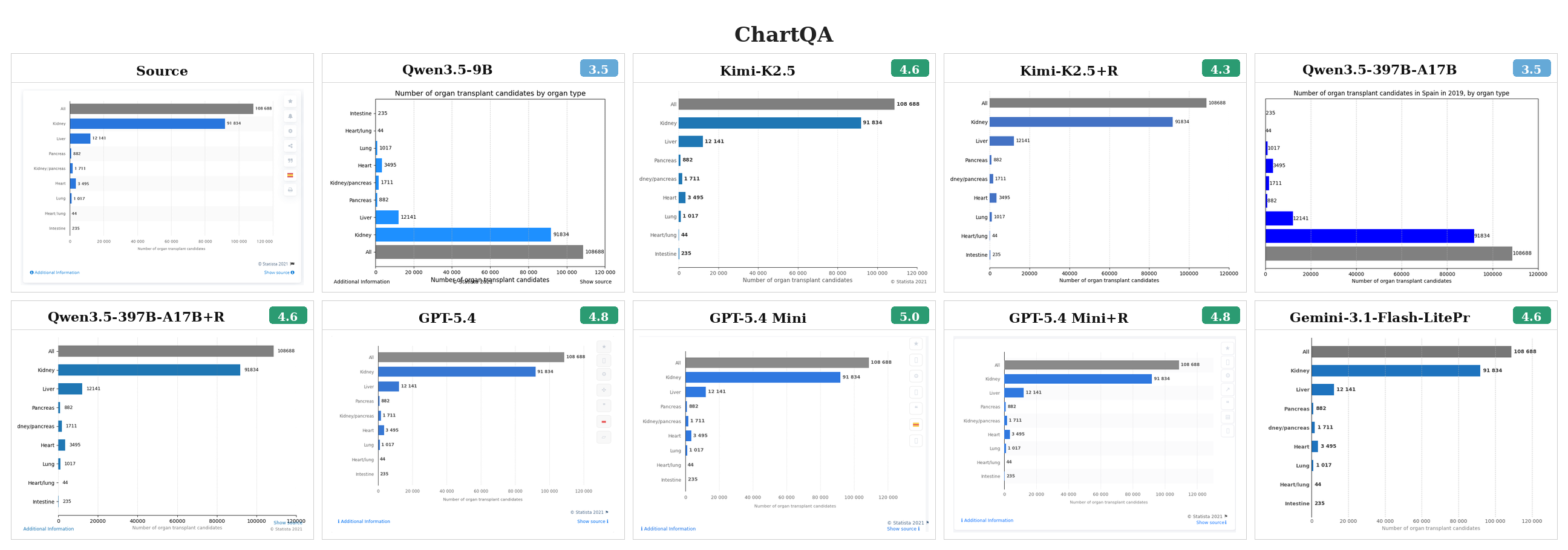}
    \caption{ChartQA rated renders by the Qwen3.5-122B-A10B-GPTQ-Int4 rater on test examples.}
    \label{fig:rating-example-chartqa}
\end{figure}

\begin{figure}[H]
    \centering
    \includegraphics[width=\textwidth]{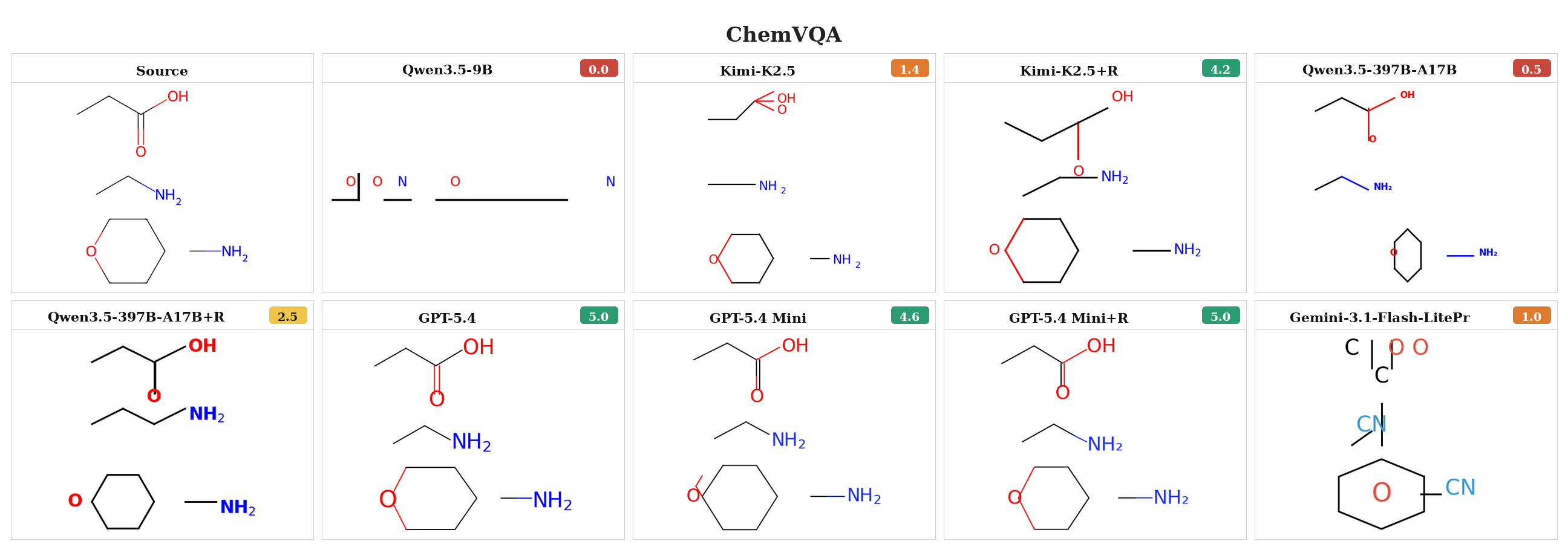}
    \caption{ChemVQA-2K rated renders by the Qwen3.5-122B-A10B-GPTQ-Int4 rater on test examples.}
    \label{fig:rating-example-chemvqa-2k}
\end{figure}

\begin{figure}[H]
    \centering
    \includegraphics[width=\textwidth]{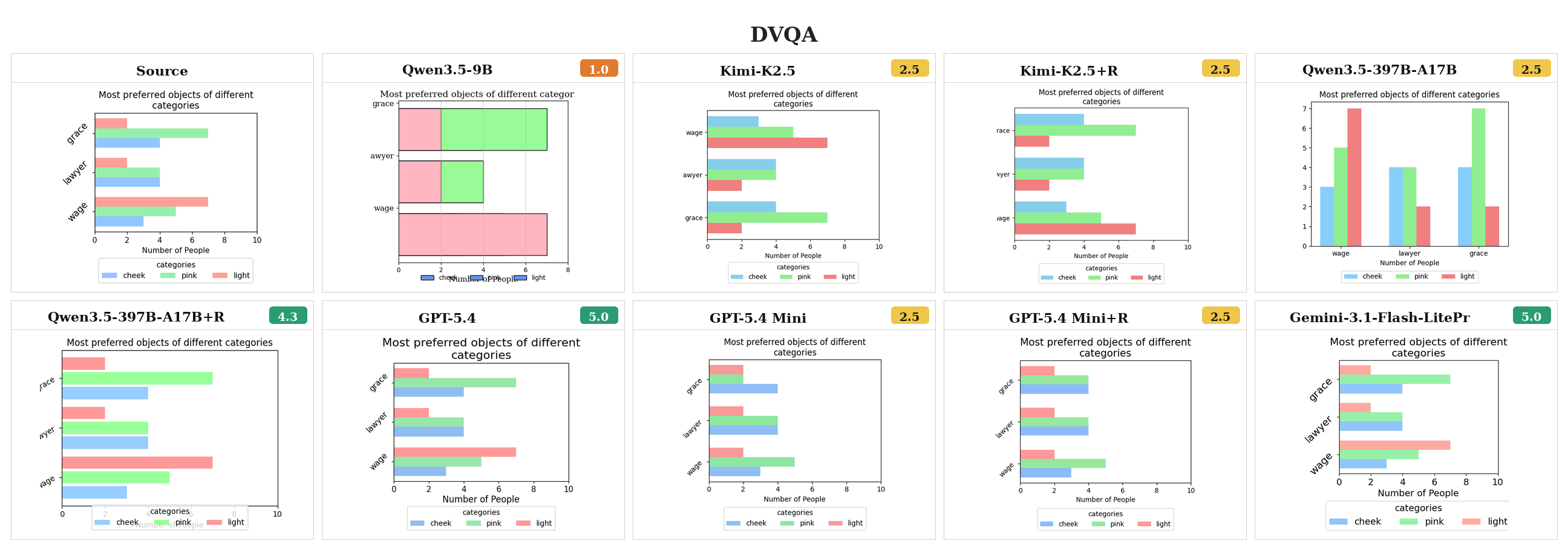}
    \caption{DVQA rated renders by the Qwen3.5-122B-A10B-GPTQ-Int4 rater on test examples.}
    \label{fig:rating-example-dvqa}
\end{figure}

\begin{figure}[H]
    \centering
    \includegraphics[width=\textwidth]{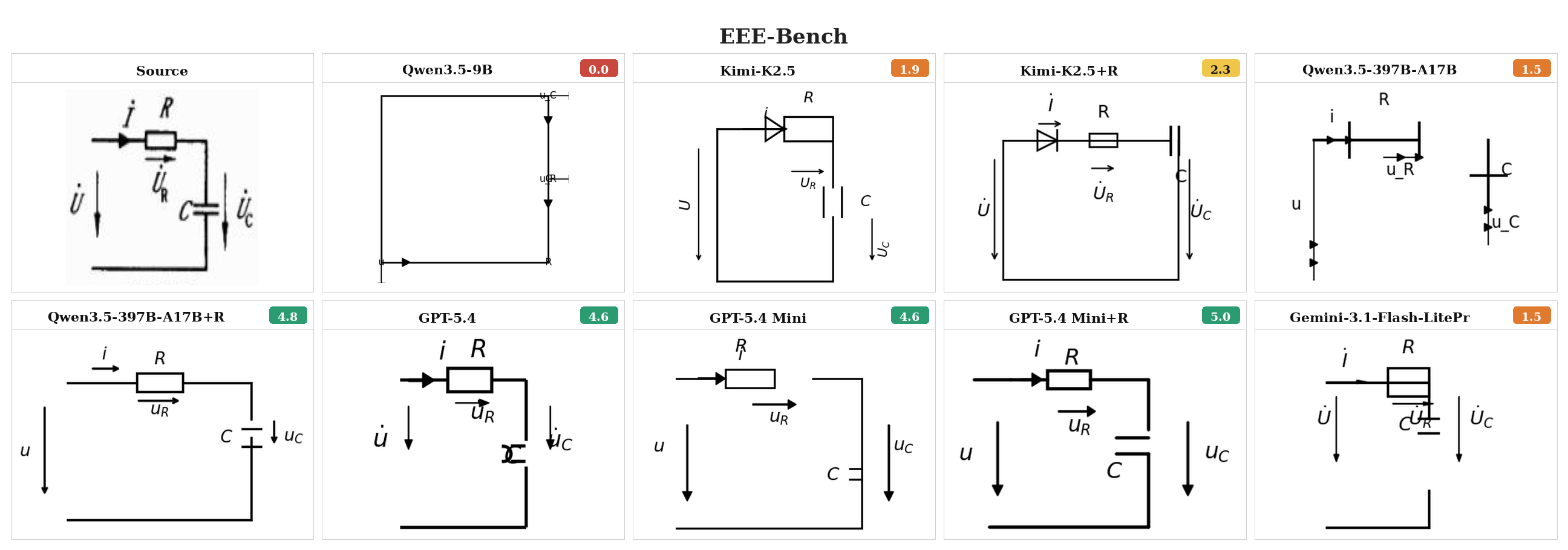}
    \caption{EEE-Bench rated renders by the Qwen3.5-122B-A10B-GPTQ-Int4 rater on test examples.}
    \label{fig:rating-example-eee-bench}
\end{figure}

\begin{figure}[H]
    \centering
    \includegraphics[width=\textwidth]{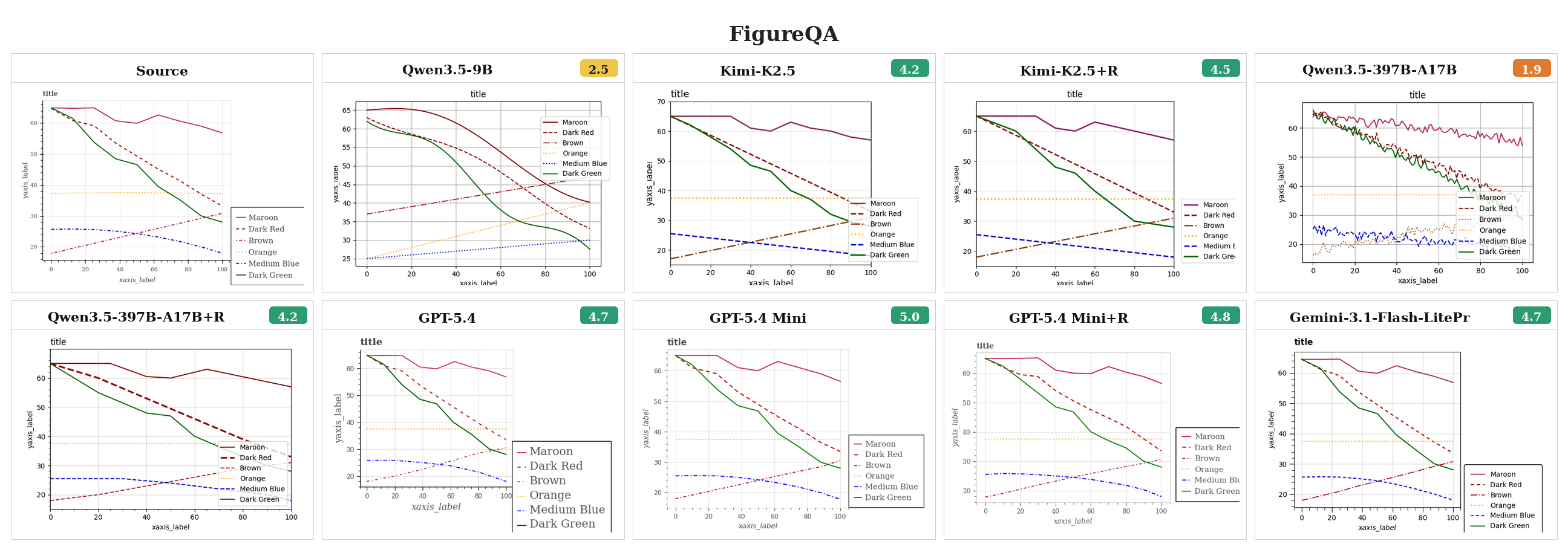}
    \caption{FigureQA rated renders by the Qwen3.5-122B-A10B-GPTQ-Int4 rater on test examples.}
    \label{fig:rating-example-figureqa}
\end{figure}

\begin{figure}[H]
    \centering
    \includegraphics[width=\textwidth]{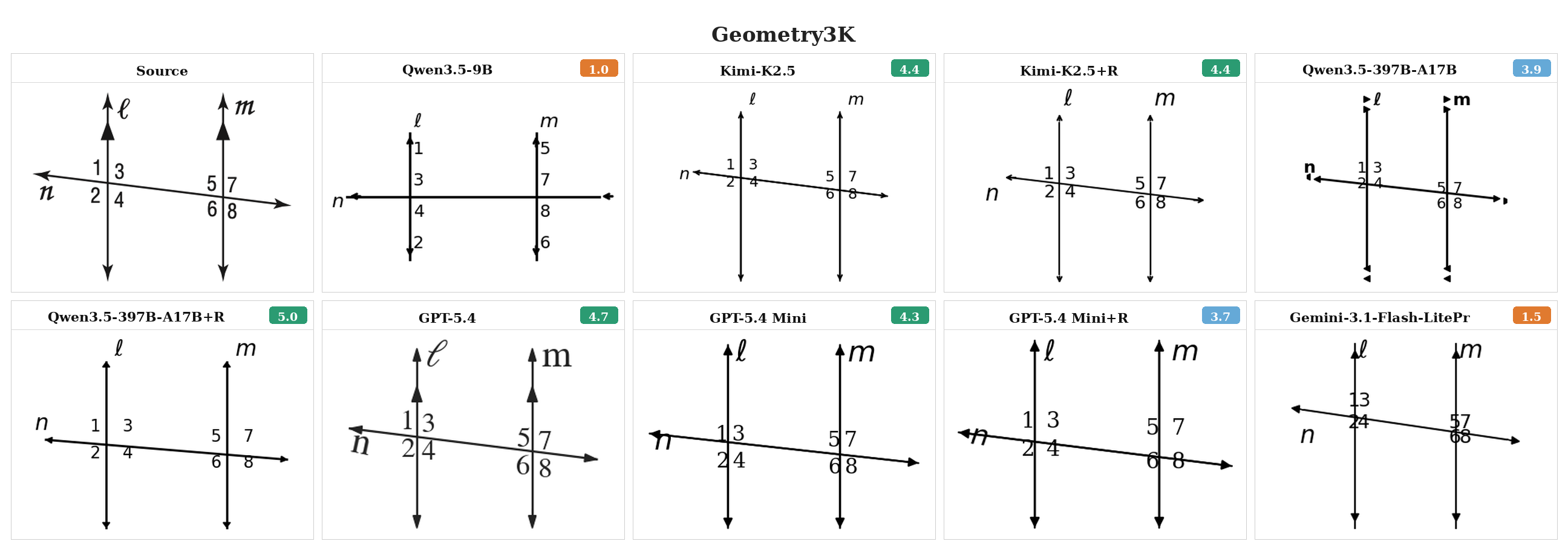}
    \caption{Geometry3K rated renders by the Qwen3.5-122B-A10B-GPTQ-Int4 rater on test examples.}
    \label{fig:rating-example-geometry3k}
\end{figure}

\begin{figure}[H]
    \centering
    \includegraphics[width=\textwidth]{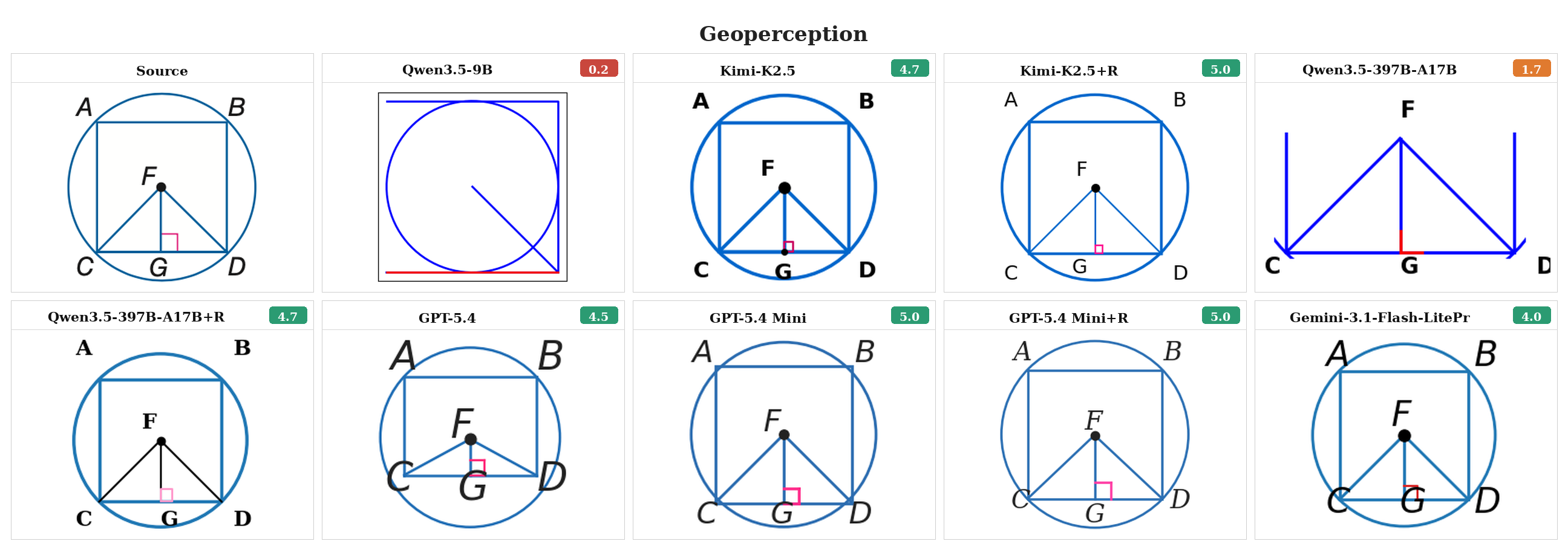}
    \caption{Geoperception rated renders by the Qwen3.5-122B-A10B-GPTQ-Int4 rater on test examples.}
    \label{fig:rating-example-geoperception}
\end{figure}

\begin{figure}[H]
    \centering
    \includegraphics[width=\textwidth]{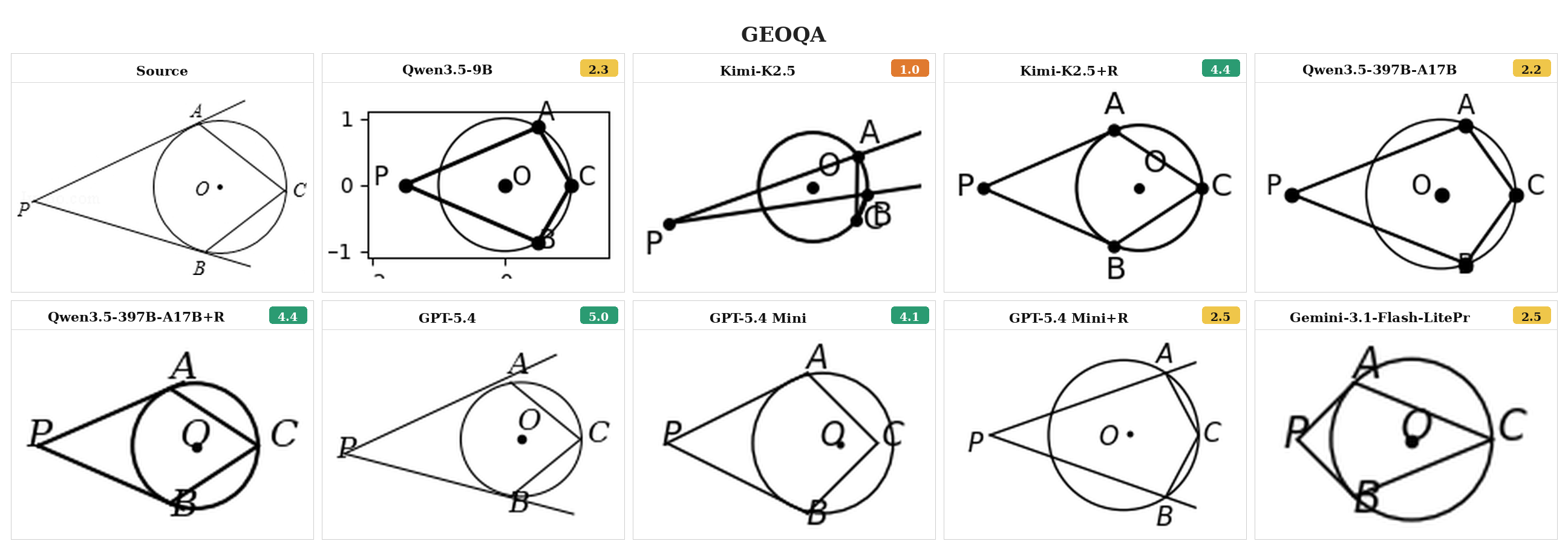}
    \caption{GeoQA-8K rated renders by the Qwen3.5-122B-A10B-GPTQ-Int4 rater on test examples.}
    \label{fig:rating-example-geoqa-8k}
\end{figure}

\begin{figure}[H]
    \centering
    \includegraphics[width=\textwidth]{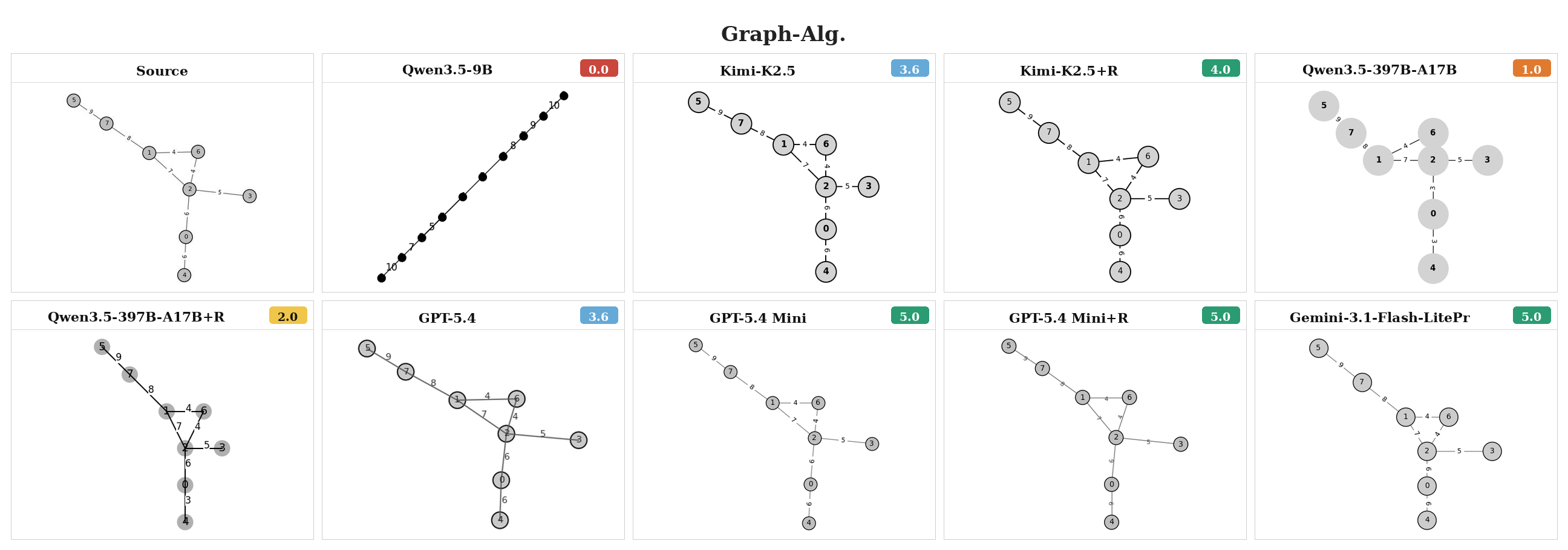}
    \caption{Graph Algorithms rated renders by the Qwen3.5-122B-A10B-GPTQ-Int4 rater on test examples.}
    \label{fig:rating-example-graph-algorithms}
\end{figure}

\begin{figure}[H]
    \centering
    \includegraphics[width=\textwidth]{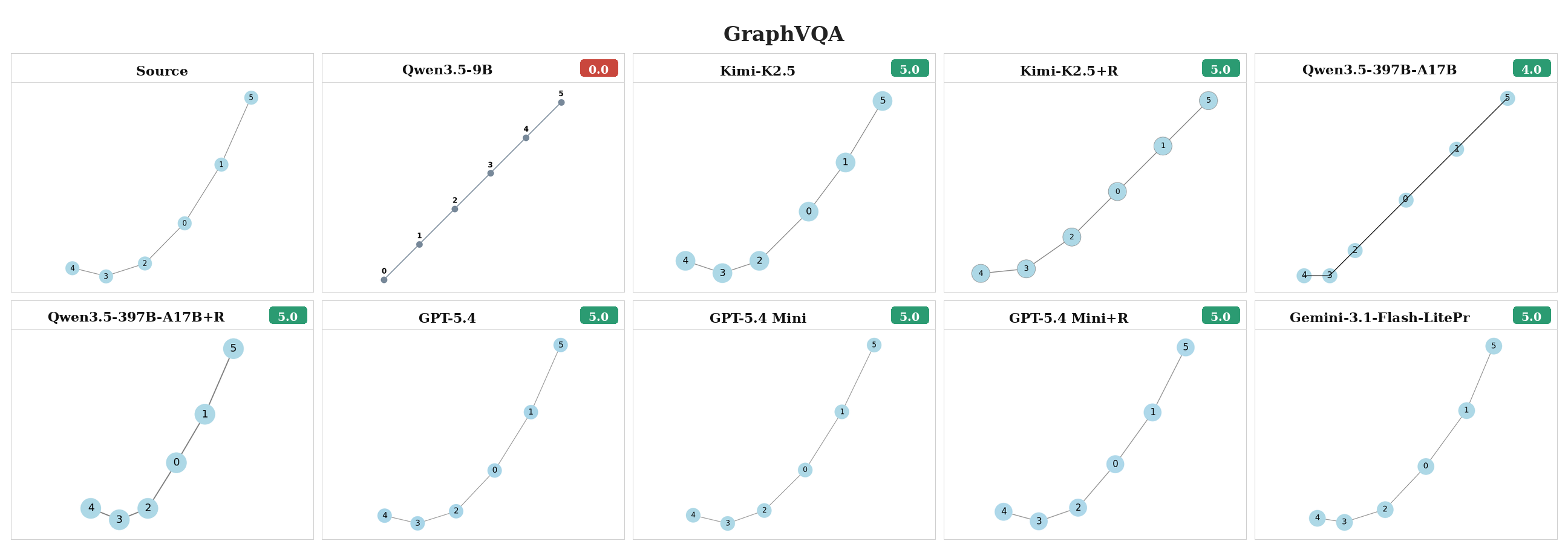}
    \caption{GraphVQA-Swift rated renders by the Qwen3.5-122B-A10B-GPTQ-Int4 rater on test examples.}
    \label{fig:rating-example-graphvqa-swift}
\end{figure}

\begin{figure}[H]
    \centering
    \includegraphics[width=\textwidth]{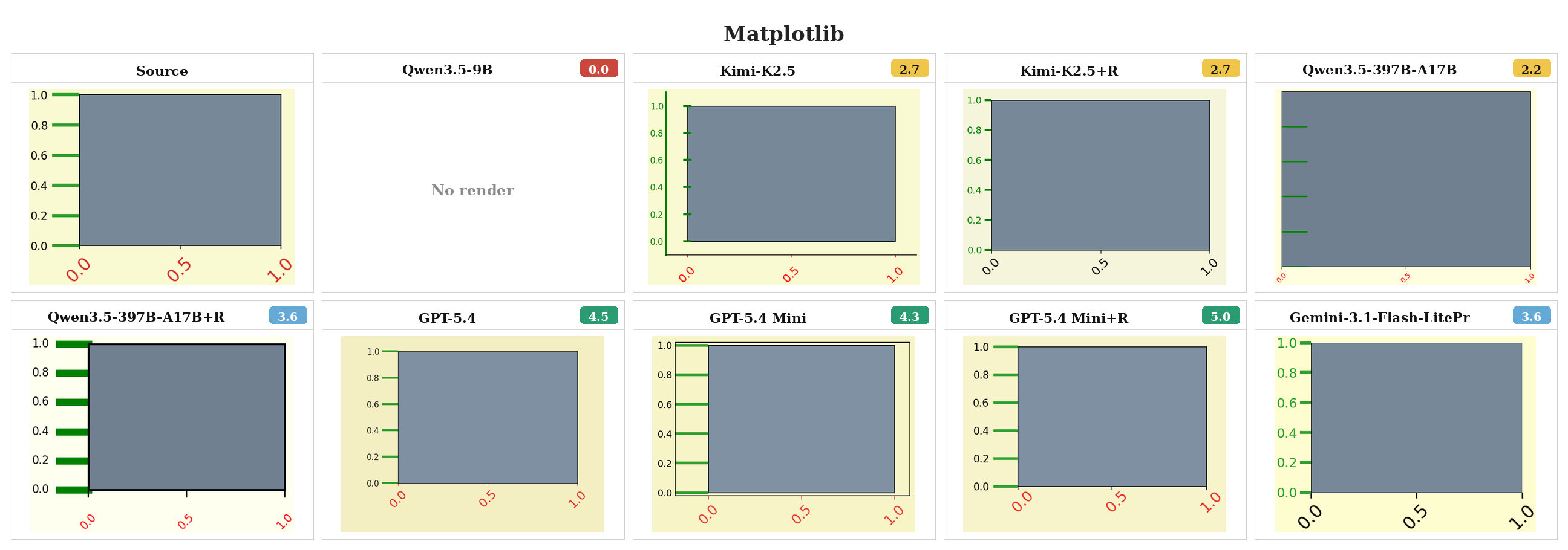}
    \caption{Matplotlib rated renders by the Qwen3.5-122B-A10B-GPTQ-Int4 rater on test examples.}
    \label{fig:rating-example-matplotlib}
\end{figure}

\begin{figure}[H]
    \centering
    \includegraphics[width=\textwidth]{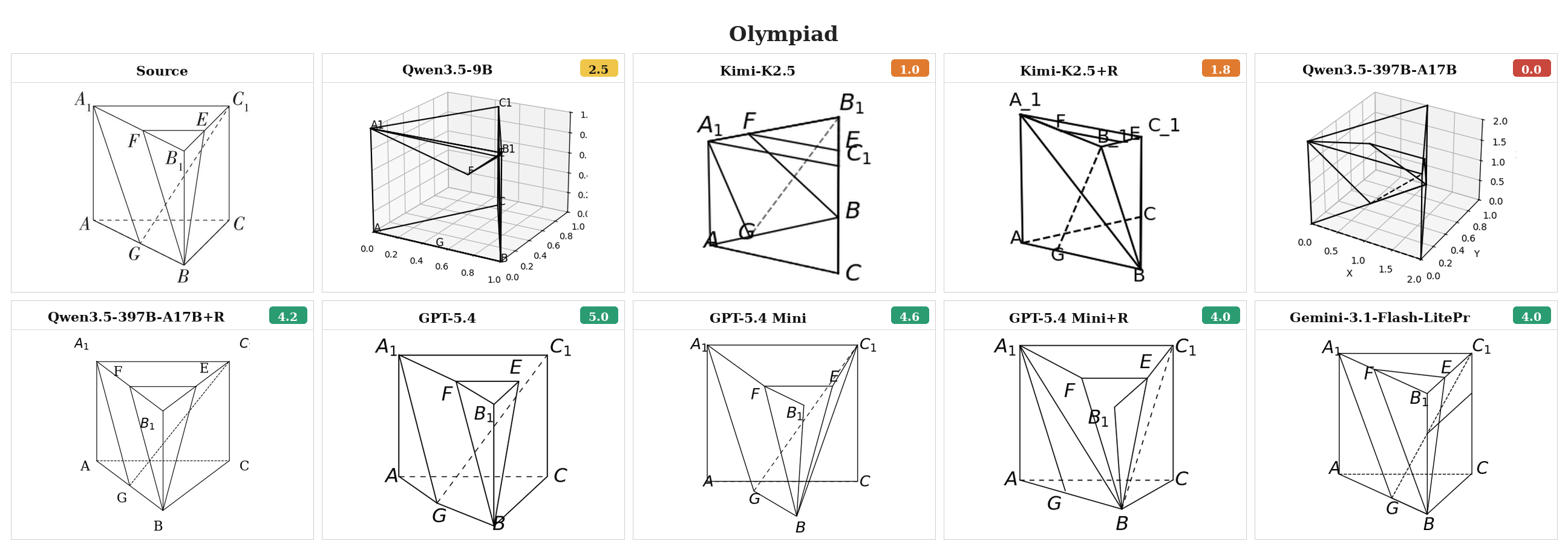}
    \caption{OlympiadBench rated renders by the Qwen3.5-122B-A10B-GPTQ-Int4 rater on test examples.}
    \label{fig:rating-example-olympiadbench}
\end{figure}

\begin{figure}[H]
    \centering
    \includegraphics[width=\textwidth]{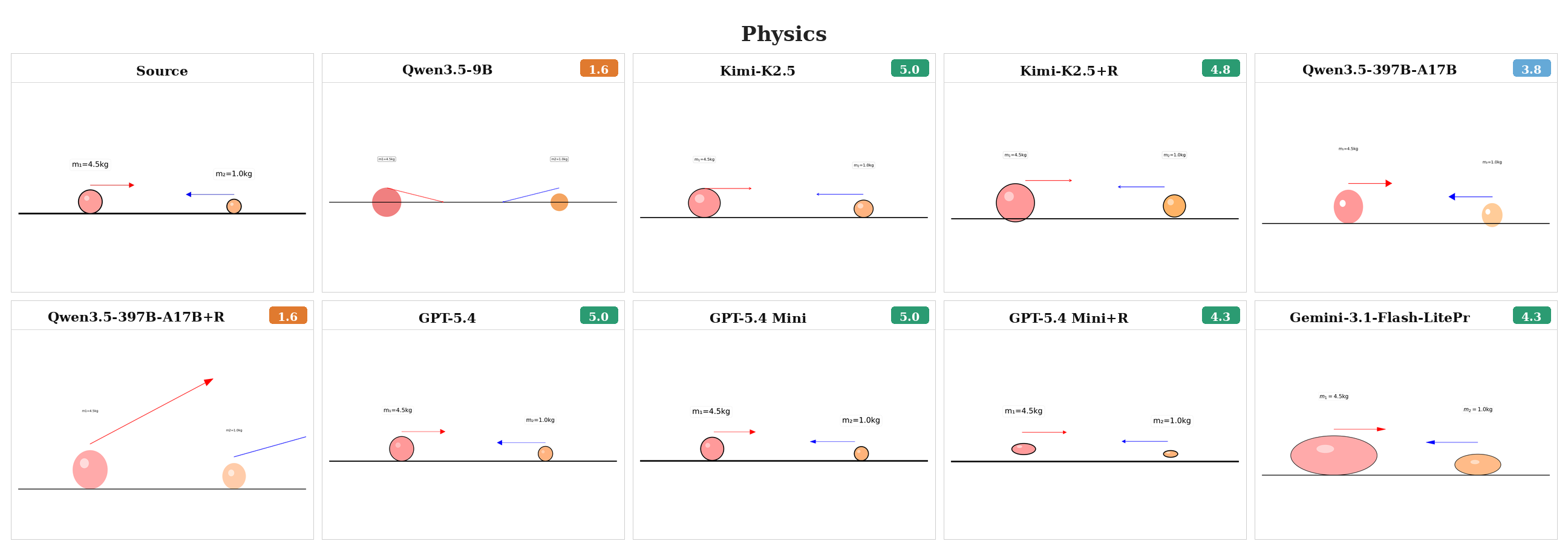}
    \caption{Physics rated renders by the Qwen3.5-122B-A10B-GPTQ-Int4 rater on test examples.}
    \label{fig:rating-example-physics}
\end{figure}

\begin{figure}[H]
    \centering
    \includegraphics[width=\textwidth]{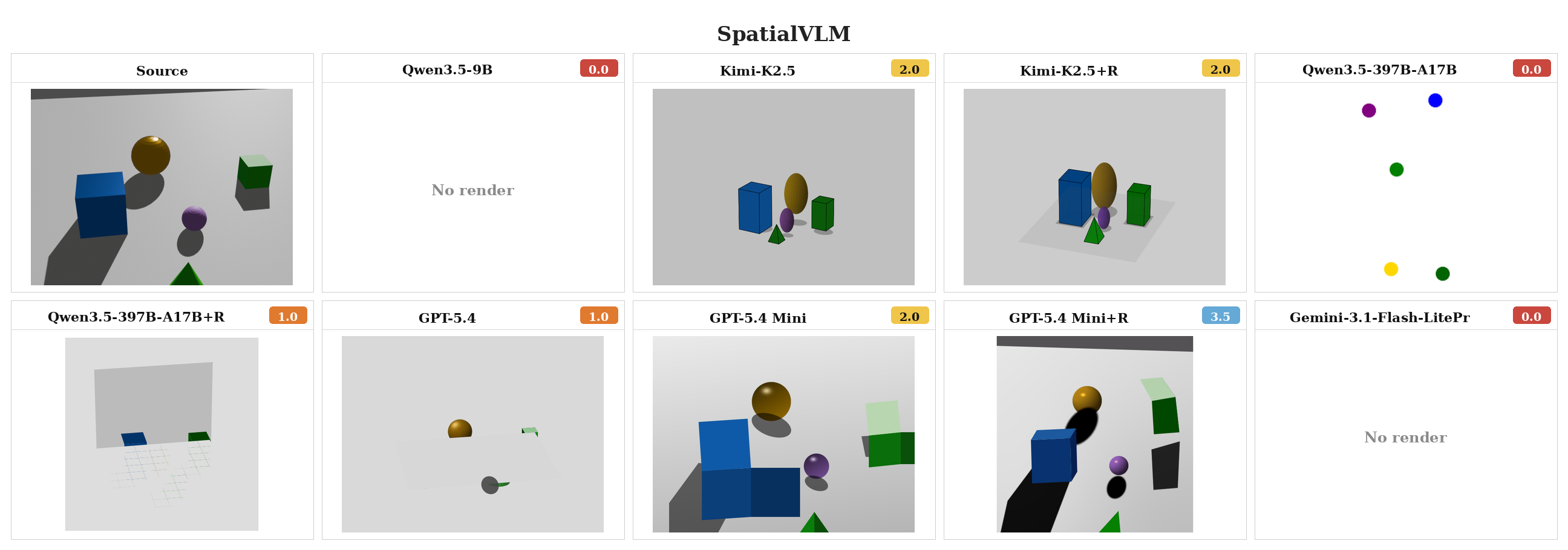}
    \caption{SpatialVLM-QA rated renders by the Qwen3.5-122B-A10B-GPTQ-Int4 rater on test examples.}
    \label{fig:rating-example-spatialvlm-qa}
\end{figure}

\begin{figure}[H]
    \centering
    \includegraphics[width=\textwidth]{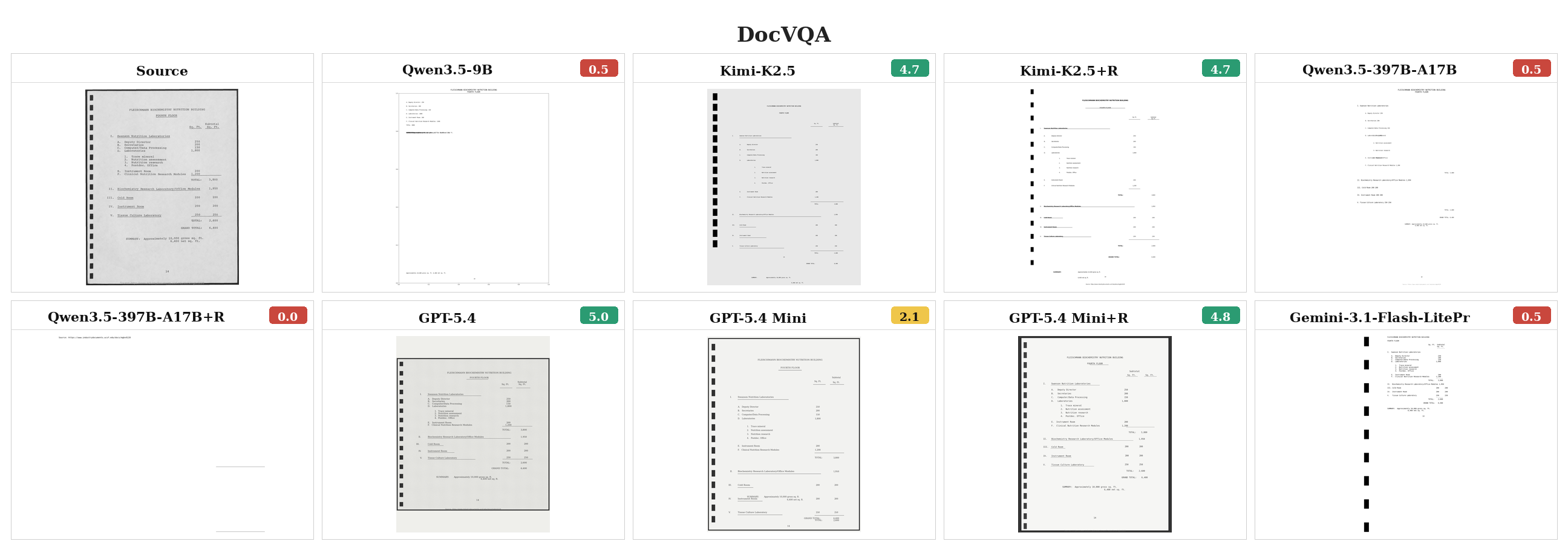}
    \caption{DocVQA rated renders by the Qwen3.5-122B-A10B-GPTQ-Int4 rater on test examples.}
    \label{fig:rating-example-docvqa}
\end{figure}

\clearpage

\section{Evaluator Robustness Qualitative Examples}
\label{app:evaluator-robustness-examples}

\begin{figure*}[h]
    \centering
    \includegraphics[width=\textwidth]
    {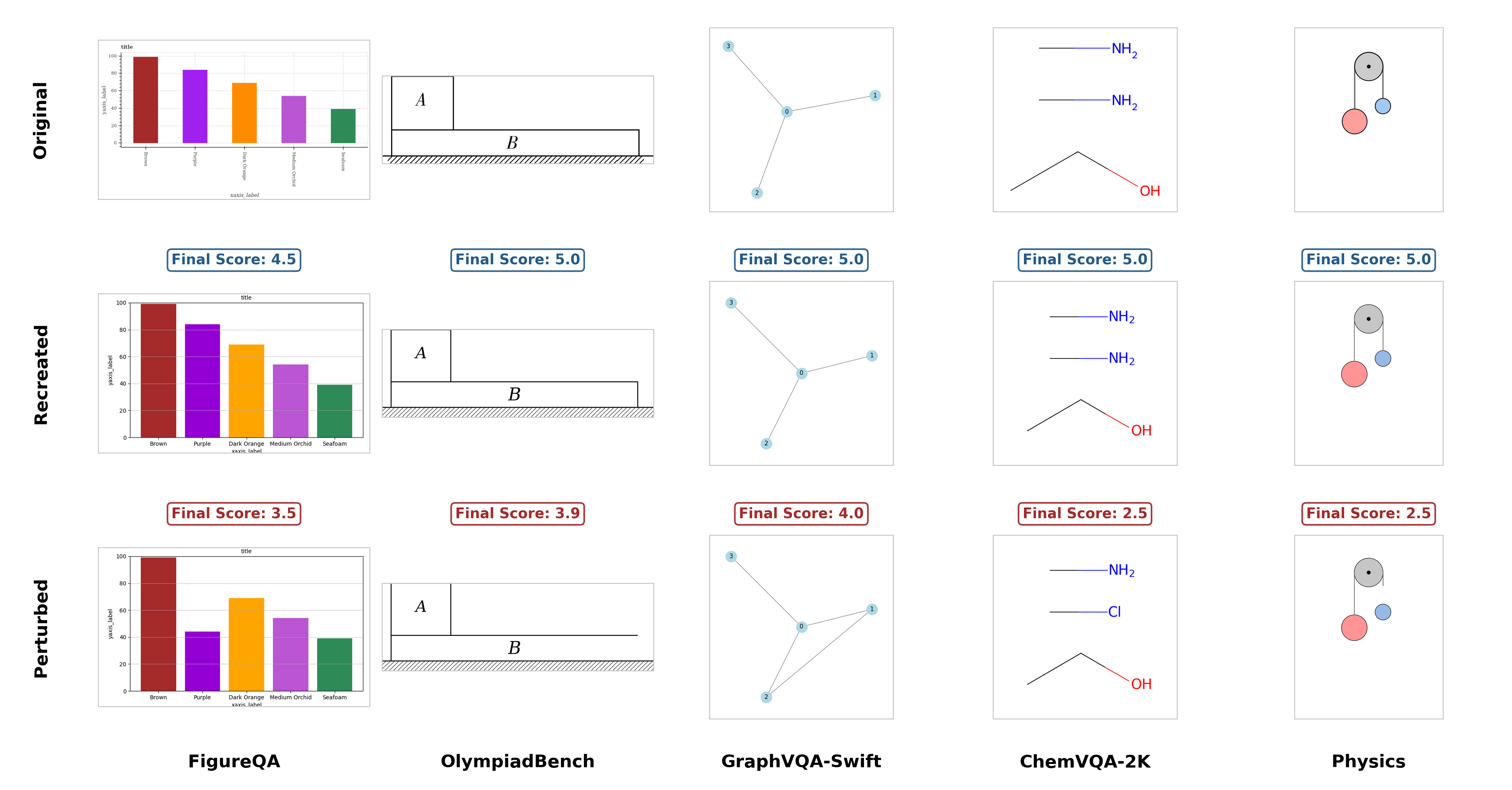}
    \caption{
    Qualitative examples evaluated by Qwen3.5-122B. The rows show the source
    image, its reconstruction, and the perturbed reconstruction. The targeted
    category scores decrease as follows: FigureQA, Numeric Information Fidelity
    ($5.0 \rightarrow 4.0$); OlympiadBench, Readability and Completeness
    ($5.0 \rightarrow 3.0$); GraphVQA-Swift, Graph Structure Fidelity
    ($5.0 \rightarrow 2.0$); ChemVQA-2K, Chemical Symbol Accuracy
    ($5.0 \rightarrow 2.0$); and Physics, Physical Setup Fidelity
    ($5.0 \rightarrow 2.0$).
    }
    \label{fig:eval-robustness-qwen}
\end{figure*}

\begin{figure*}[h]
    \centering
    \includegraphics[width=\textwidth]
    {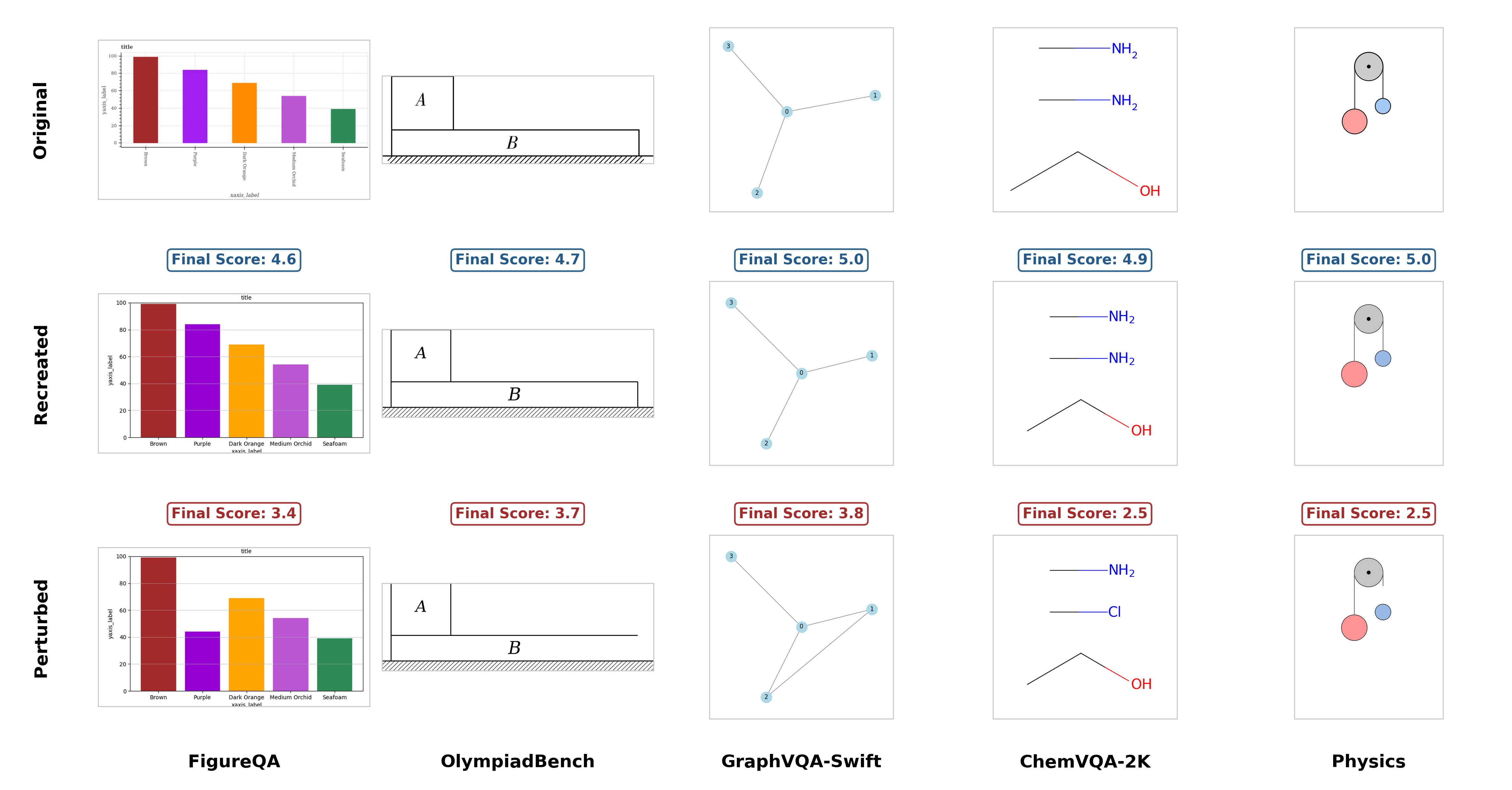}
    \caption{
    The same controlled perturbations evaluated by Kimi-K2.6. The targeted
    category scores decrease as follows: FigureQA, Numeric Information Fidelity
    ($5.0 \rightarrow 2.5$); OlympiadBench, Readability and Completeness
    ($4.6 \rightarrow 3.5$); GraphVQA-Swift, Graph Structure Fidelity
    ($5.0 \rightarrow 2.0$); ChemVQA-2K, Chemical Symbol Accuracy
    ($5.0 \rightarrow 2.0$); and Physics, Physical Setup Fidelity
    ($5.0 \rightarrow 2.5$).
    }
    \label{fig:eval-robustness-kimi}
\end{figure*}

Figures~\ref{fig:eval-robustness-qwen} and
\ref{fig:eval-robustness-kimi} show controlled perturbations evaluated by
Qwen3.5-122B and Kimi-K2.6, respectively. Each perturbation introduces one
targeted error while leaving the remaining image unchanged. The displayed
values are the final rubric scores.

\clearpage

\section{Tool-Use Ablation Qualitative Examples}
\label{app:tool-use-ablations}

\subsection{Excalidraw Recreation}
\label{app:tool-use-excalidraw}
Excalidraw illustrates a practical Vision2Code setting: the model predicts an editable structured artifact, which is rendered back to an image.

\begin{figure}[H]
\centering
\scriptsize
\setlength{\tabcolsep}{3pt}
\renewcommand{\arraystretch}{0.92}
\begin{tabular}{@{}p{0.48\linewidth}p{0.48\linewidth}@{}}
\centering\textbf{Source} & \centering\arraybackslash\textbf{Recreated from generated Excalidraw JSON} \\
\includegraphics[width=\linewidth,height=0.12\textheight,keepaspectratio]{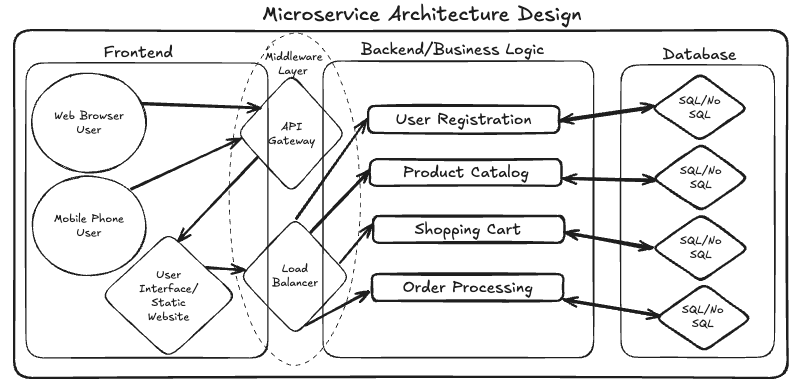} &
\includegraphics[width=\linewidth,height=0.12\textheight,keepaspectratio]{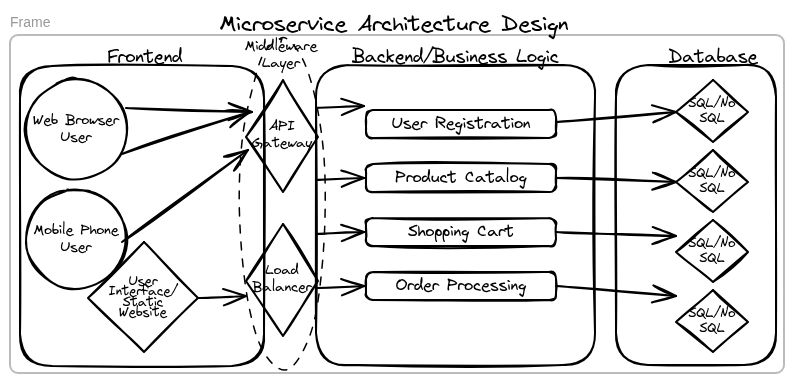} \\
\includegraphics[width=\linewidth,height=0.12\textheight,keepaspectratio]{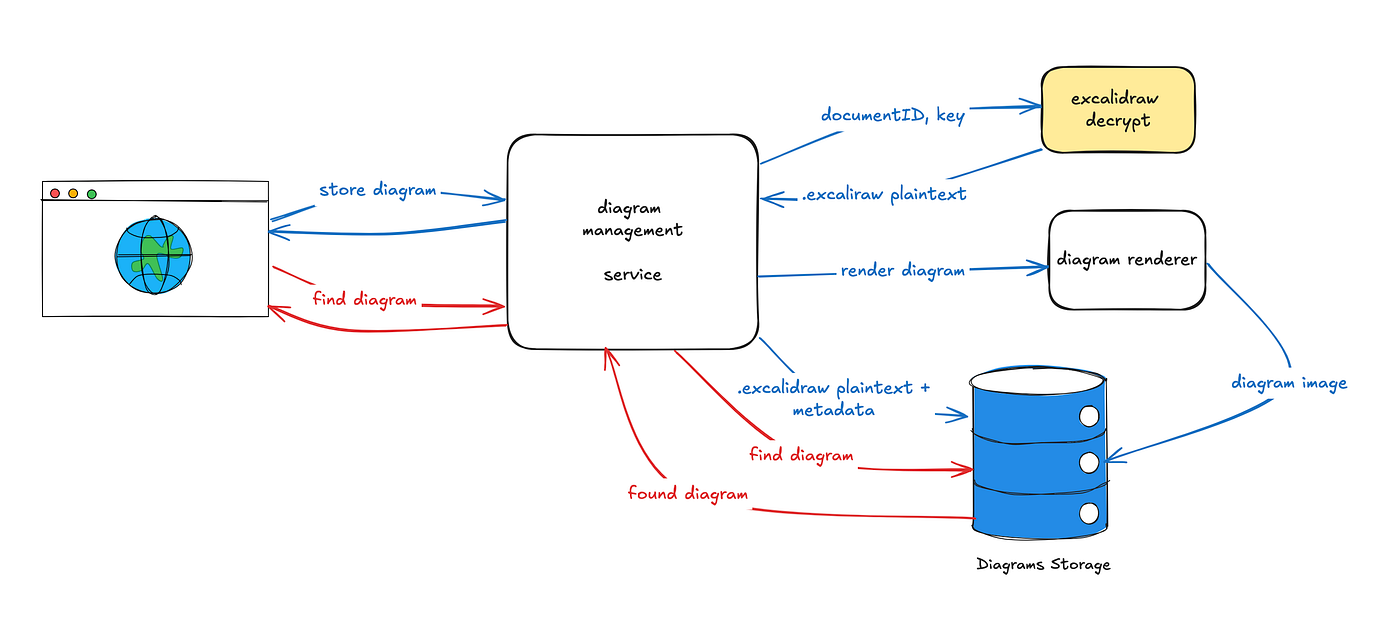} &
\includegraphics[width=\linewidth,height=0.12\textheight,keepaspectratio]{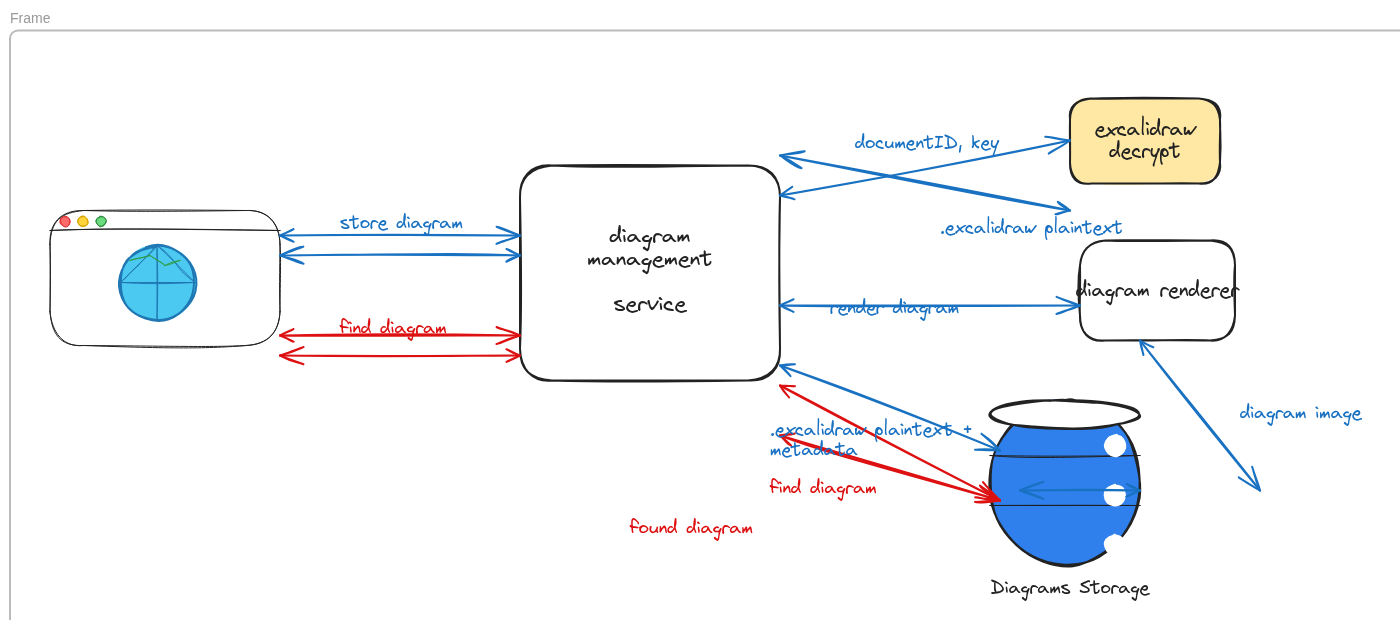} \\
\includegraphics[width=\linewidth,height=0.12\textheight,keepaspectratio]{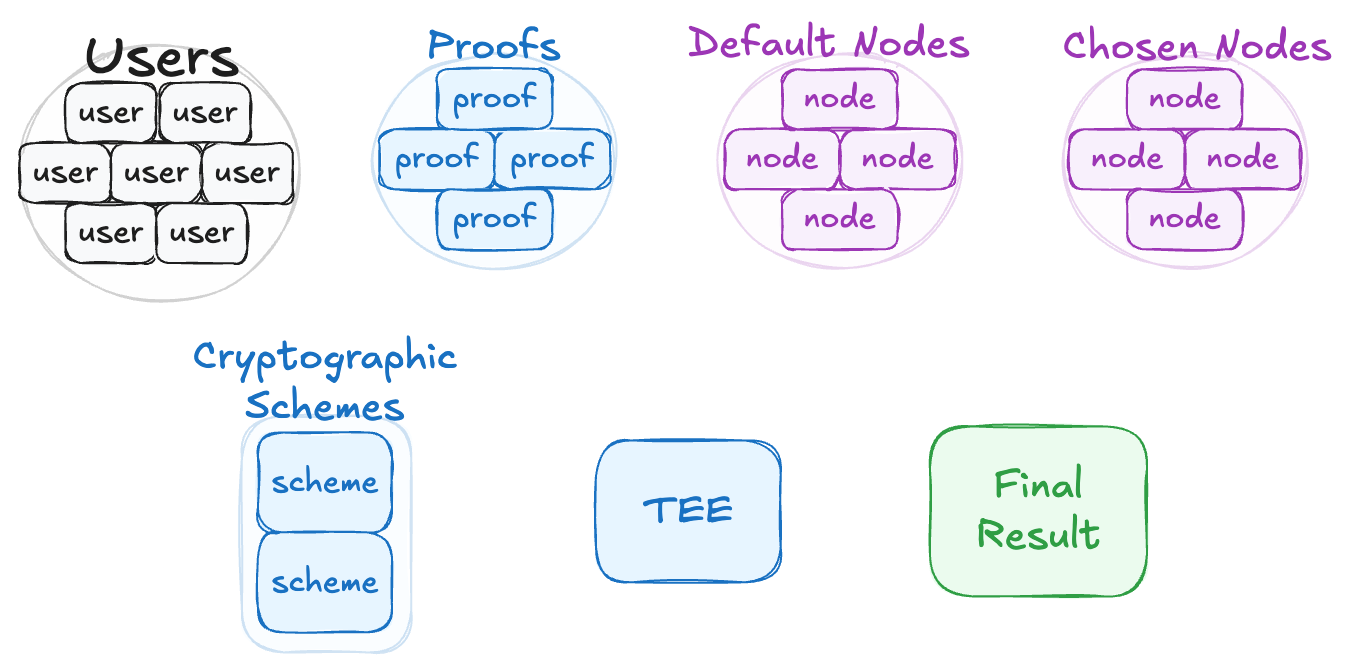} &
\includegraphics[width=\linewidth,height=0.12\textheight,keepaspectratio]{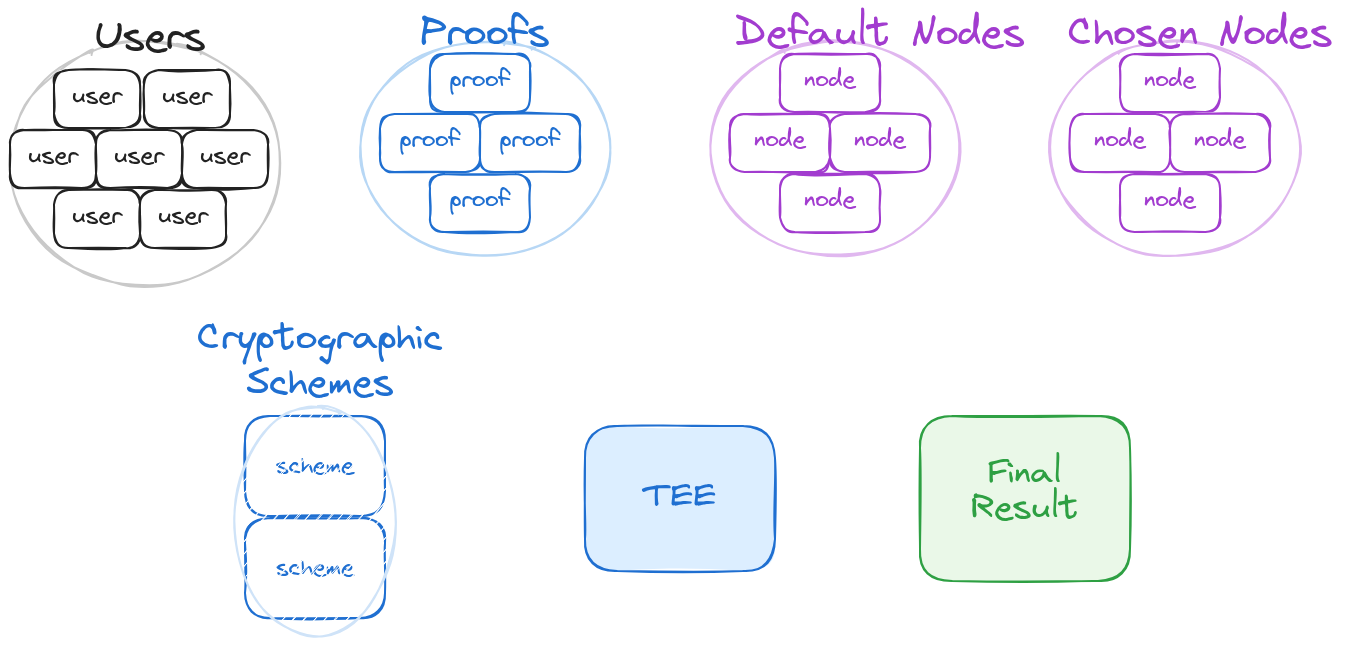} \\
\includegraphics[width=\linewidth,height=0.12\textheight,keepaspectratio]{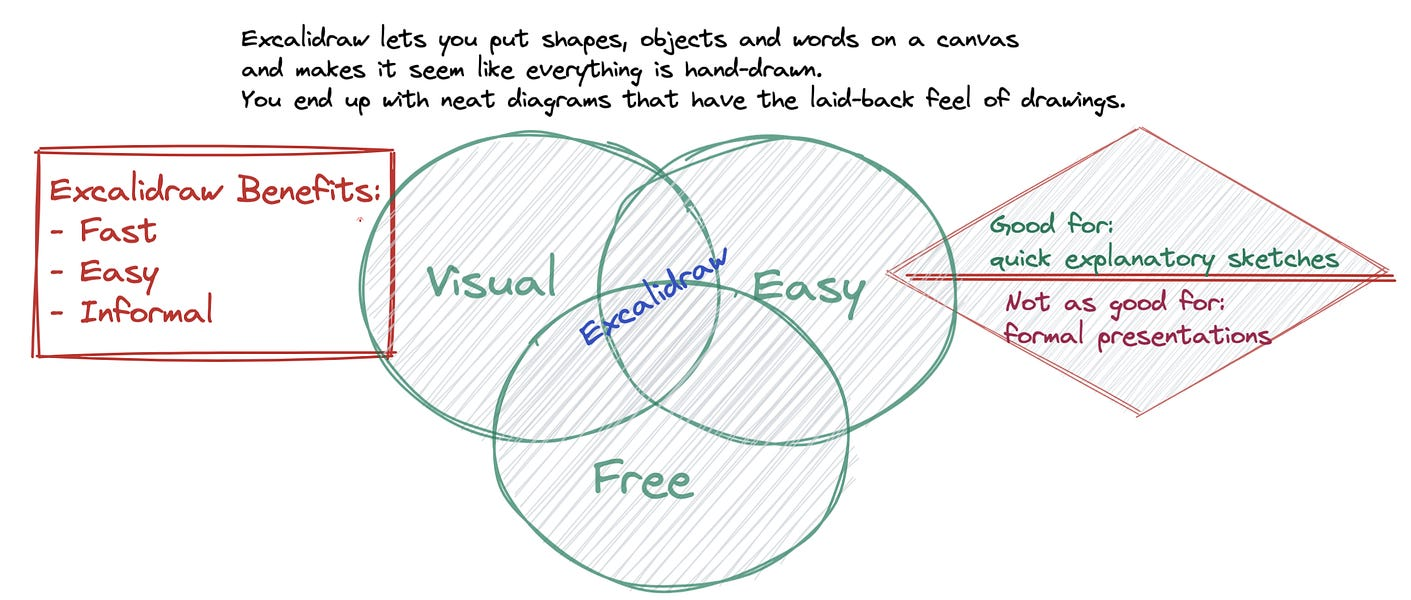} &
\includegraphics[width=\linewidth,height=0.12\textheight,keepaspectratio]{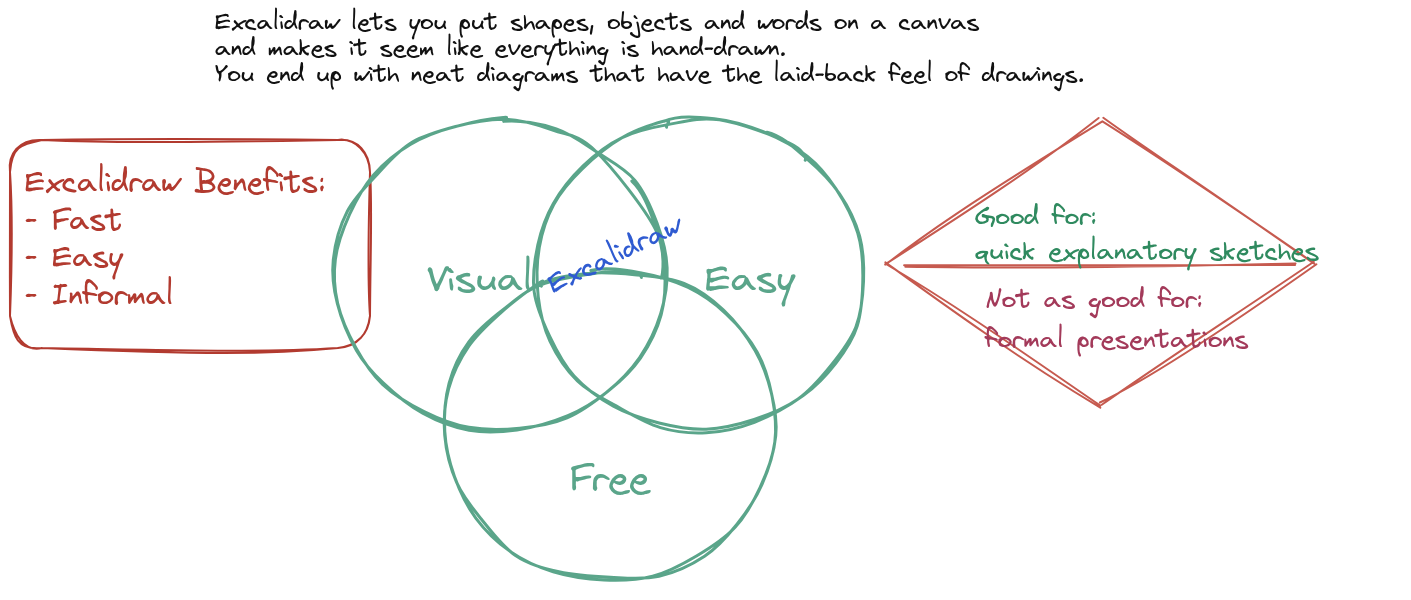} \\
\includegraphics[width=\linewidth,height=0.12\textheight,keepaspectratio]{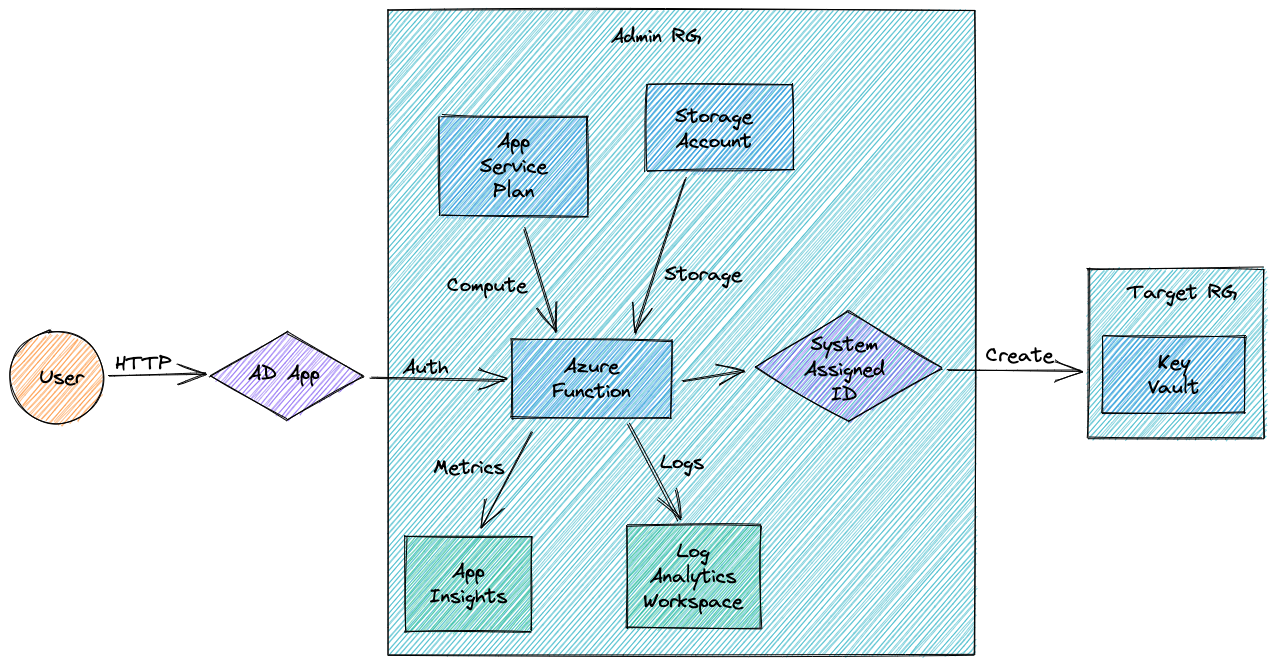} &
\includegraphics[width=\linewidth,height=0.12\textheight,keepaspectratio]{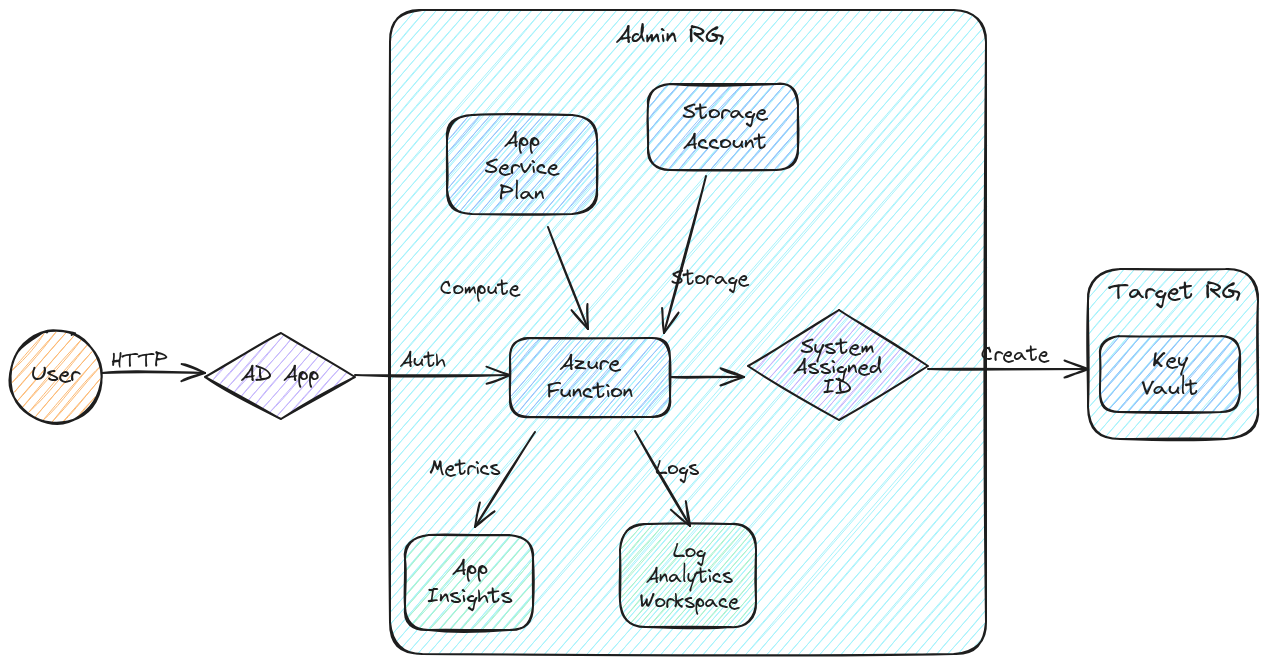} \\
\includegraphics[width=\linewidth,height=0.12\textheight,keepaspectratio]{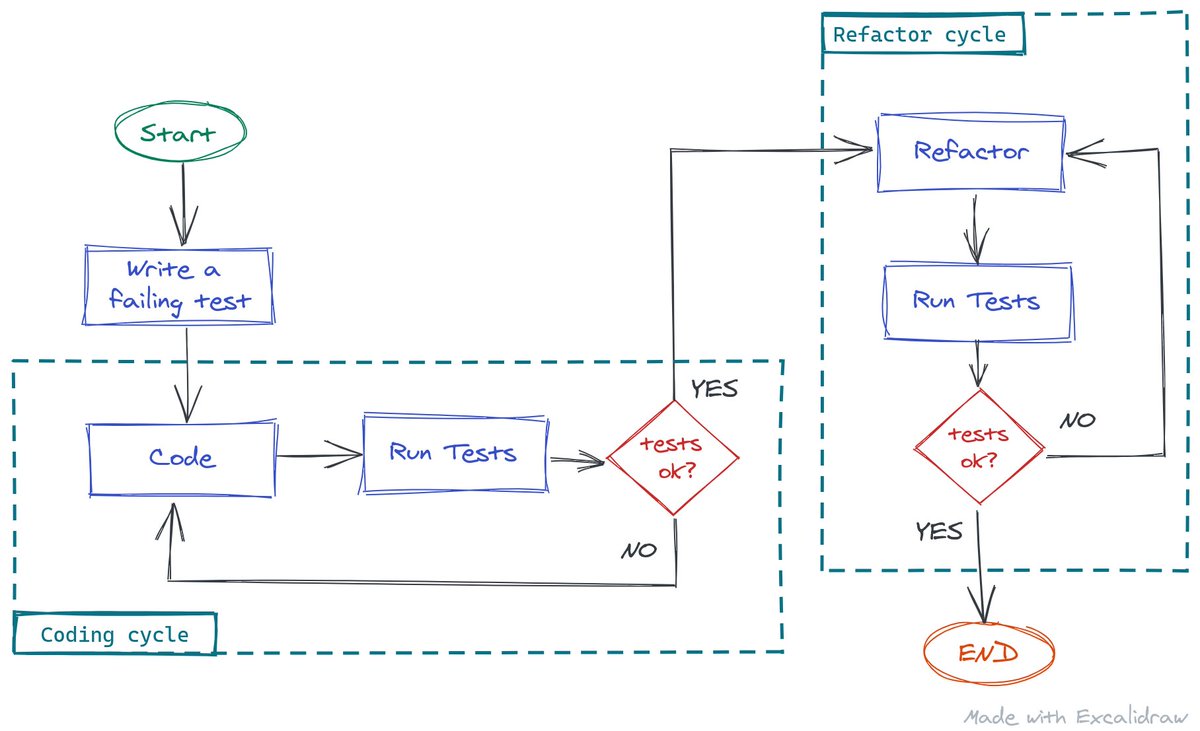} &
\includegraphics[width=\linewidth,height=0.12\textheight,keepaspectratio]{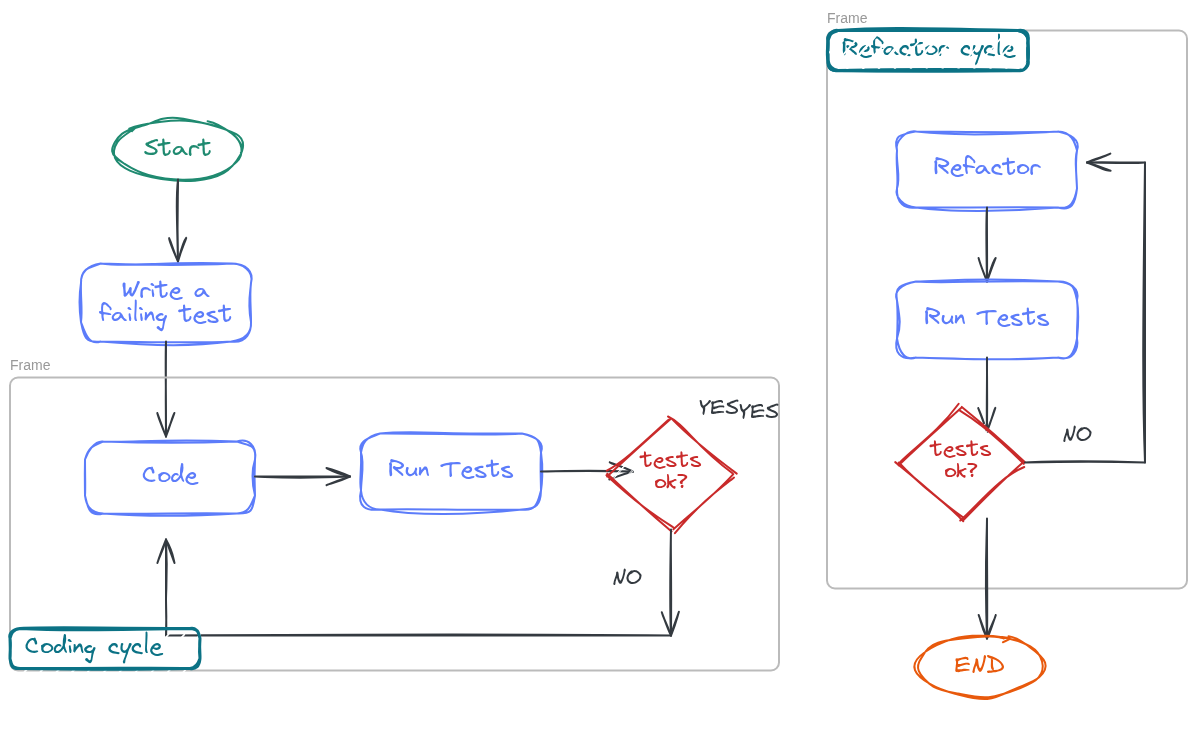} \\
\end{tabular}
\caption{Qualitative examples from the Excalidraw tool-use ablation. The model writes an Excalidraw scene JSON object, which is rendered with the official Excalidraw renderer.}
\label{fig:appendix-tool-use-excalidraw}
\end{figure}

\subsection{LaTeX Document Recreation}
\label{app:tool-use-latex}

\begin{figure}[H]
\centering
\scriptsize
\setlength{\tabcolsep}{3pt}
\renewcommand{\arraystretch}{0.92}
\begin{tabular}{@{}p{0.48\linewidth}p{0.48\linewidth}@{}}
\centering\textbf{Source} & \centering\arraybackslash\textbf{Recreated from generated LaTeX} \\
\includegraphics[width=\linewidth,height=0.4\textheight,keepaspectratio]{figures/tool_ablation/latex/1_source.png} &
\includegraphics[width=\linewidth,height=0.4\textheight,keepaspectratio]{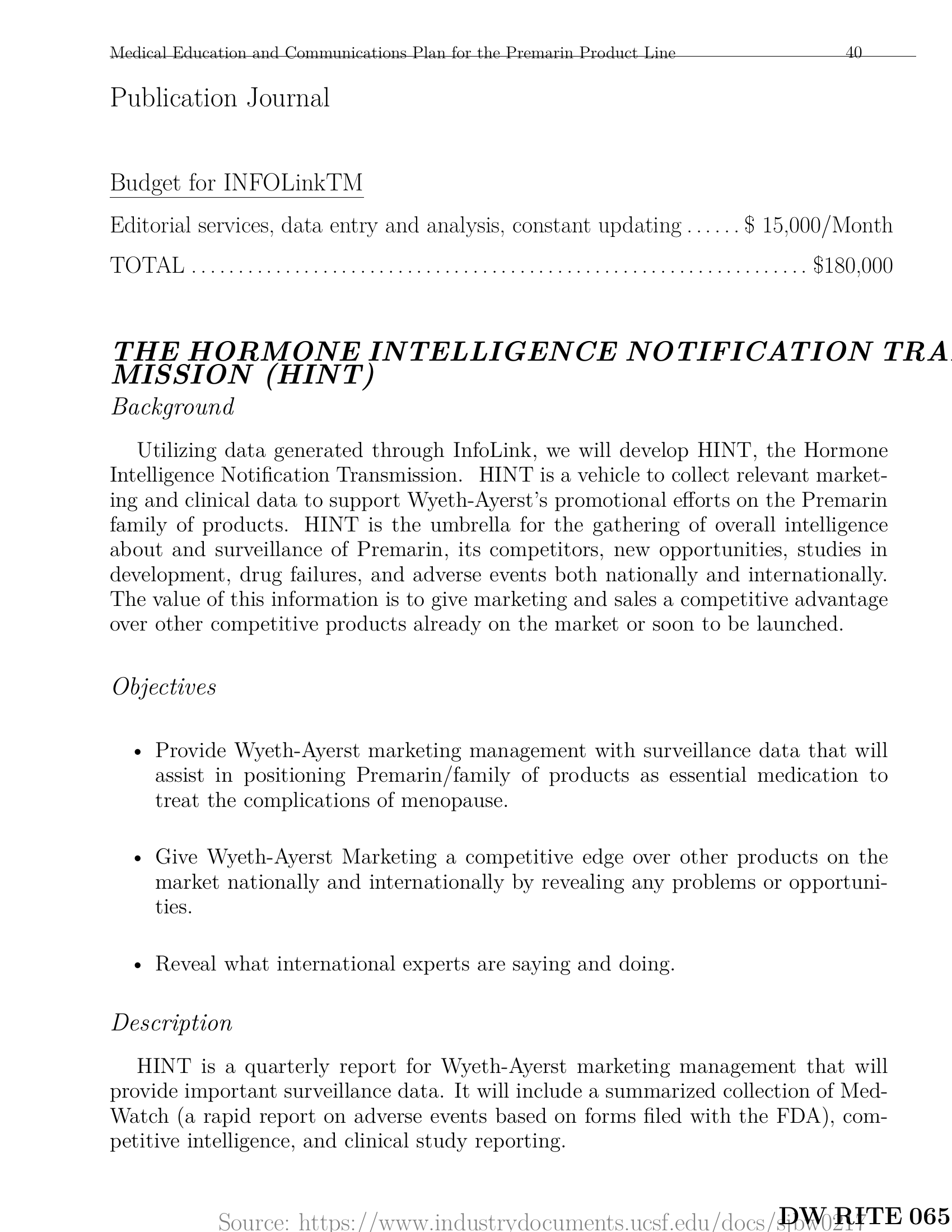} \\
\includegraphics[width=\linewidth,height=0.4\textheight,keepaspectratio]{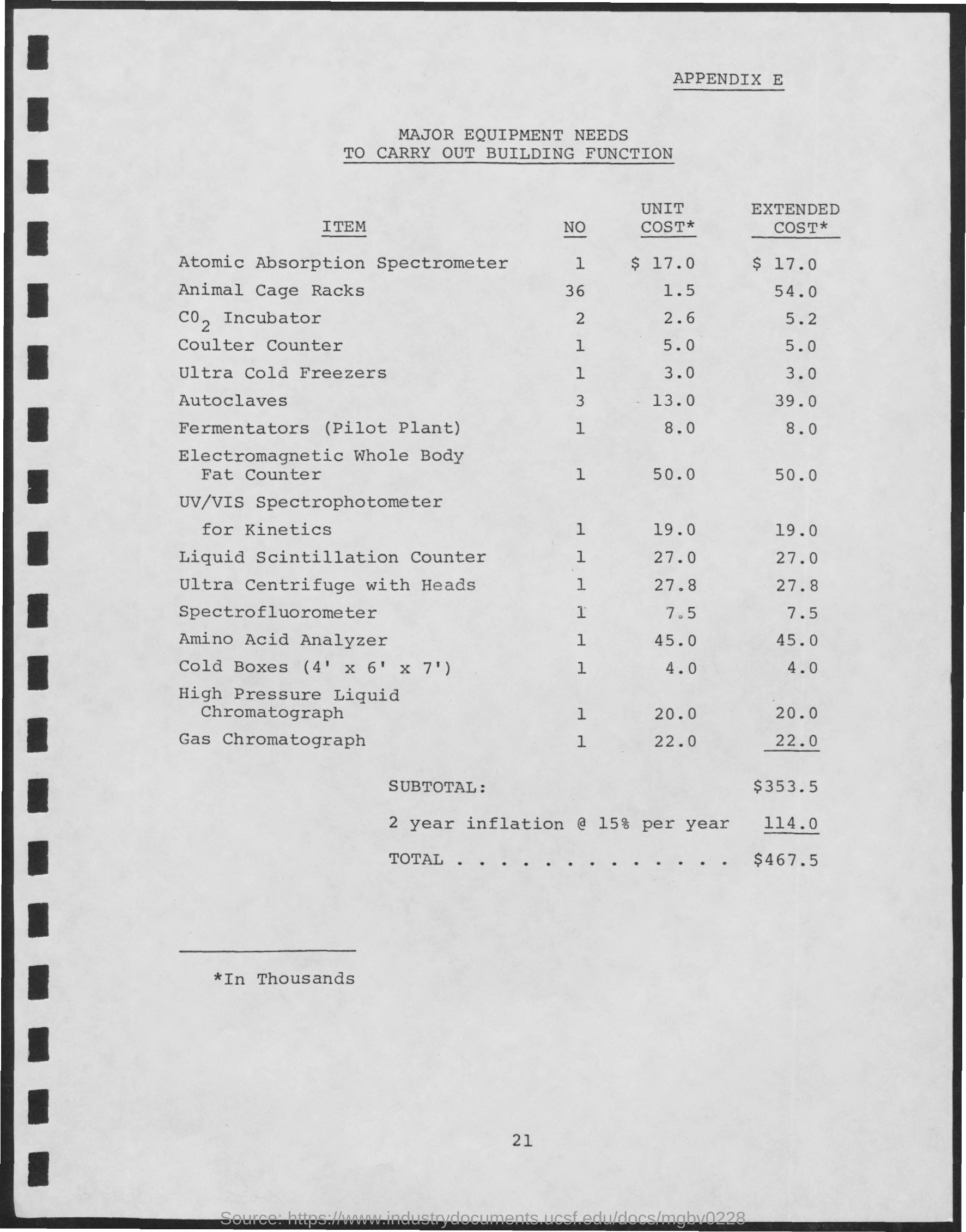} &
\includegraphics[width=\linewidth,height=0.4\textheight,keepaspectratio]{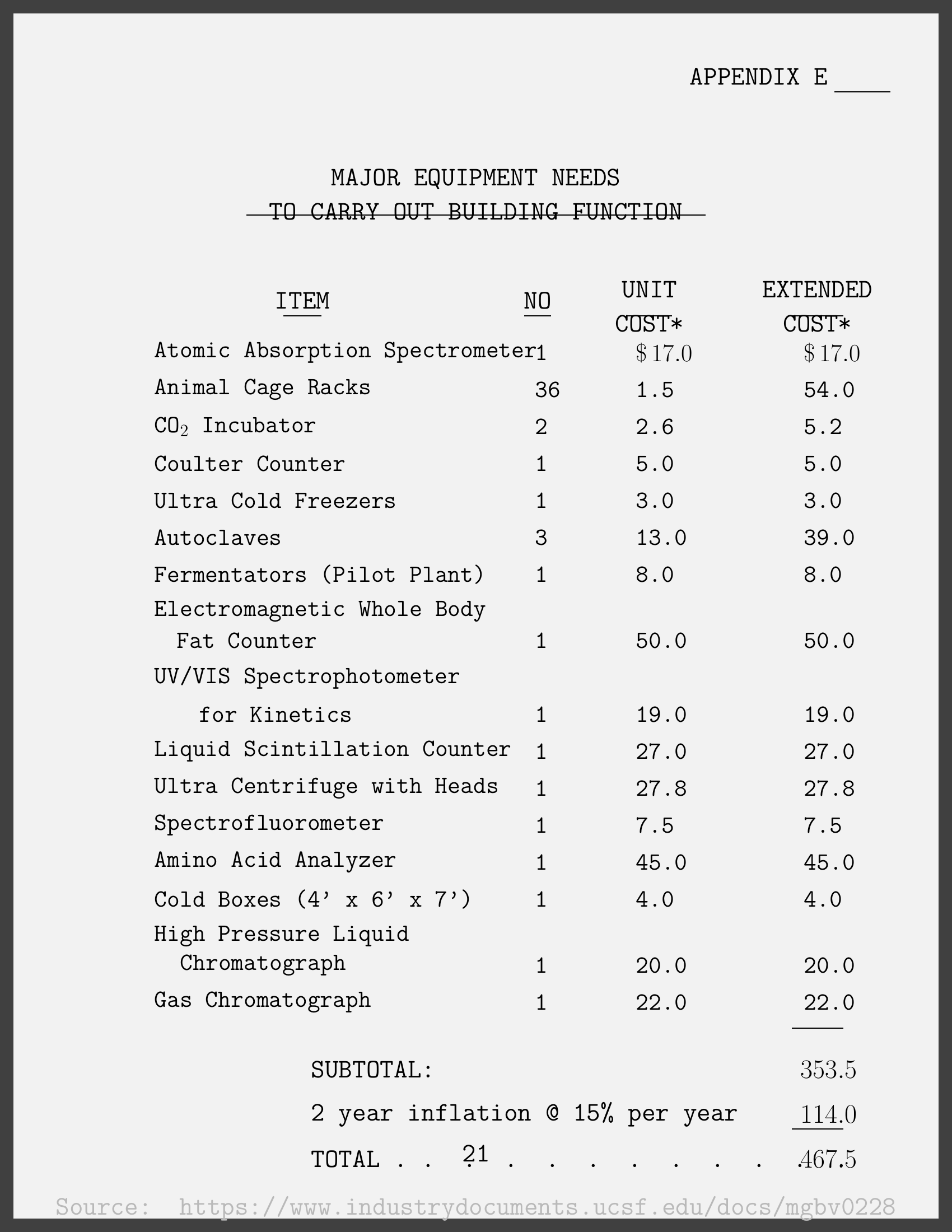} \\
\end{tabular}
\caption{Qualitative examples from the LaTeX document tool-use ablation. The model writes standalone LaTeX source, which is compiled and rasterized before evaluation.}
\label{fig:appendix-tool-use-latex}
\end{figure}

\end{document}